\documentclass[10pt,twocolumn,letterpaper]{article}

\usepackage[pagenumbers]{cvpr} %

\usepackage[utf8]{inputenc} %
\usepackage[T1]{fontenc}    %
\usepackage{url}            %
\usepackage{booktabs}       %
\usepackage{amsfonts}       %
\usepackage{nicefrac}       %
\usepackage{microtype}      %
\usepackage{xcolor}         %
\usepackage{algorithm} %
\usepackage{algorithmicx} \usepackage{algcompatible}
\usepackage{algpseudocode}
\usepackage{amsmath}
\usepackage{graphicx}
\usepackage{array}
\usepackage{booktabs}
\usepackage{makecell}
\usepackage{adjustbox}
\usepackage{siunitx}
\usepackage{multirow}
\usepackage{graphicx}

\definecolor{cvprblue}{rgb}{0.21,0.49,0.74}
\usepackage[pagebackref,breaklinks,colorlinks,allcolors=cvprblue]{hyperref}

\def\paperID{21580} %
\def\confName{CVPR}
\def\confYear{2026}

\title{Identifying Models Behind Text-to-Image Leaderboards}

\author{Ali Naseh\thanks{Correspondence to anaseh@cs.umass.edu}\\
University of Massachusetts Amherst\\
\and
Yuefeng Peng\\
University of Massachusetts Amherst\\
\and
Anshuman Suri\\
Northeastern University\\
\and
Harsh Chaudhari\\
Northeastern University\\
\and
Alina Oprea\\
Northeastern University\\
\and
Amir Houmansadr\\
University of Massachusetts Amherst
}

\begin{document}
\maketitle
\begin{abstract}
Text-to-image (T2I) models are increasingly popular, producing a large share of AI-generated images online. To compare model quality, voting-based leaderboards have become the standard, relying on anonymized model outputs for fairness. In this work, we show that such anonymity can be easily broken. We find that generations from each T2I model form distinctive clusters in the image embedding space, enabling accurate deanonymization without prompt control or training data. Using 22 models and 280 prompts (150K images), our centroid-based method achieves high accuracy and reveals systematic model-specific signatures. We further introduce a prompt-level distinguishability metric and conduct large-scale analyses showing how certain prompts can lead to near-perfect distinguishability. Our findings expose fundamental security flaws in T2I leaderboards and motivate stronger anonymization defenses.
\end{abstract}
    
\section{Introduction}
\label{sec:intro}

Text-to-image (T2I) models have become increasingly popular and are widely used across creative \citep{zhou2024generative}, professional \citep{zhang2025generative,park2024we}, and social media  \citep{moller2025impact} platforms. A substantial fraction of online images are now AI-generated \citep{yang2024characteristics,matatov2024examining,wei2024understanding}, reflecting the growing influence of these models in creating visual content. Given this widespread adoption, a natural question arises: \emph{how should we evaluate and compare these models?} This question has made leaderboards a central mechanism for benchmarking and comparing generative models.

Generative model leaderboards typically fall into two categories: benchmark-based \citep{mteb_leaderboard, alpaca_eval, lai2023openllmbenchmark, paech2023eq} and voting-based \citep{mteb_arena, 2023ChatArena, artificial}. Benchmark-based leaderboards rely on private or standardized test sets to evaluate models quantitatively. However, for image generation, automatic metrics often correlate poorly with human judgment \citep{corneanu2025structured, lee2023holistic}, making purely benchmark-driven evaluations unreliable. Consequently, voting-based leaderboards have become the dominant evaluation paradigm for T2I models. In these leaderboards, users are provided two anonymized generations produced by different models along with the corresponding prompt, and are asked to vote for the better image. Model anonymity is therefore vital for a fair evaluation process.

Recent research \citep{huang2025exploring, min2025improving,suri2026exploiting} has highlighted the vulnerabilities of leaderboards for text-based large language models (LLMs), showing that adversaries can manipulate rankings by deanonymizing models and strategically upvoting or downvoting specific models to alter their rankings. Yet, the deanonymization of \emph{state-of-the-art} T2I models in leaderboard settings remains largely unexplored. In this work, we demonstrate that in the T2I modality, deanonymization is significantly easier than in text-based leaderboards, even under minimal adversarial assumptions. We show that an adversary with no control over prompts and only black-box access to model APIs can successfully infer which model generated a given leaderboard image.

Our hypothesis stems from the observation that different T2I models, trained on distinct datasets, architectures, and parameter scales, exhibit characteristic generation patterns for the same textual prompt. While generations from different models tend to diverge systematically in style, composition, or fine-grained features, generations from the \emph{same} model for a fixed prompt are typically consistent and exhibit low intra-model variance (see \Cref{fig:prompt_diversity}). Thus, in a semantically meaningful embedding space, images from one model form tight clusters that are well separated from others. We find that modern image encoders provide a strong representation for capturing these differences. To further quantify this phenomenon, we define a \emph{prompt-level distinguishability metric} that measures how well the generations of different models separate in the embedding space, enabling us to identify prompts that reveal maximal model-specific signals.

\begin{figure*}[ht]
    \centering
    \includegraphics[width=0.9\textwidth]{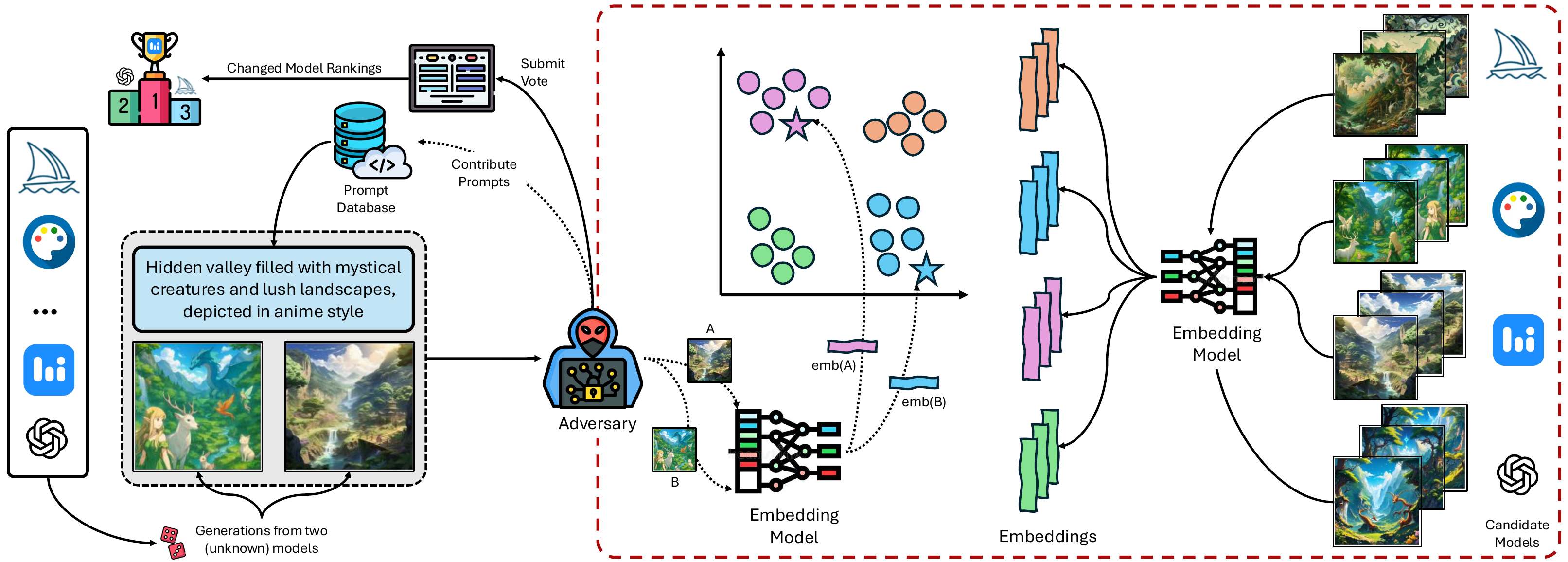}
    \caption{Overview of our setting: (a) voting-based leaderboard, (b) adversarial deanonymization pipeline, and (c) model-specific clustering in embedding space.}

    \label{fig:overview}
\end{figure*}

Our experiments span 280 prompts collected from Artificial Analysis, a popular T2I leaderboard \citep{artificial} and 22 state-of-the-art T2I models representing diverse architectures, providers, and model sizes, yielding over 150{,}000 generated images. We confirm that model-specific signatures are pervasive and can be exploited for high-accuracy deanonymization. Beyond our training-free embedding-based approach, we evaluate multiple fingerprinting and attribution techniques proposed in prior works \citep{marra2019gans, dzanic2020fourier} and find that simple centroid-based classification can outperform all of them. We also compare against supervised training-based methods, finding that they generalize poorly and offer limited benefits. Extending our analysis to a large-scale dataset exceeding two million generated images, we provide a comprehensive study of prompt-level distinguishability across topics. Finally, we discuss potential countermeasures for leaderboard design aimed at reducing such vulnerabilities. We hope our findings encourage future leaderboard operators to adopt stronger anonymization and evaluation protocols for fair model comparison.

\section{Related Work}

\paragraph{Leaderboard Attacks.}
Leaderboards for generative AI are generally either benchmark-based \citep{muennighoff2023mteb, minixhofer2024ttsds} or voting-based \citep{tts-arena-v2} (\eg Chatbot Arena \citep{chiang2024chatbot}). Both types are vulnerable to manipulation attacks. \citet{huang2025exploring} demonstrate that malicious participants can deanonymize models in Chatbot Arena and artificially promote their own models through poisoned votes. \citet{zhao-etal-2025-challenges} show how inserting as little as 10\% adversarial/low-quality votes can shift a model’s rank by up to five places. \citet{min2025improving} analyze Elo-style rating systems, showing they can be gamed via ``omnipresent rigging,'' where a few hundred strategically placed votes can boost a model's rank substantially, even without targeting the victim directly. \citet{suri2026exploiting} show that benchmark leaderboards can be subverted by models trained on test sets. Prior works either target LLMs, whose deanonymization strategies may not apply to T2I models, or inject backdoors into models, an approach that may interfere with the model's training objective and only applies to models in an adversary's control.

\newcommand{\img}[1]{\raisebox{-0.5\height}{\includegraphics[height=1.5cm]{#1}}}

\begin{figure*}[t]
\centering
\renewcommand{\arraystretch}{1}
\setlength{\tabcolsep}{1pt}

\begin{minipage}{0.78\linewidth}
\centering
{\small \textbf{Prompt:} ``An impressionistic painting of a bustling city street in the rain, vibrant umbrellas dotting the crowd.''}
\end{minipage}

{\tiny
\resizebox{0.8\linewidth}{!}{%
\begin{tabular}{@{}>{\centering\arraybackslash}m{2.0cm}ccccc@{}}

\toprule
\textbf{Model} & \multicolumn{5}{c}{\textbf{Generated Images (five different seeds)}}\\
\midrule

\makecell[c]{FLUX.1-dev} &
\img{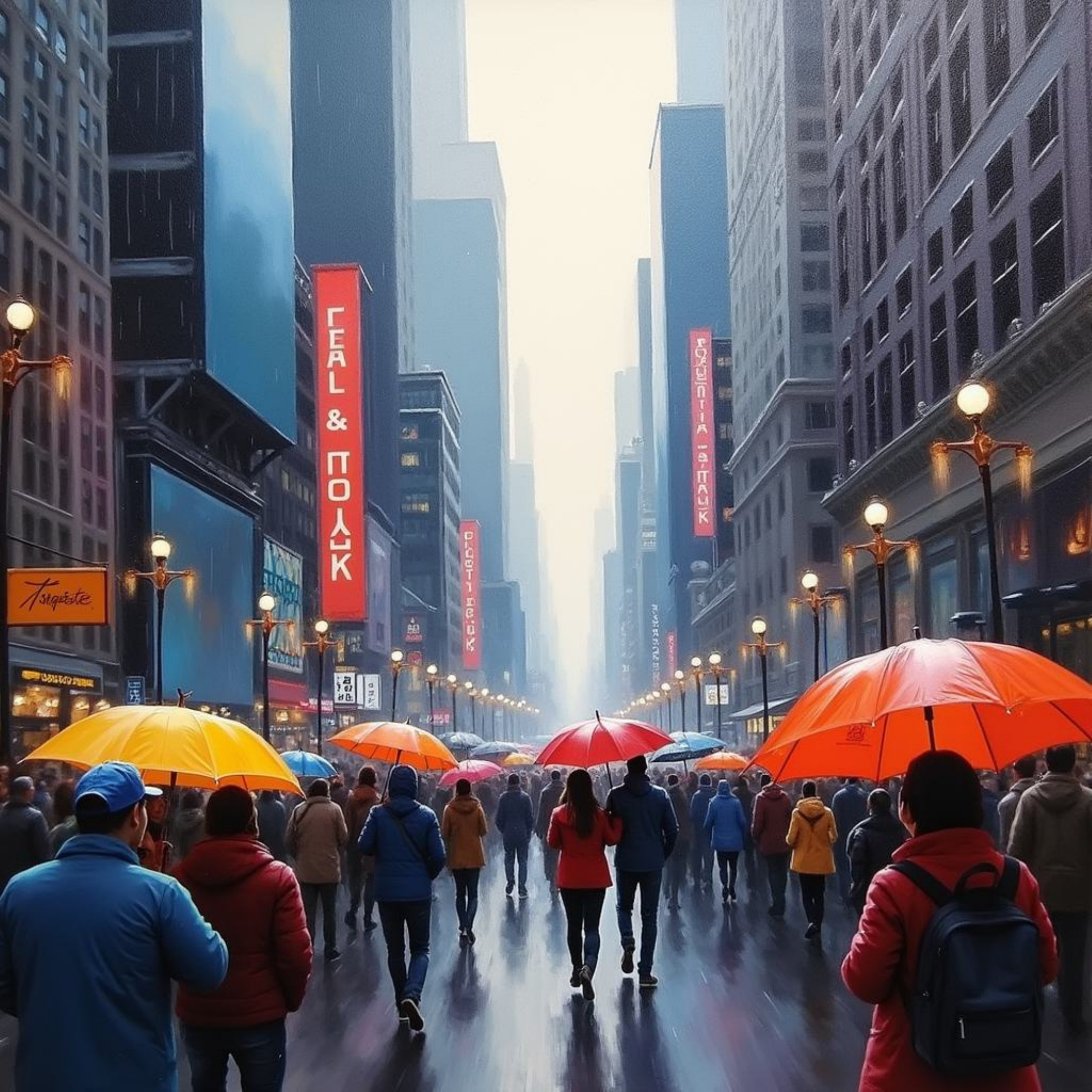} &
\img{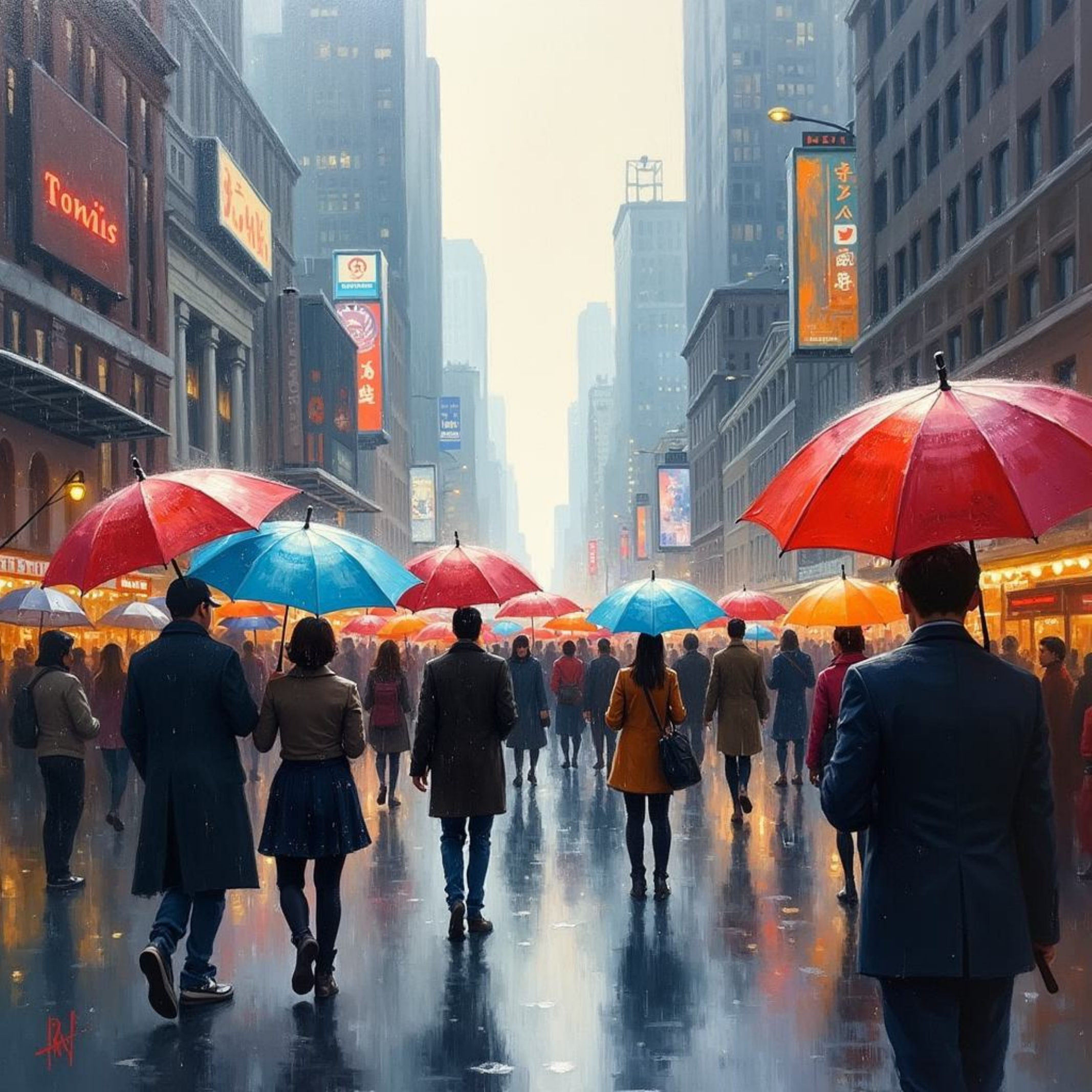} &
\img{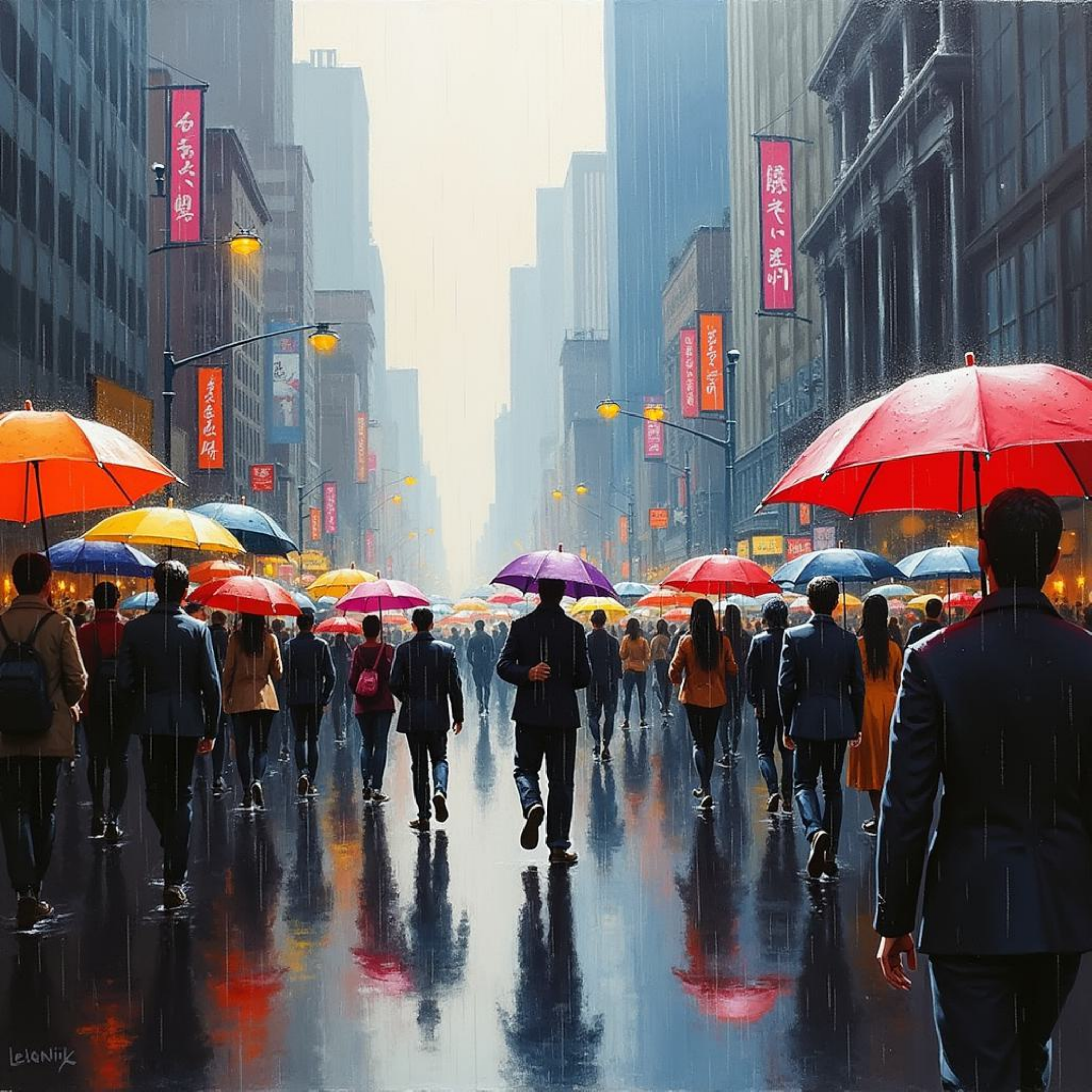} &
\img{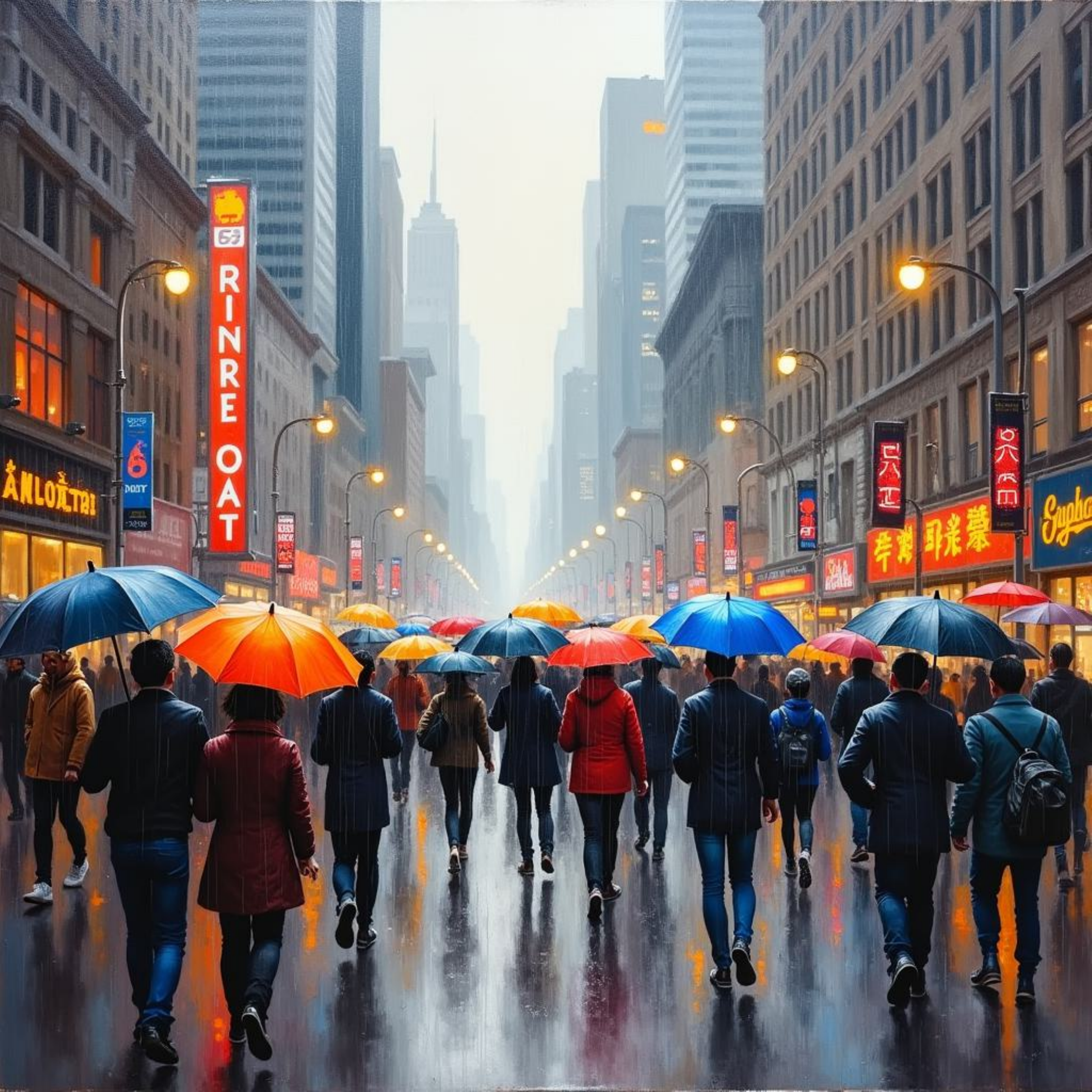} &
\img{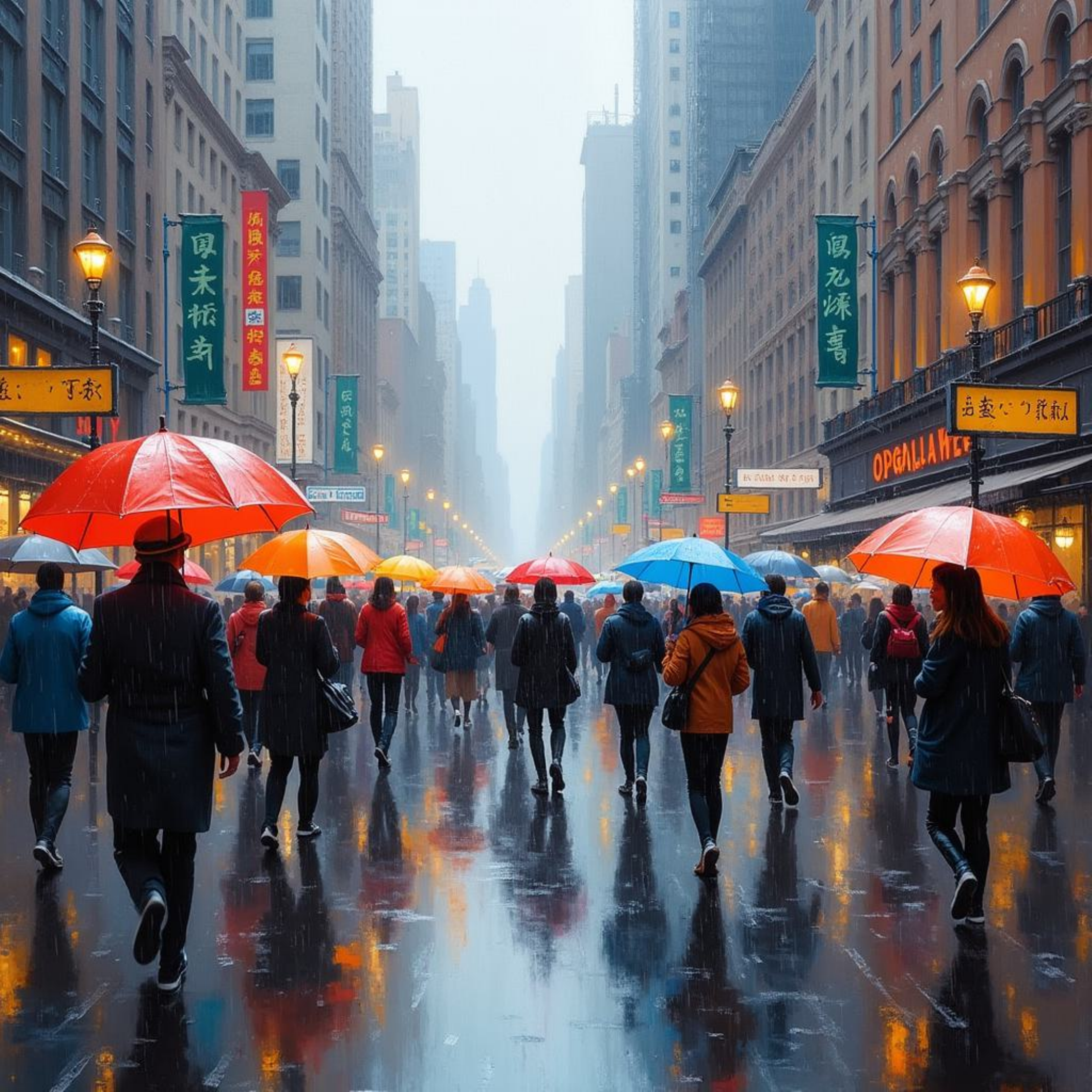} \\

\makecell[c]{SDXL Turbo} &
\img{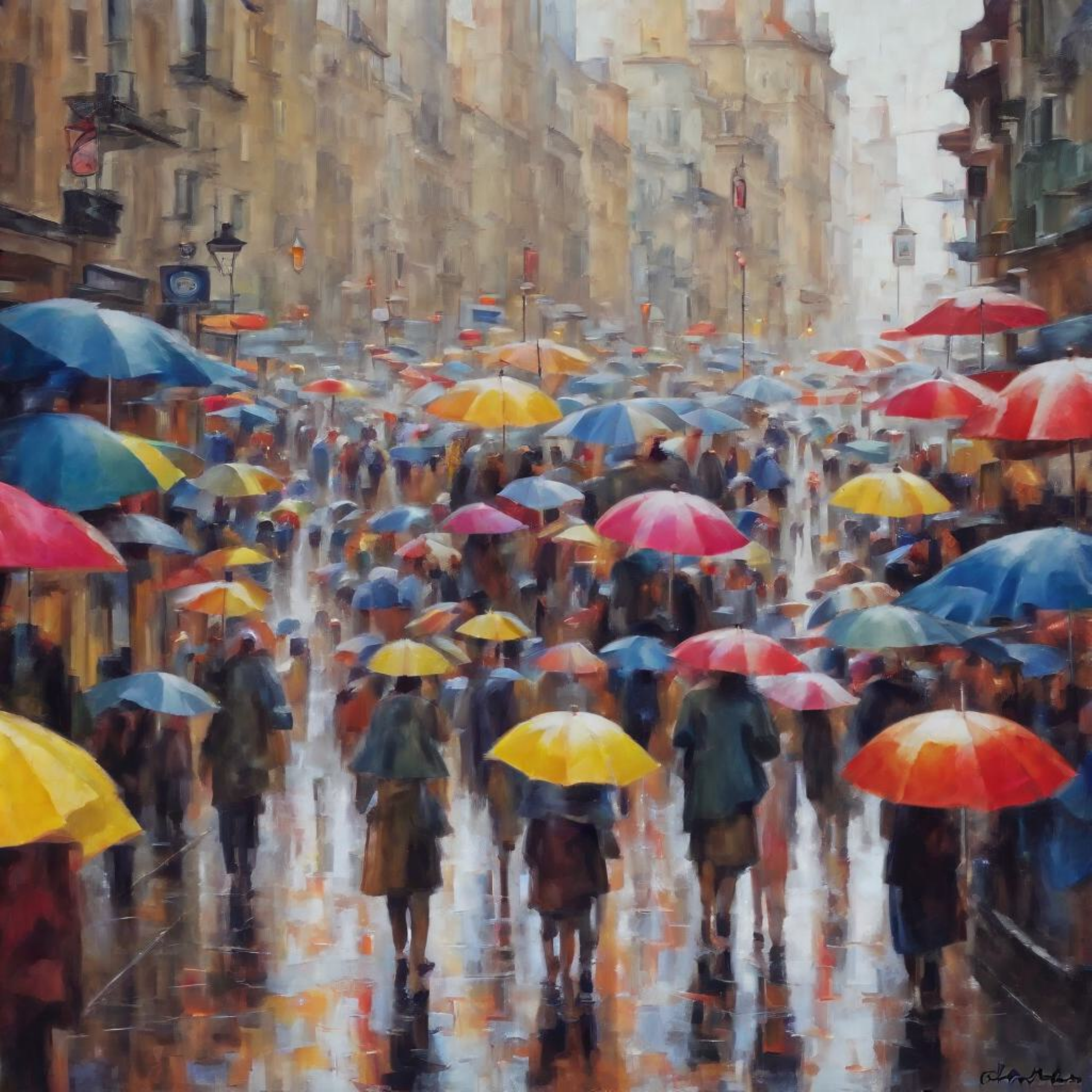} &
\img{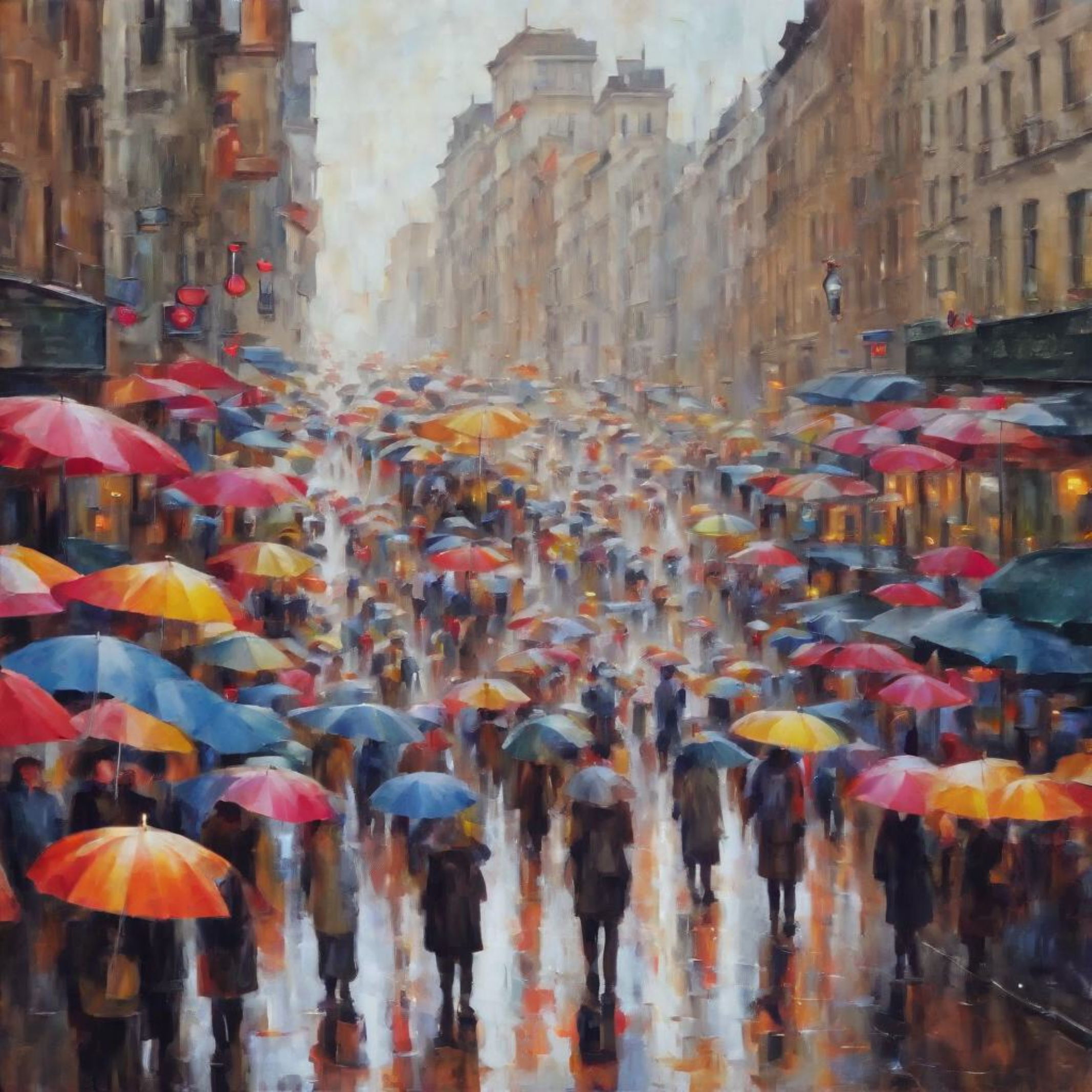} &
\img{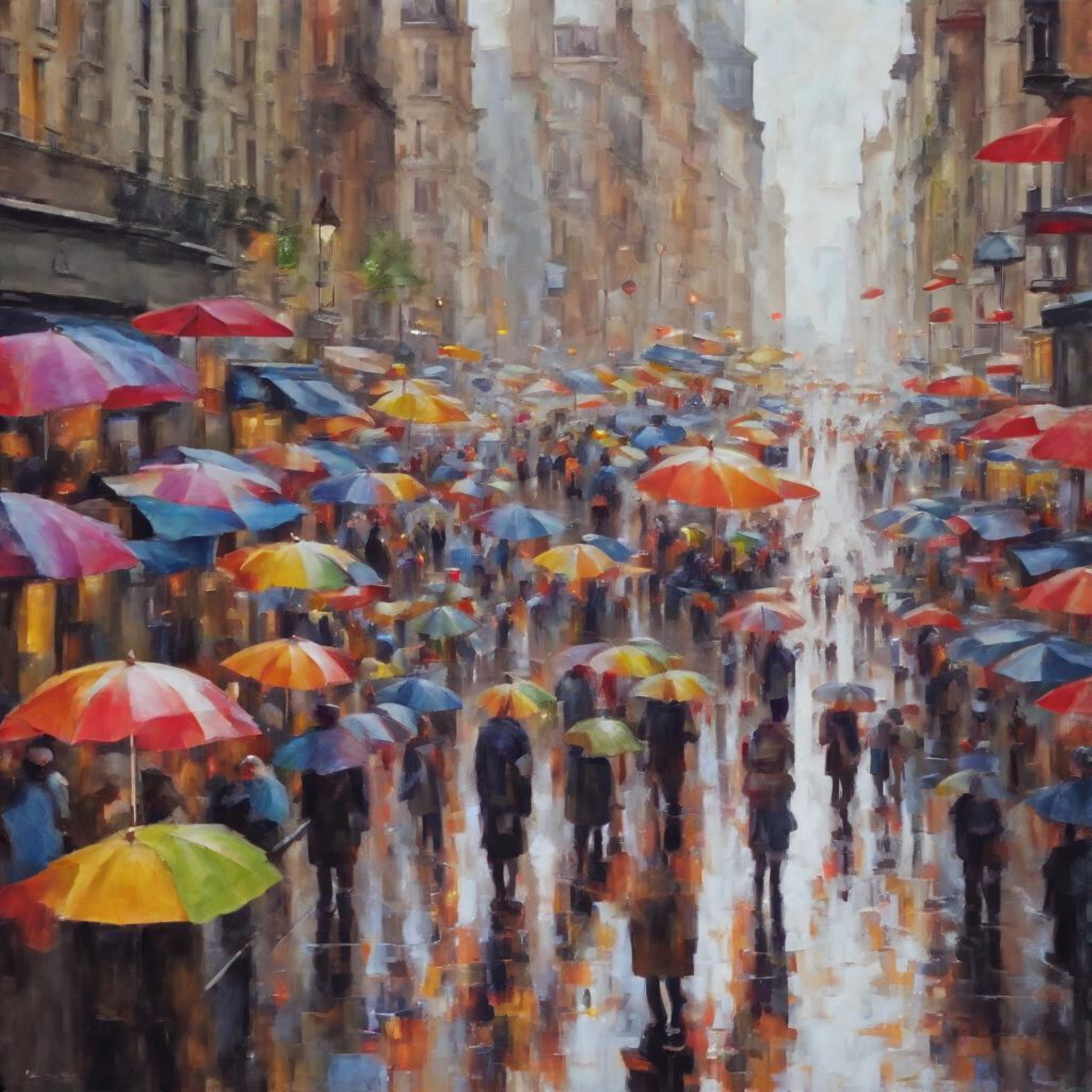} &
\img{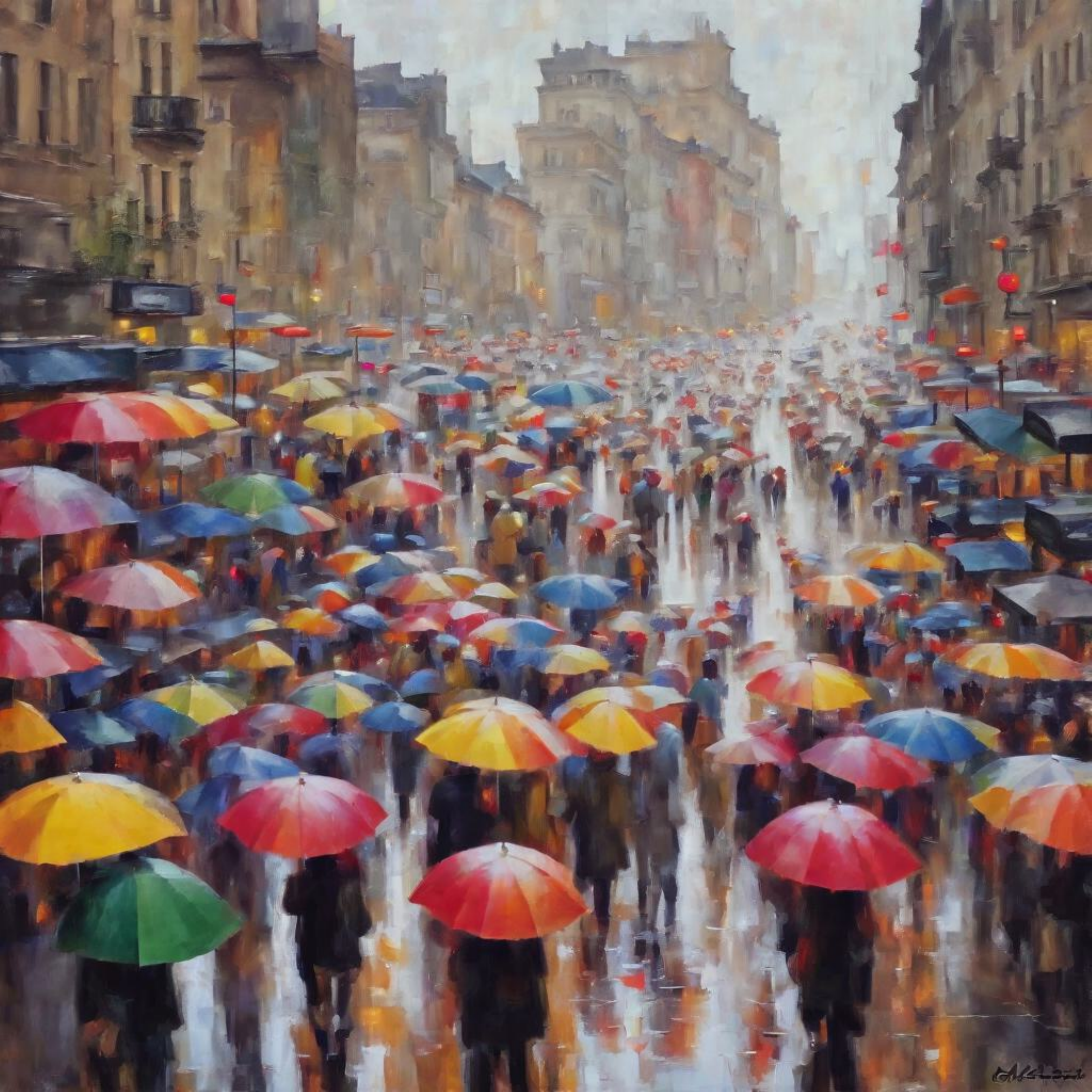} &
\img{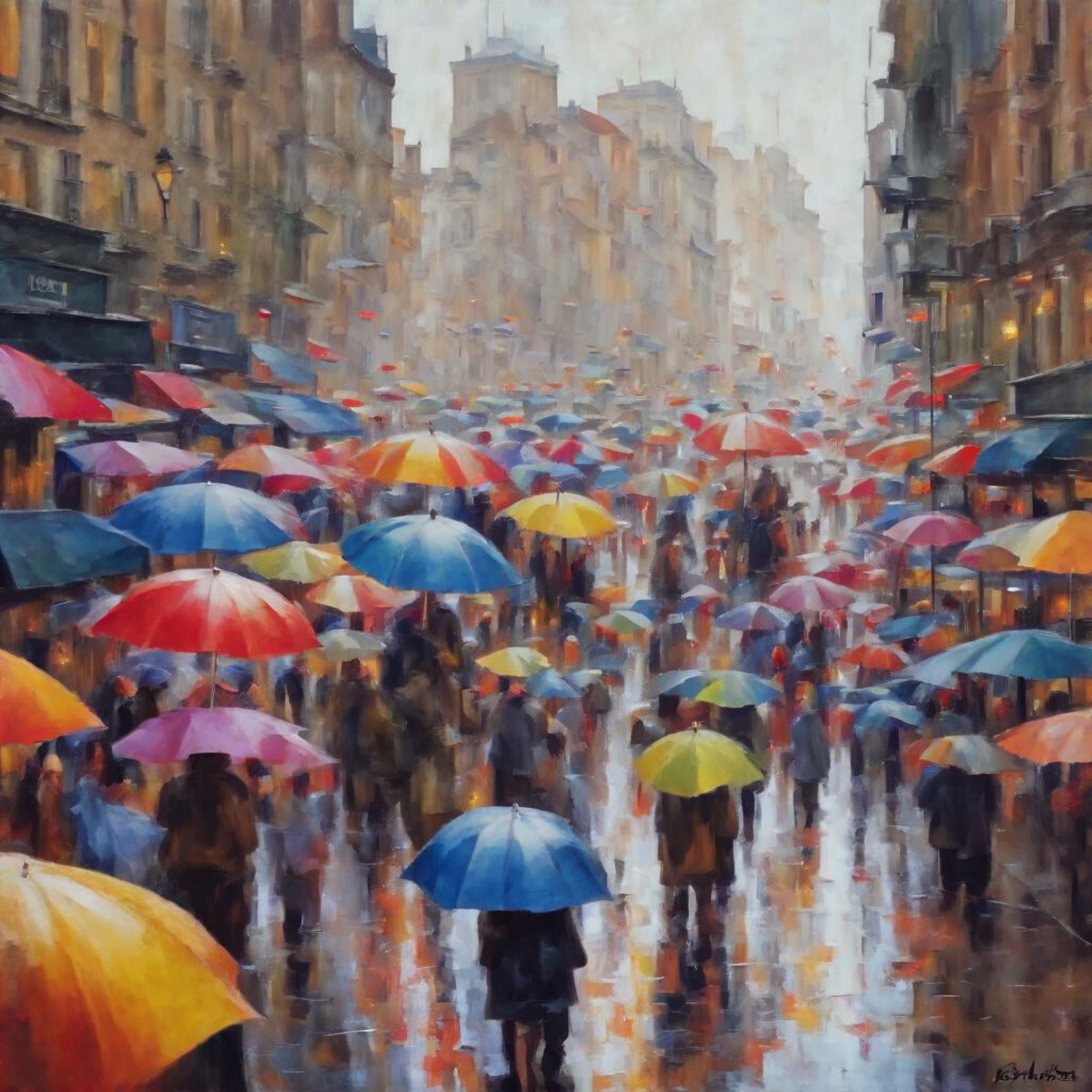} \\

\makecell[c]{Playground v2} &
\img{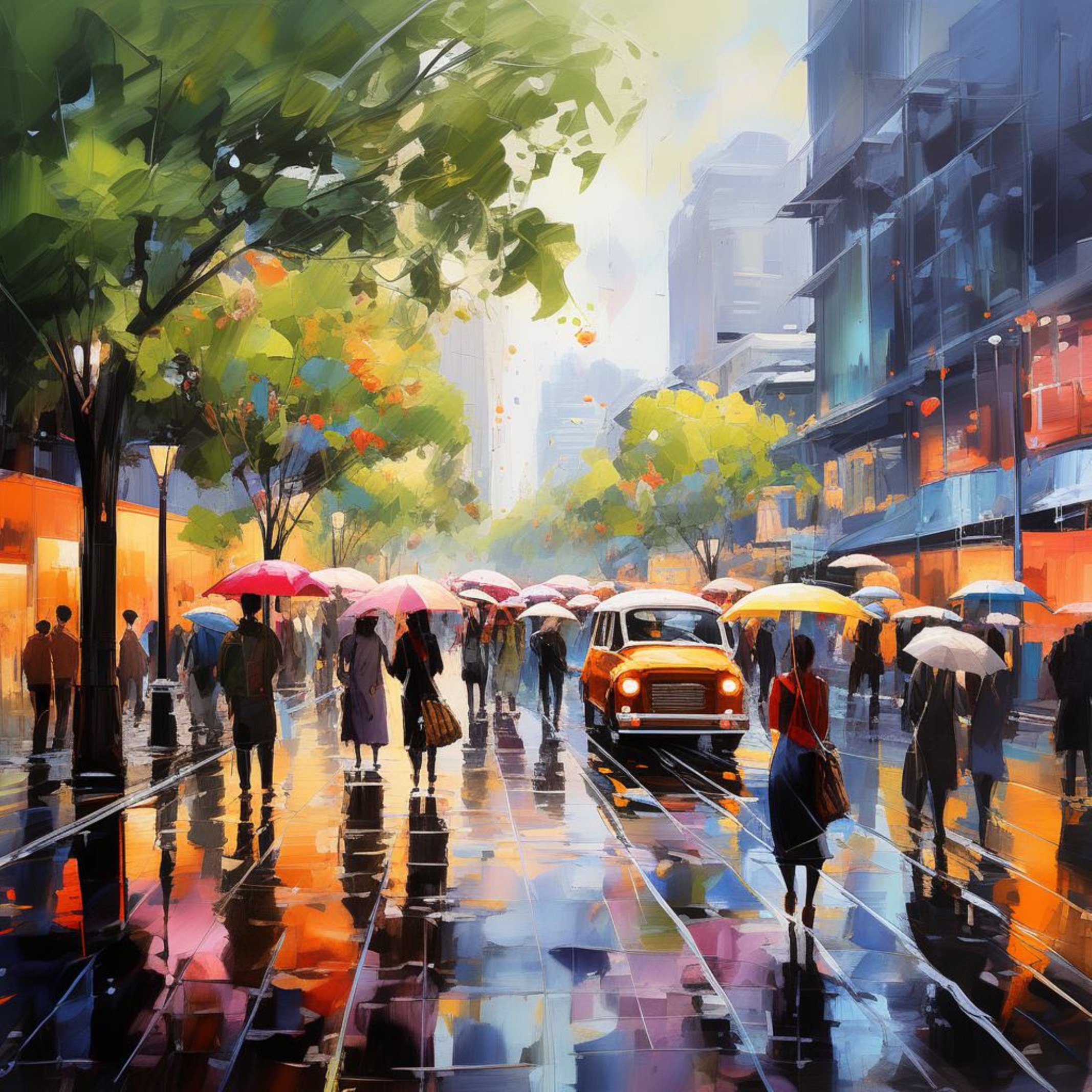} &
\img{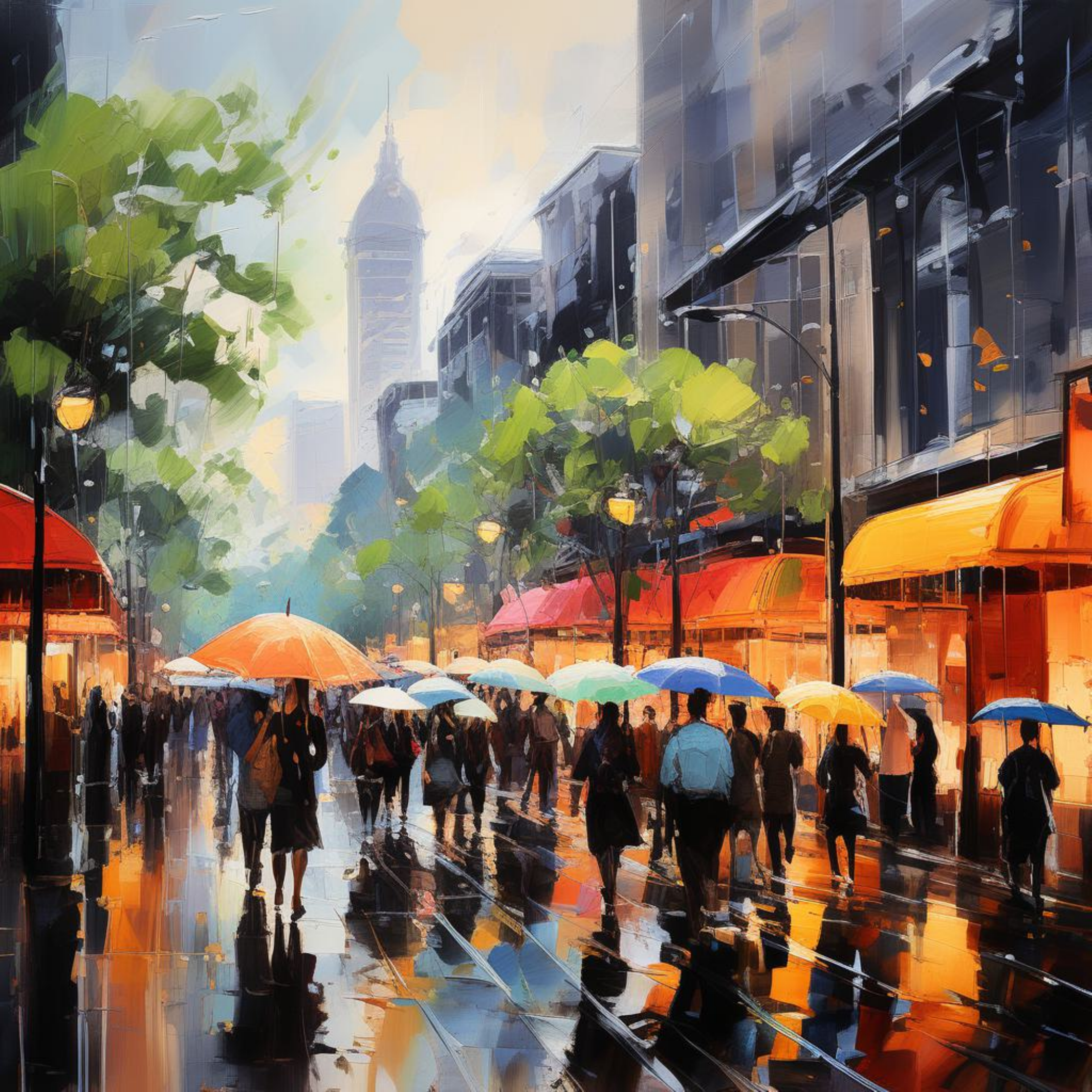} &
\img{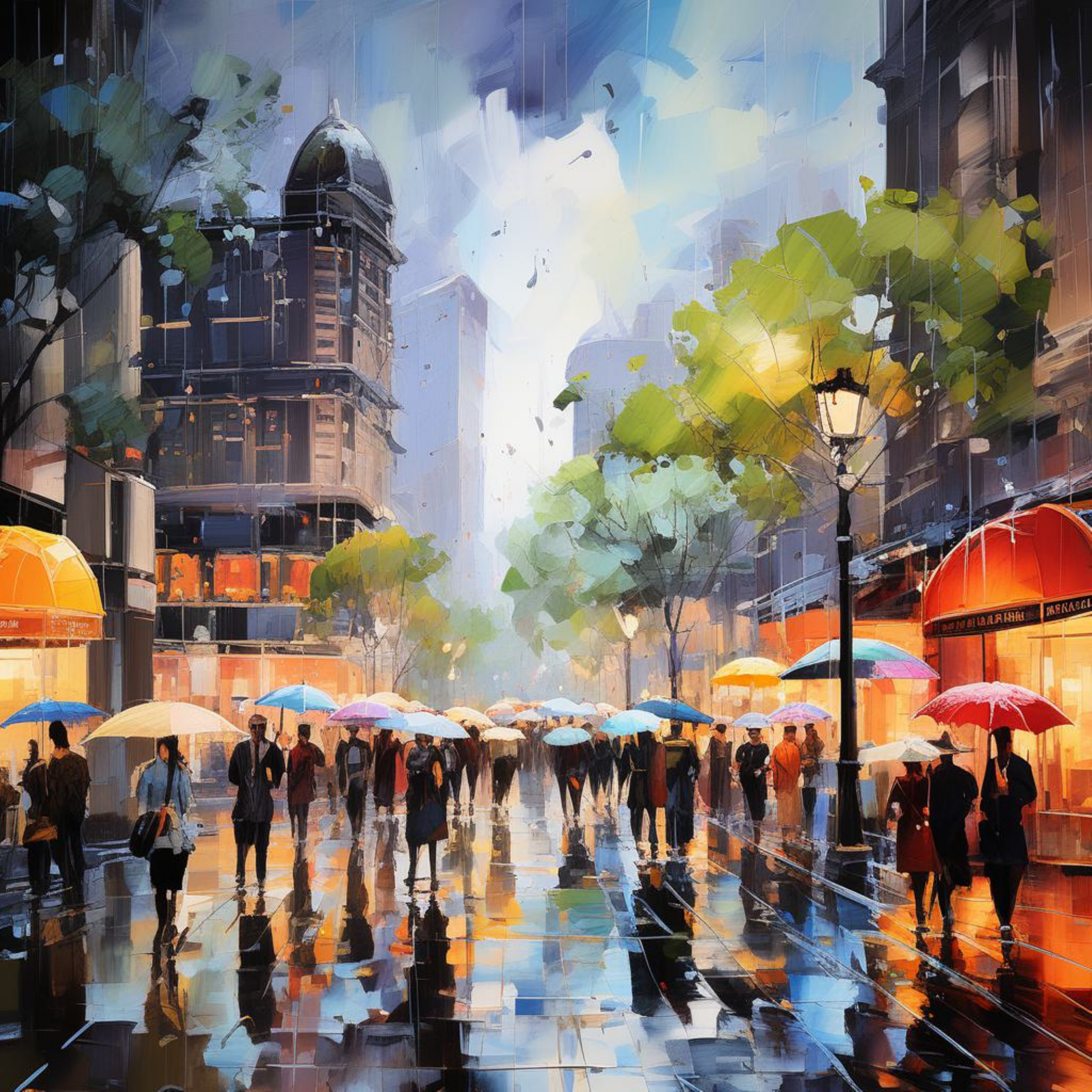} &
\img{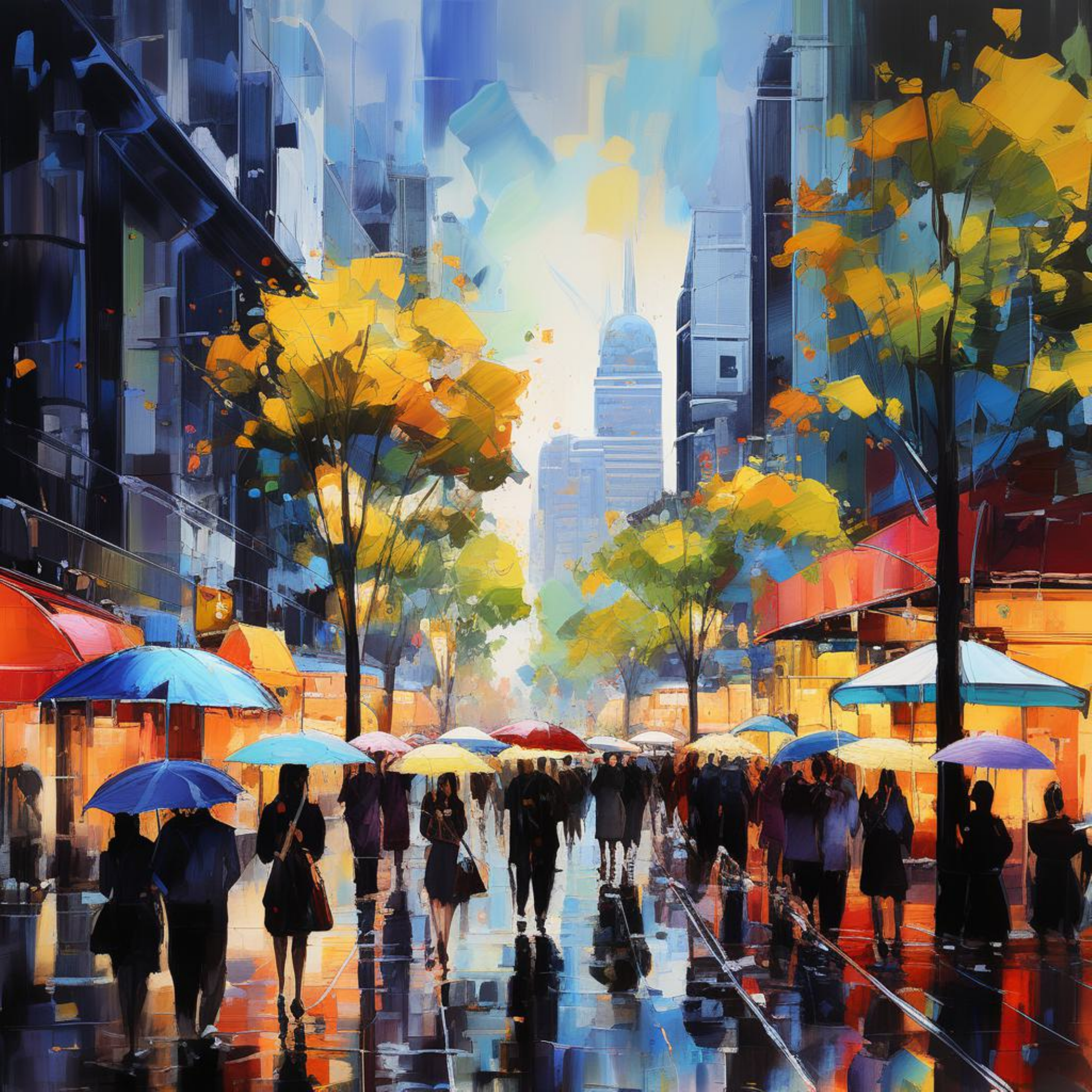} &
\img{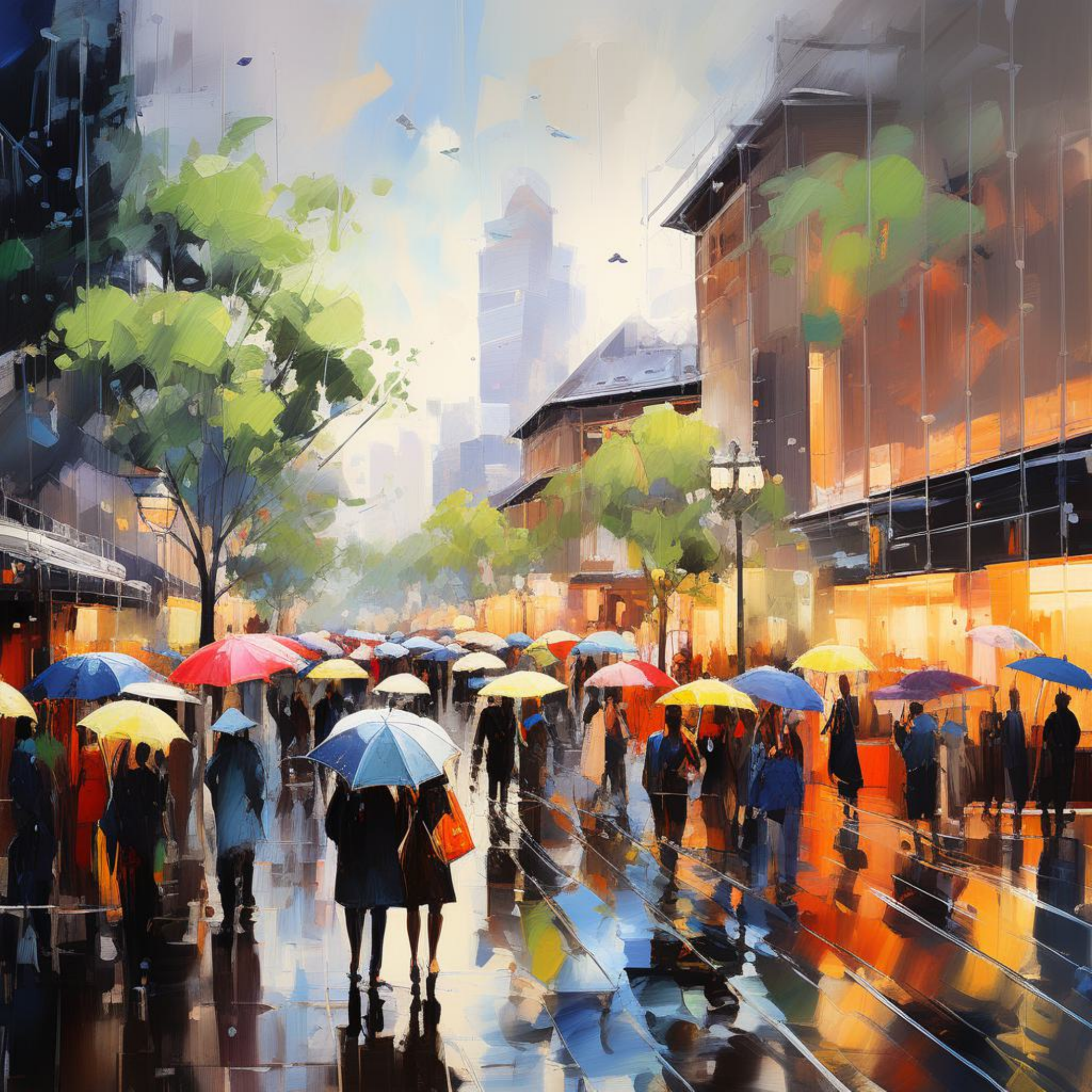} \\

\bottomrule
\end{tabular}%
} %
} %

\caption{Model-specific generation patterns for a fixed prompt, showing low intra-model diversity and clear inter-model differences.}
\label{fig:prompt_diversity}
\end{figure*}

\paragraph{Model Attribution.}
Model attribution seeks to identify the source model of a generated sample, typically assuming access through an API that can be queried. Earlier studies addressed GANs \citep{yu2019attributing} and generative text models \citep{pasquini2025llmmap}, while more recent work focuses on text-to-image systems. Some approaches rely on adversarial perturbations to differentiate models \citep{guo2024one}, requiring multiple API calls per sample, while others train large-scale classifiers on tens of thousands of generations \citep{yao2025authprint}, leading to poor generalization and high false positive rates on unconstrained prompts. Other attribution efforts include geometric or manifold-based fingerprinting, such as the Riemannian method of \citet{song2025riemannian} and ManiFPT \citep{song2024manifpt}, both of which assume access to real data and thousands of samples per model. A complementary line of work embeds fingerprints during model training or fine-tuning \citep{yu2021artificial, kim2024wouaf, nie2023attributing}, requiring access to model weights or datasets. These techniques aim at provenance verification rather than post hoc attribution, making them unsuitable for our black-box leaderboard scenario. In contrast, our approach operates entirely at inference time, without any training or access to real data, and shows that model-specific signatures naturally emerge in the embedding space, enabling practical deanonymization in realistic leaderboard environments.

\section{Threat Model}

In voting-based text-to-image (T2I) leaderboards, users are provided a pair of image generations without any information about the underlying generation model, along with the prompt used to generate these images. Users are then expected to vote for the ``better'' image. These prompts (and their corresponding image generations) are shown in random order, and users usually have no control over the prompts or the order in which they are shown. In most cases, users can also choose to abstain from voting and skip. Model anonymity is an essential component of the evaluation process, as it helps remove any possible bias relating to the underlying model/organization, as well as ensure that these arenas cannot be easily gamed.

\paragraph{Adversary’s Objective.}
The adversary aims to deanonymize the displayed generations \ie to infer which model produced each image. Once the generating model is identified, the adversary can cast strategic votes to promote or demote specific models, thereby manipulating the leaderboard ranking. We consider two different granularities for the objective:
\begin{enumerate}[label=(\alph*)]
    \item exactly identifying the underlying model,
    \item distinguishing a specific model from \textit{any} other model. 
\end{enumerate}
The first objective is the most generalizable, and aligns with open-ended adversaries that may want fine-grained control over ranking manipulation. The second objective aligns best with focused adversaries that may want to specifically upvote their own model or downvote models from competitors.

\paragraph{Adversary’s Capabilities.}
The adversary has no control over the selection or order in which models are selected by the arena for image generation. However, the adversary does know the list of participating models (publicly shown in the leaderboard rankings; see \Cref{fig:overview}). For the prompt used for image generation, we consider two levels of control:
\begin{enumerate}[label=(\alph*)]
    \item \textbf{No prompt control:} The adversary has no control over the prompts used by the leaderboard.
    \item \textbf{Can submit prompts:} The adversary can submit arbitrary prompts to the leaderboard.
\end{enumerate}
The former corresponds to the default case, where most voting-based T2I leaderboards \citep{artificial, 2023ChatArena} do not allow control over the prompt, while the latter helps measure worst-case deanonymization that an adversary could achieve.

Models on leaderboards are available either as open-source models that can be freely downloaded and used, or available via API calls. However, an adversary that only wants to distinguish its own model (or family of models) from others may not want to (or have) resources to reproduce generations for other models. Thus, we additionally consider a \textbf{resource-constrained} adversary that is limited only to using its own model or family of models.

\section{Embedding-Based Deanonymization Attack}
\label{sec:method}

\subsection{Intuition}
The behavior of generative models, including T2I systems, is shaped by their training data, architecture, and scale \citep{li2024scalability}. These factors yield characteristic generation patterns that manifest even when models are exposed to identical prompts. As a result, generations from different models may vary not only in the elements explicitly specified in the prompt but also in aspects left unspecified: style, composition, color palette, or fine-grained textures.

Despite the stochastic nature of T2I sampling, generations produced by the \emph{same} model for a fixed prompt tend to exhibit low intra-model variability: sampling different random seeds results in images with subtle variations. Conversely, generations produced by \emph{different} models often exhibit high inter-model variability (see \Cref{fig:prompt_diversity}): stylistic or perceptual attributes differ in consistent and model-specific ways.
We refer to these two types of differences as:
\begin{itemize}
    \item \textbf{Intra-model variation:} variability between generations of a single model for the same prompt.
    \item \textbf{Inter-model variation:} variability between generations of different models for the same prompt.
\end{itemize}

Empirically, inter-model variation typically dominates intra-model variation. This asymmetry implies that, in a suitable representation space, generations from each model form tight and separable clusters. In this work, we show that suitable image-embedding spaces capture semantic and stylistic differences more effectively than raw pixels. This separation enables model-specific clustering and forms the basis of our deanonymization strategy (see \Cref{fig:overview}).

\subsection{Deanonymization Method}

We formalize the deanonymization problem as follows. Let $\mathcal{C} = \{M_1,\dots,M_n\}$ denote the set of participating T2I models, and let $p$ be the prompt used by the leaderboard. The adversary observes a leaderboard-provided image $I^\ast$ and aims to infer its generating model without access to model identities during voting.

Our approach, based on nearest-centroid classification \cite{manning2008vector}, relies on an image encoder $\phi(\cdot)$ (e.g., CLIP) that maps images into an embedding space where model-specific signatures become evident. For each model $M_i$, the adversary uses the prompt $p$ to generate $k$ reference images:
\[
I_{i,1}, \dots, I_{i,k}.
\]
Each reference image is embedded as $e_{i,j} = \phi(I_{i,j})$. The adversary then computes a centroid
\[
c_i = \frac{1}{k} \sum_{j=1}^k e_{i,j},
\]
representing the typical embedding of model $M_i$ for prompt $p$.
Given the embedding $e^\ast = \phi(I^\ast)$ of the leaderboard image, the adversary computes its distance to each centroid and predicts the model producing the closest cluster:
\[
\hat{M} = \arg\min_{M_i \in \mathcal{C}} \| e^\ast - c_i \|_2.
\]
This centroid-based approach exploits the low intra-model variation and high inter-model separation present in the embedding space. In practice, it achieves high deanonymization accuracy even for small $k$, and it requires no training, prompting, or prior data collection.

\subsection{One-vs-Rest Deanonymization}

In some scenarios, the adversary is only interested in determining whether the displayed image originates from a specific target model $M^\ast$. We consider two variants of this setting depending on the adversary’s resource constraints.

\paragraph{Case 1: Access to All Models.}
If the adversary can query all models in $\mathcal{C}$, the problem reduces to a binary decision built on the centroid-based approach introduced earlier. The adversary generates $k$ reference images for each model, embeds them, and computes their centroids $\{c_1,\dots,c_{|\mathcal{C}|}\}$. Given the embedding $e^\ast$ of the leaderboard image, the adversary predicts that $M^\ast$ generated the image if
\[
\arg\min_{M_i\in\mathcal{C}} \| e^\ast - c_i \|_2 = M^\ast,
\]
otherwise it predicts that the image was produced by a different model.

\paragraph{Case 2: Access Only to the Target Model.}\label{sec:case2}
In the most restrictive setting, the adversary can query only its target model $M^\ast$. For the prompt $p$, the adversary generates $k$ reference images, computes their embeddings $\{x_1,\dots,x_k\}$, and forms a centroid $c$.
We define a threshold based on the $\alpha$-quantile of in-cluster distances:
\[
    \lambda_\alpha = \mathrm{quantile}_\alpha\!\left(\|x_i - c\|_2\right).
\]
Given a test embedding $z$, we compute its distance to the centroid and use the followign rule to classify the image as originating from $M^\ast$:
\[
\|z - c\|_2 \le \lambda_\alpha
\]
This one-vs-rest formulation enables detection even without access to any other models, highlighting how strongly model-specific signatures can manifest in embedding space.

\subsection{Distinguishability Metric}
\label{sec:distinguishability}

To quantify how separable different models' generations are in the embedding space, we introduce a \emph{distinguishability} metric that evaluates prompt-level separability. This metric helps identify prompts that naturally expose stronger model-specific signatures.

\paragraph{Model-level Separability.}
For each prompt $p_i$ and model $M_j$, let $\{e_{i,j}^{(1)}, \dots, e_{i,j}^{(k)}\}$ denote the embeddings of $k$ images generated by $M_j$ on $p_i$. For each embedding $e_{i,j}^{(\ell)}$, we identify its nearest neighbor in the joint embedding set of all models for the same prompt. If the nearest neighbor also belongs to $M_j$, we treat $e_{i,j}^{(\ell)}$ as correctly clustered. Formally,
\[
\text{frac}(i,j)
    = \frac{1}{k} \sum_{\ell=1}^{k}
      \mathbb{I}\!\left[\mathrm{NN}(e_{i,j}^{(\ell)}) \in M_j\right],
\]
where $\mathrm{NN}(\cdot)$ is the nearest neighbor in embedding space. If $\text{frac}(i,j) > \tau$ for a fixed threshold $\tau \in (0,1)$, the cluster corresponding to $(p_i, M_j)$ is deemed \emph{separable}.

\paragraph{Prompt-level Distinguishability.}
The distinguishability score of prompt $p_i$ is defined as
\[
D(i) = \frac{1}{|\mathcal{C}|}
       \sum_{M_j \in \mathcal{C}}
       \mathbb{I}\!\left[\text{frac}(i,j) > \tau\right],
\]
\ie the fraction of models whose generations form separable clusters under prompt $p_i$.
High distinguishability scores reflect prompts where inter-model differences dominate intra-model variation, offering a principled way to rank prompts by vulnerability to deanonymization.

\section{Experiments}
\label{sec:results}

\begin{table*}[t]
    \centering
    \footnotesize
    \caption{Deanonymization performance for varying types of attacks.}
    \label{tab:attribution-results}
    \begin{tabular}{llccc}
        \toprule
        \multirow{2}{*}{\textbf{Category}} 
            & \multirow{2}{*}{\textbf{Method}} 
            & \multicolumn{3}{c}{\textbf{Accuracy} (\%)} \\
        \cmidrule(lr){3-5}
         & & Top-1 & Top-2 & Top-3 \\
        \midrule

        \multirow{2}{*}{\textbf{IT-FP}}
            & Marra et al.~\citep{marra2019gans}  
                & $24.40_{\pm 5.10}$ 
                & $31.20_{\pm 4.80}$ 
                & $36.50_{\pm 5.60}$ \\

            & Dzanic et al.~\citep{dzanic2020fourier} 
                & $11.79_{\pm 4.71}$ 
                & $21.57_{\pm 7.15}$ 
                & $28.43_{\pm 8.86}$ \\
        \midrule

        \multirow{3}{*}{\textbf{Classifier}}
            & Image 
                & $54.86_{\pm 4.91}$ 
                & $67.00_{\pm 4.73}$ 
                & $72.86_{\pm 4.37}$ \\

            & Image Embedding 
                & $43.00_{\pm 6.77}$ 
                & $55.86_{\pm 7.85}$ 
                & $63.36_{\pm 7.85}$ \\

            & Image + Text Embedding 
                & $42.50_{\pm 5.49}$ 
                & $57.71_{\pm 7.85}$ 
                & $65.50_{\pm 4.89}$ \\
        \midrule

        \multirow{1}{*}{\textbf{IT-Emb}}
            & Embedding-1 
                & $\mathbf{90.86}_{\pm 2.35}$ 
                & $\mathbf{96.14}_{\pm 1.71}$ 
                & $\mathbf{97.50}_{\pm 1.64}$ \\
        \bottomrule
    \end{tabular}
\end{table*}

\subsection{Setup}

\paragraph{Models and Evaluation Datasets.}
We consider 22 state-of-the-art text-to-image (T2I) models from eight different companies, spanning both open-weight and commercial closed-source systems. Our set includes models with distinct architectures as well as different variants (\eg model sizes) within the same architecture or company. The full list and detailed specifications are provided in \Cref{tab:models} (Appendix). Unless otherwise specified, all main experiments use a ViT image encoder \citep{ilharco_gabriel_2021_5143773} trained on LAION dataset \citep{schuhmann2022laion} to compute embeddings.
To evaluate our inference-time attack, we manually collected 280 prompts from one of the most popular public T2I leaderboards. These prompts cover diverse visual and conceptual themes, closely reflecting real user queries submitted to online platforms.

\paragraph{Generation Setup and Evaluation Metrics.}
For each prompt, we generate 30 images per model, resulting in $280 \times 22 \times 30=184,800$ total images. During evaluation, each prompt–model pair is treated as a separate instance, and we repeat all randomized prompt–model assignments five times to account for variability in generation and sampling. Reported metrics are averaged across these repetitions. We evaluate all attacks using top-1, top-2, and top-3 identification accuracy. These metrics quantify the fraction of test images for which the correct generating model appears within the top-$k$ nearest centroids.

\subsection{Baselines}
Since there is no current established method specifically for deanonymizing T2I models, we report baseline results using inference-time fingerpriting techniques and classifier-based approaches.

\subsubsection{Inference-Time Fingerprints.}
In the context of image deanonymization, prior works on image fingerprinting are natural baseline candidates. However, most  fingerprinting techniques require access to the training pipeline (\eg to embed fingerprints) or rely on knowledge of the training data distribution, and thus do not directly apply to our threat model.
Moreover, these techniques were originally developed for early low-resolution GAN models and do not generalize well to modern text-to-image (T2I) systems. Among the remaining applicable methods, we evaluate two representative inference-time fingerprinting approaches. \citet{marra2019gans} compute a per-model fingerprint by averaging noise residuals (the difference between the original and denoised images) over multiple generations, then attribute a target image by correlating its residual with each model’s fingerprint. \citet{dzanic2020fourier} propose a frequency-domain method that models each image’s high-frequency Fourier decay using simple power-law parameters to capture systematic spectral artifacts left by generative networks. Using the 280 evaluation prompts, we compute fingerprints for each model’s generations and assign each test image to the most correlated fingerprint.

\subsubsection{Training-Based Classification.}
As an additional baseline, we train supervised classifiers to identify the generating model from labeled generations. We consider three architectures: 
(1)~an image-only classifier, 
(2)~a classifier trained on pre-extracted image embeddings, and 
(3)~a multimodal classifier combining image and text embeddings.

\paragraph{Dataset.}
To train these classifiers, we collect over two million user-submitted prompts from Midjourney’s public Discord server. To ensure diversity across topics and visual concepts, we embed all prompts and cluster them into approximately 6{,}000 groups. We then sample 2{,}000 clusters and, for each, select 22 prompts—one per model—to generate images using all 22 T2I models. This results in 44{,}000 prompt–image pairs (2{,}000 per model) covering diverse semantic domains. For training and validation, we perform cluster-based splitting: 10\% of clusters farthest from the training clusters are reserved for validation to encourage generalization across unseen concepts.

\begin{figure*}[t]
    \centering
    \includegraphics[width=\linewidth]{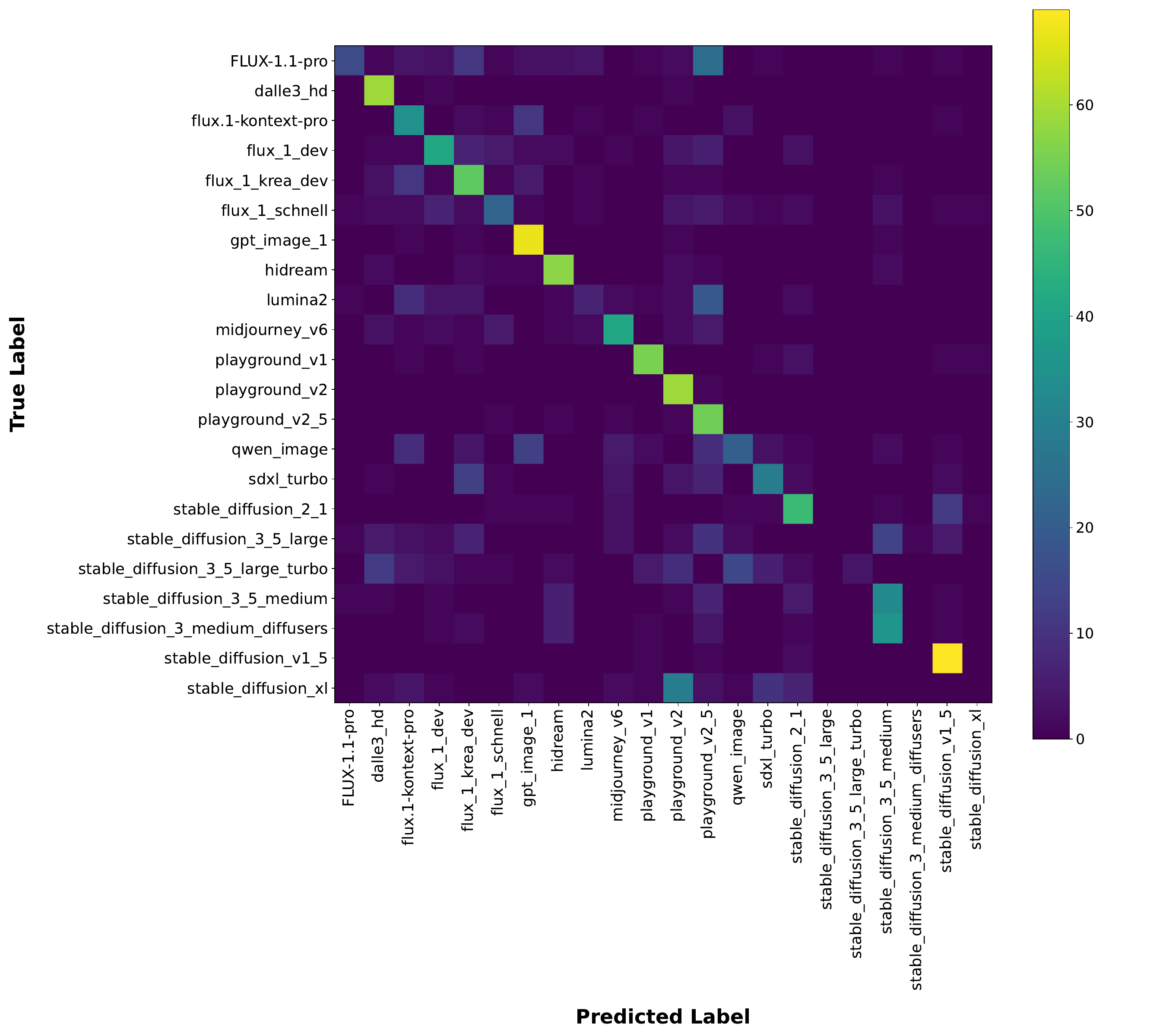}
    \caption{Confusion matrix of the image-only classifier aggregated over five runs, showing per-model prediction patterns and overall variability across models.}
\label{fig:confusion_matrix}
\end{figure*}

\paragraph{Image-Only Classifier.}
We train a ResNet-50 \cite{he2016deep} model from scratch on the generated images. While the model achieves high validation accuracy on the training dataset, its performance drops notably on evaluation prompts, as shown in \Cref{tab:attribution-results}. The confusion matrix (\Cref{fig:confusion_matrix}, Appendix) further reveals strong confusion between models of similar architecture or training source, suggesting that pixel-level features alone do not capture model-specific signatures effectively.

\paragraph{Embedding-based Classifiers.}
We train a two-layer MLP (with $[512, 128]$ hidden dimensions) on top of image-embeddings extracted using a pre-trained image encoder. We use the same MLP architecture for the image and text embedding model; the embeddings are concatenated and provided as input to the model. We experimented with various MLP configurations and hyper-parameters, and report results for the best-performing configuration. 

\subsection{Results} 
Traditional fingerprinting and classification-based approaches fail to generalize to unseen prompts, yielding low identification accuracy on the leaderboard dataset (\Cref{tab:attribution-results}).
In contrast, our inference-time, training-free clustering method achieves substantially higher performance. Using 30 generations per $\langle$p,\,model$\rangle$ pair, our method attains a top-1 accuracy of 91\%, outperforming baselines.

This setting represents the most general and realistic attack scenario. The adversary has no control over the leaderboard’s prompt selection, no access to its model-assignment mechanism, and only black-box query access to the participating models---an assumption that naturally holds since these models must be publicly accessible to appear on a leaderboard. Despite these minimal capabilities, the adversary can deanonymize leaderboard generations with striking accuracy, underscoring the severity of this vulnerability.

\subsection{One-vs-Rest Evaluation}

In this setting, the adversary targets a single model and seeks to determine whether a given leaderboard image originates from that model. We evaluate this scenario by fixing one model at a time as the target and repeating the experiment across the 280 evaluation prompts. Each experiment is repeated over five randomized leaderboard assignments to account for prompt variability. When the adversary has query access to all model APIs (Case~1), the average top-1 accuracy reaches $99.16 \pm 0.21\%$, indicating near-perfect identification. Detailed per-model results are provided in \Cref{tab:mean_accuracy_per_model} (Appendix).

\paragraph{Without Access to Other APIs.}
To test the attack under more challenging conditions, we consider the most restrictive setting (Case~2) where the adversary can query only its target model. For each model, we estimate a similarity threshold based on in-cluster distances (\Cref{sec:case2}) and report results averaged across multiple thresholds. \Cref{tab:one_vs_rest_results} summarizes the overall performance in terms of accuracy, AUC, FPR, FNR, and TPR at 1\% and 5\% FPR. Even under this limited-access scenario, the attack maintains strong detection performance, highlighting the persistence of model-specific visual signatures. Full per-model breakdowns are shown in \Cref{tab:one-vs-rest-no-access} (Appendix).

\begin{table}[t]
\centering
\scriptsize
\caption{Performance of the distance-threshold classifier using $\alpha$-quantile in-cluster distances, reporting accuracy, AUC, FNR, FPR, and TPR at 1\% and 5\% FPR.}

\begin{tabular}{c c c c c c c}
\toprule
\multirow{2}{*}{\textbf{$\alpha$}} 
& \multirow{2}{*}{\textbf{Accuracy}}
& \multirow{2}{*}{\textbf{AUC}}
& \multirow{2}{*}{\textbf{FNR}}
& \multirow{2}{*}{\textbf{FPR}}
& \multicolumn{2}{c}{\textbf{TPR}} \\
& & & & & \textbf{@1\%FPR} & \textbf{@5\%FPR} \\
\midrule
0.80 & 0.926 & 0.928 & 0.350 & 0.060 & 0.374 & 0.677 \\
0.85 & 0.910 & 0.925 & 0.291 & 0.080 & 0.368 & 0.655 \\
0.90 & 0.885 & 0.922 & 0.224 & 0.110 & 0.330 & 0.634 \\
0.95 & 0.841 & 0.916 & 0.155 & 0.160 & 0.289 & 0.584 \\
\bottomrule
\end{tabular}
\label{tab:one_vs_rest_results}
\end{table}

\begin{figure*}[ht]
    \centering
    \includegraphics[width=1.0\linewidth]{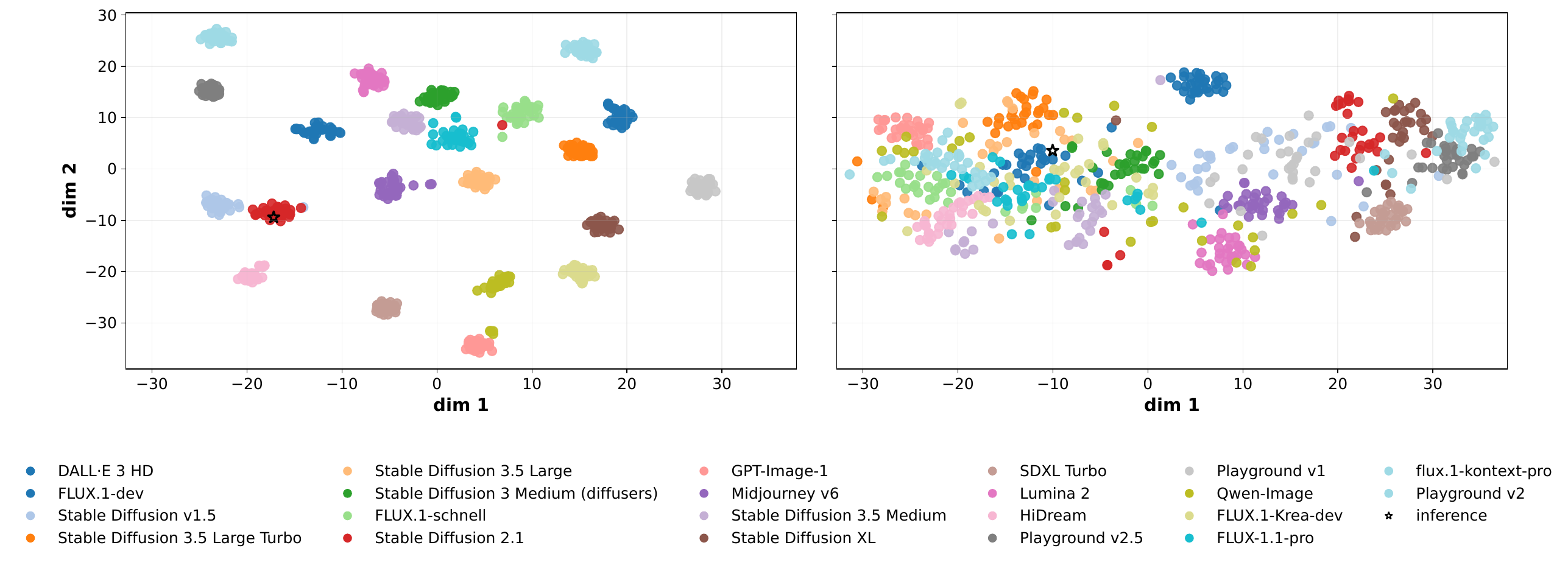}
    \caption{Embeddings for two prompts with high (left) and low (right) distinguishability, showing clear separation versus overlap among model generations.}

    \label{fig:dist_score_distribution}
\end{figure*}

\subsection{Distinguishability Score Analysis}

\begin{figure}%
    \centering
    \includegraphics[width=\linewidth]{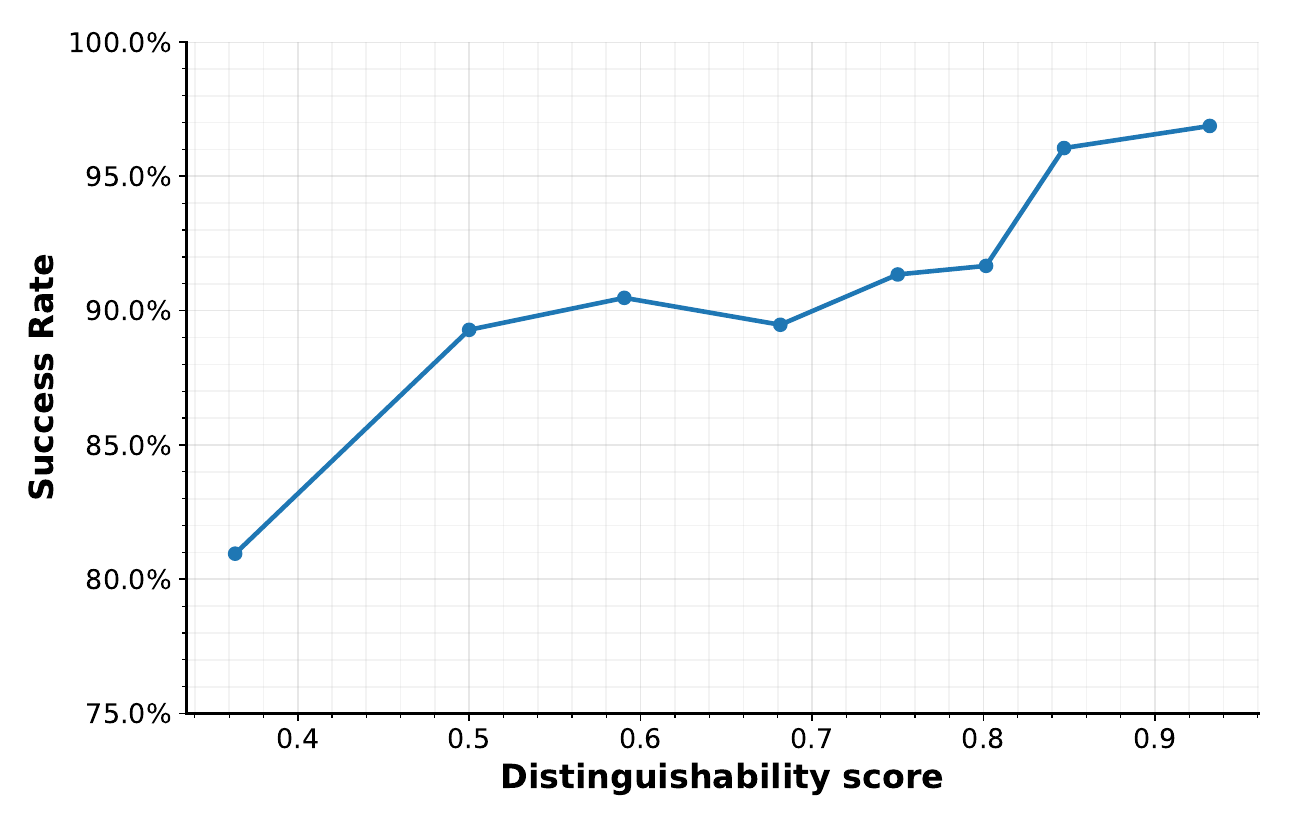}
    \caption{
Deanonymization success rate across bins of distinguishability scores. 
Higher scores correspond to greater success, confirming that the metric predicts attack performance.
}
\label{fig:success_vs_dist}
\end{figure}

In \Cref{sec:distinguishability}, we introduced a metric to quantify how separable different models' generations are for a given prompt in the embedding space. This metric provides insight into which prompts make voting-based leaderboards more vulnerable to deanonymization. As shown in \Cref{fig:success_vs_dist}, deanonymization success increases consistently with higher distinguishability scores, confirming that prompts yielding more separable model clusters lead to higher identification accuracy. \Cref{fig:dist_score_distribution} further illustrates qualitative examples with low and high distinguishability.

\begin{figure*}[t]
    \centering
    \includegraphics[width=\linewidth]{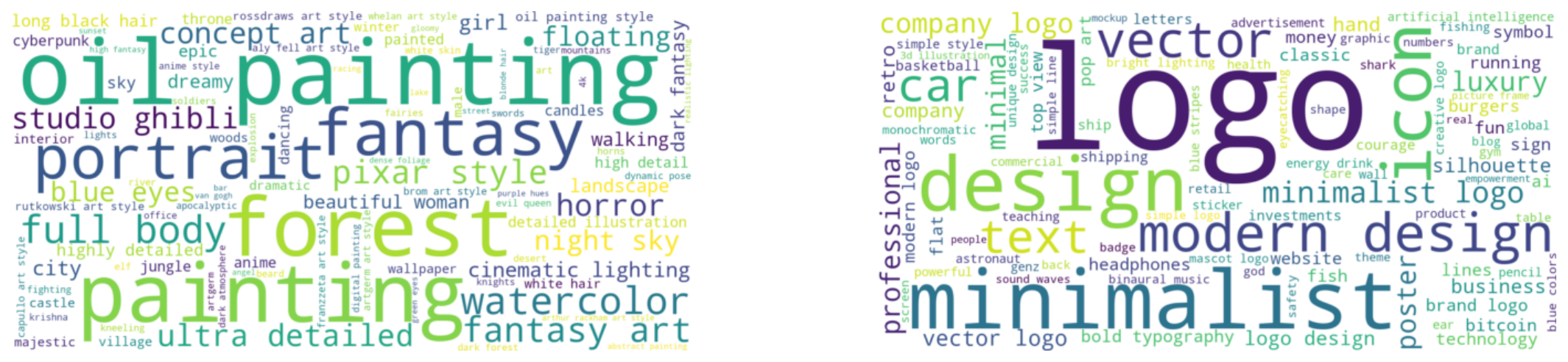}
    \caption{Word-cloud for keywords corresponding to \textbf{Left:} prompts with high distinguishability scores and \textbf{Right:} low distinguishability.
}\label{fig:frequency_distinguishability}
\end{figure*}

\begin{figure*}[t]
\centering
\renewcommand{\arraystretch}{1}
\setlength{\tabcolsep}{1pt}

\begin{minipage}{0.95\linewidth}
\centering
{\small \textbf{Prompt:} ``Design a LOGO of instant tea with white background.''}
\end{minipage}

{\tiny
\resizebox{0.95\linewidth}{!}{%
\begin{tabular}{@{}>{\centering\arraybackslash}m{2.0cm}ccccc@{}}

\toprule
\textbf{Model} & \multicolumn{5}{c}{Generated Images (five different seeds)}\\
\midrule

\makecell[c]{Stable Diffusion 3.5 \\ Medium} &
\img{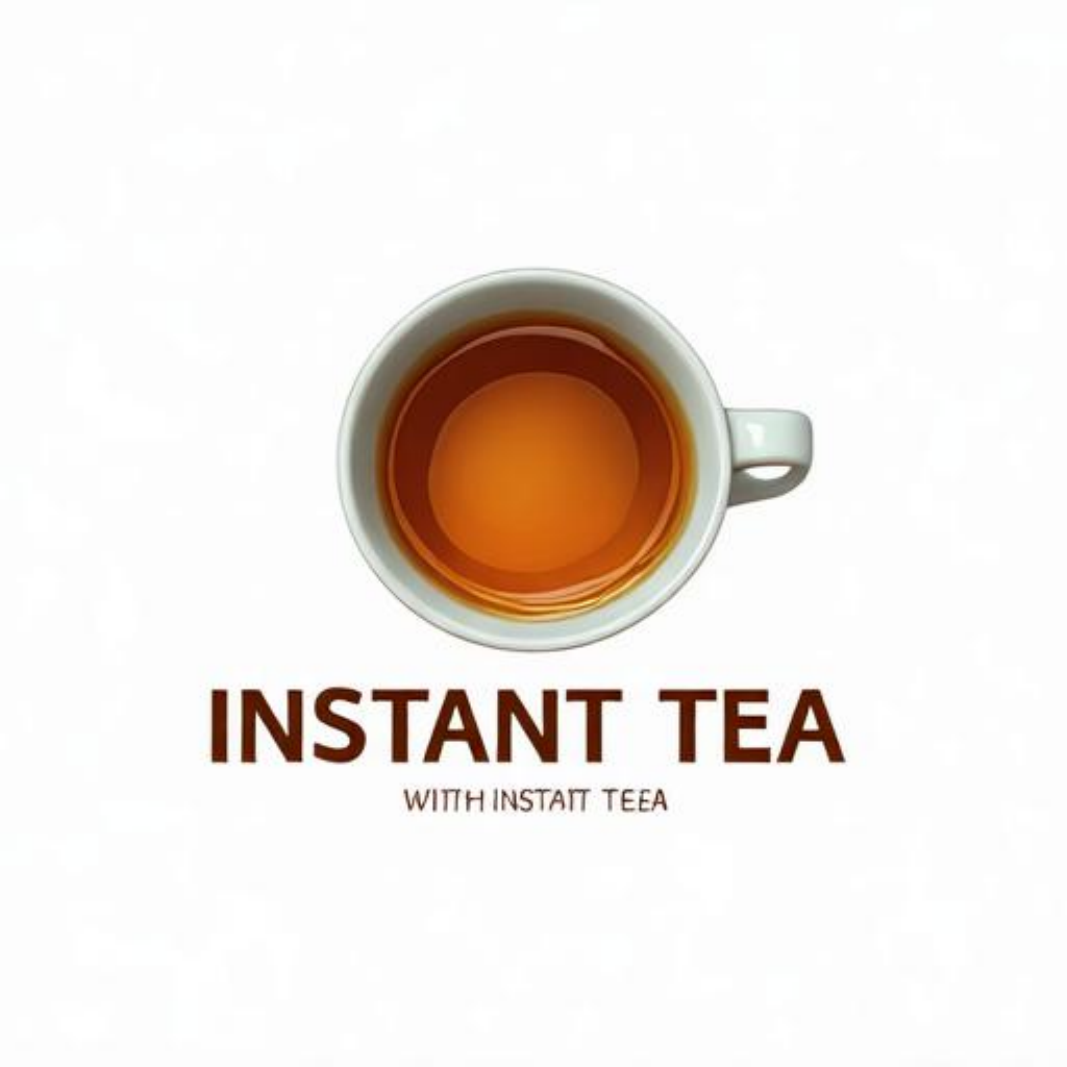} &
\img{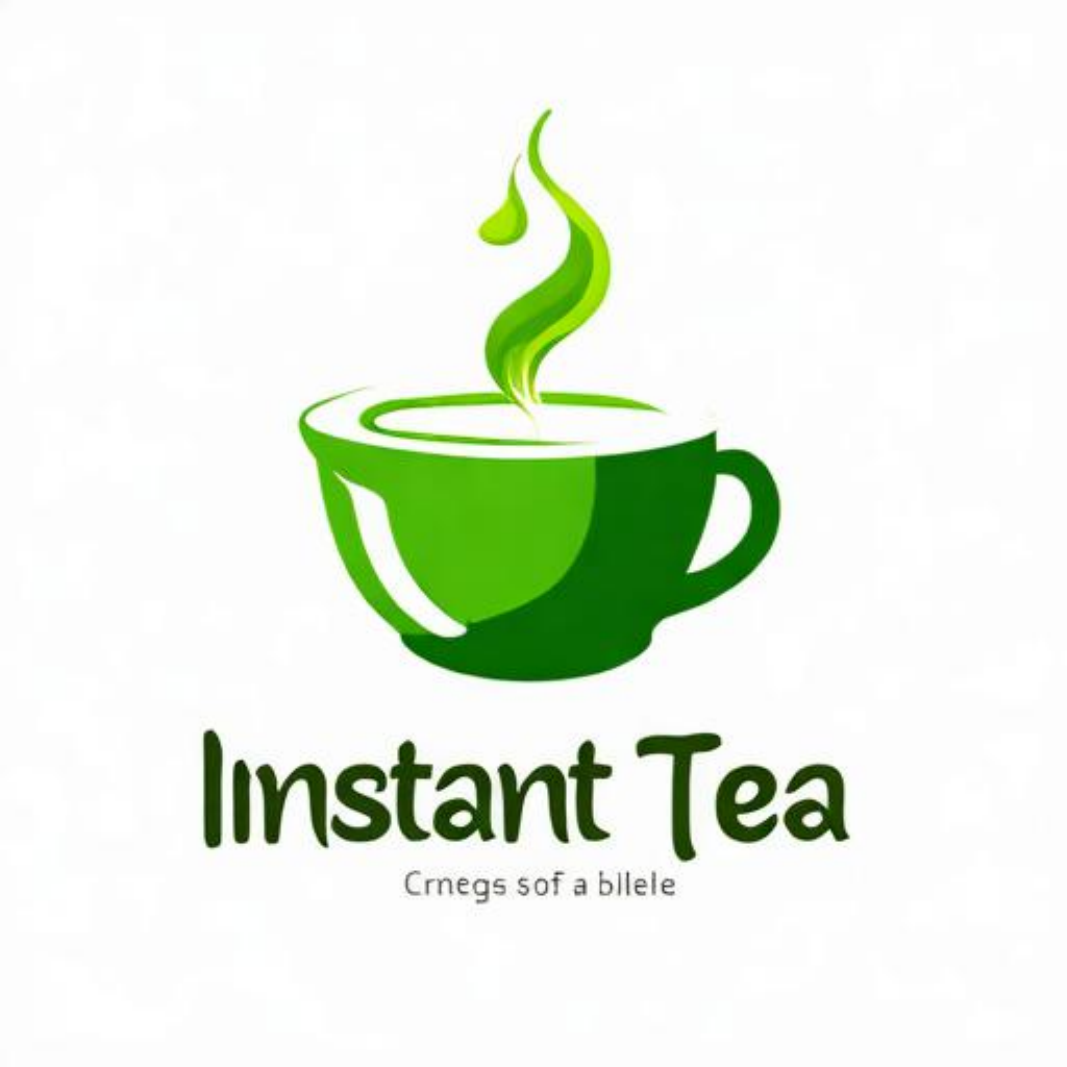} &
\img{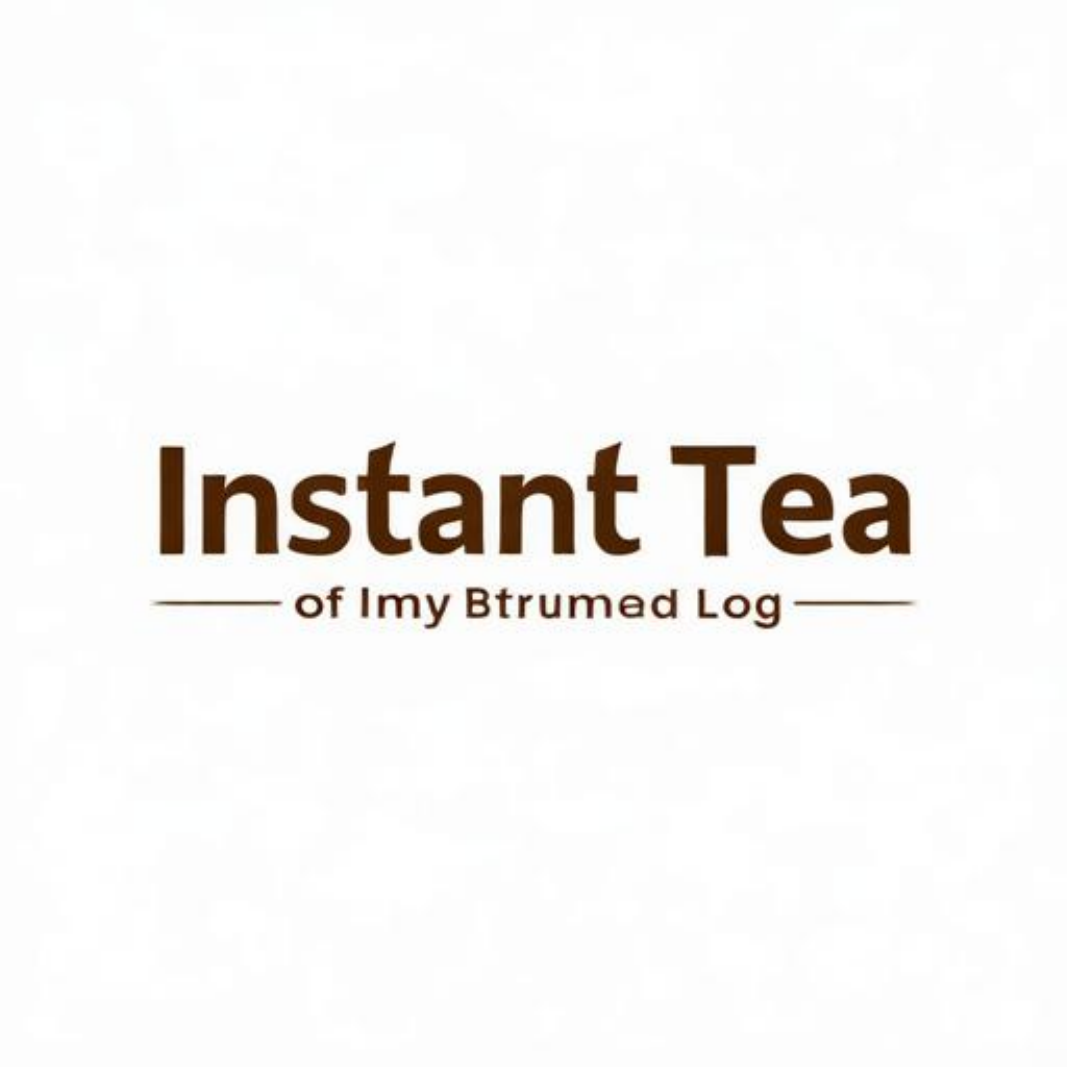} &
\img{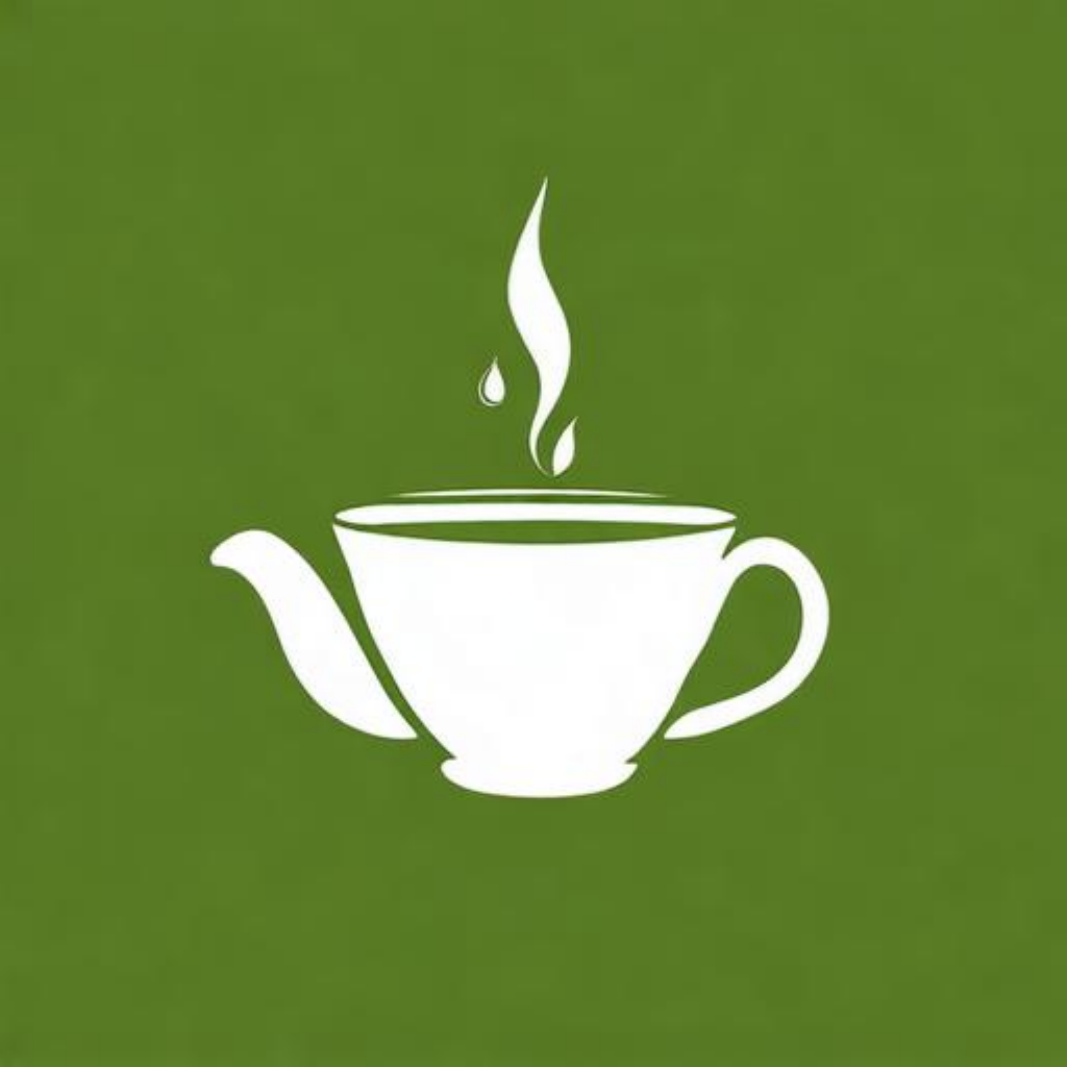} &
\img{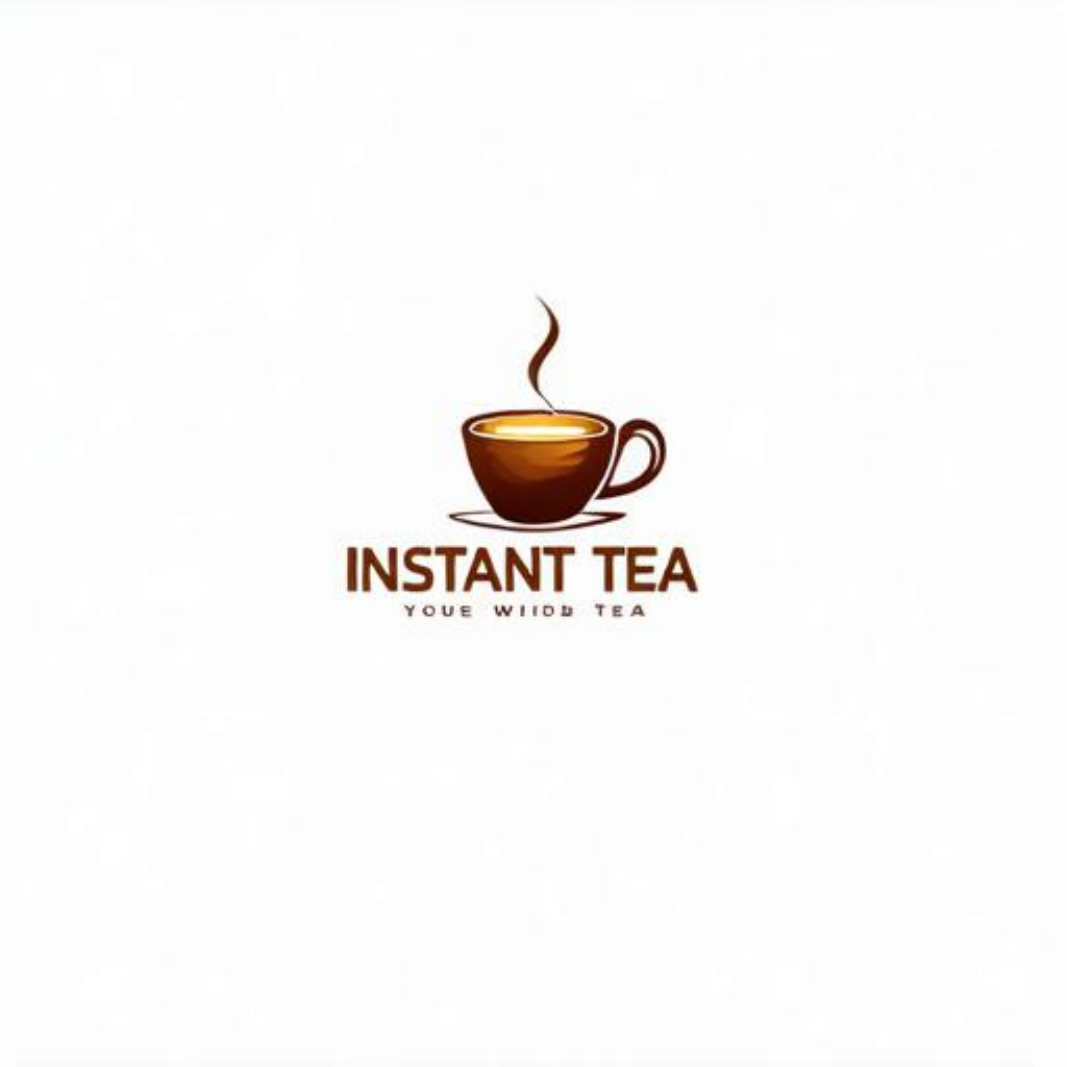} \\

\makecell[c]{Stable Diffusion 3.5 \\ Large Turbo} &
\img{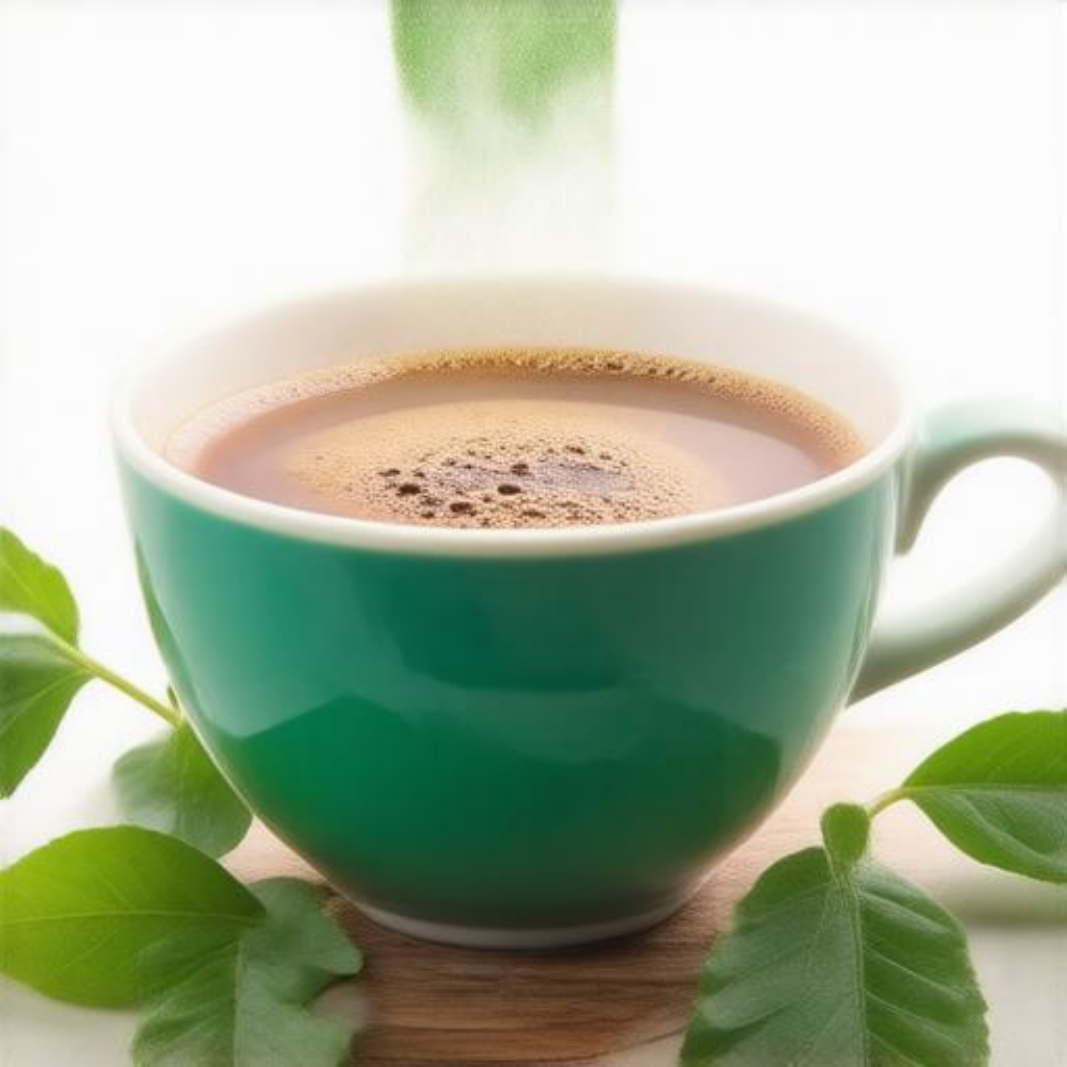} &
\img{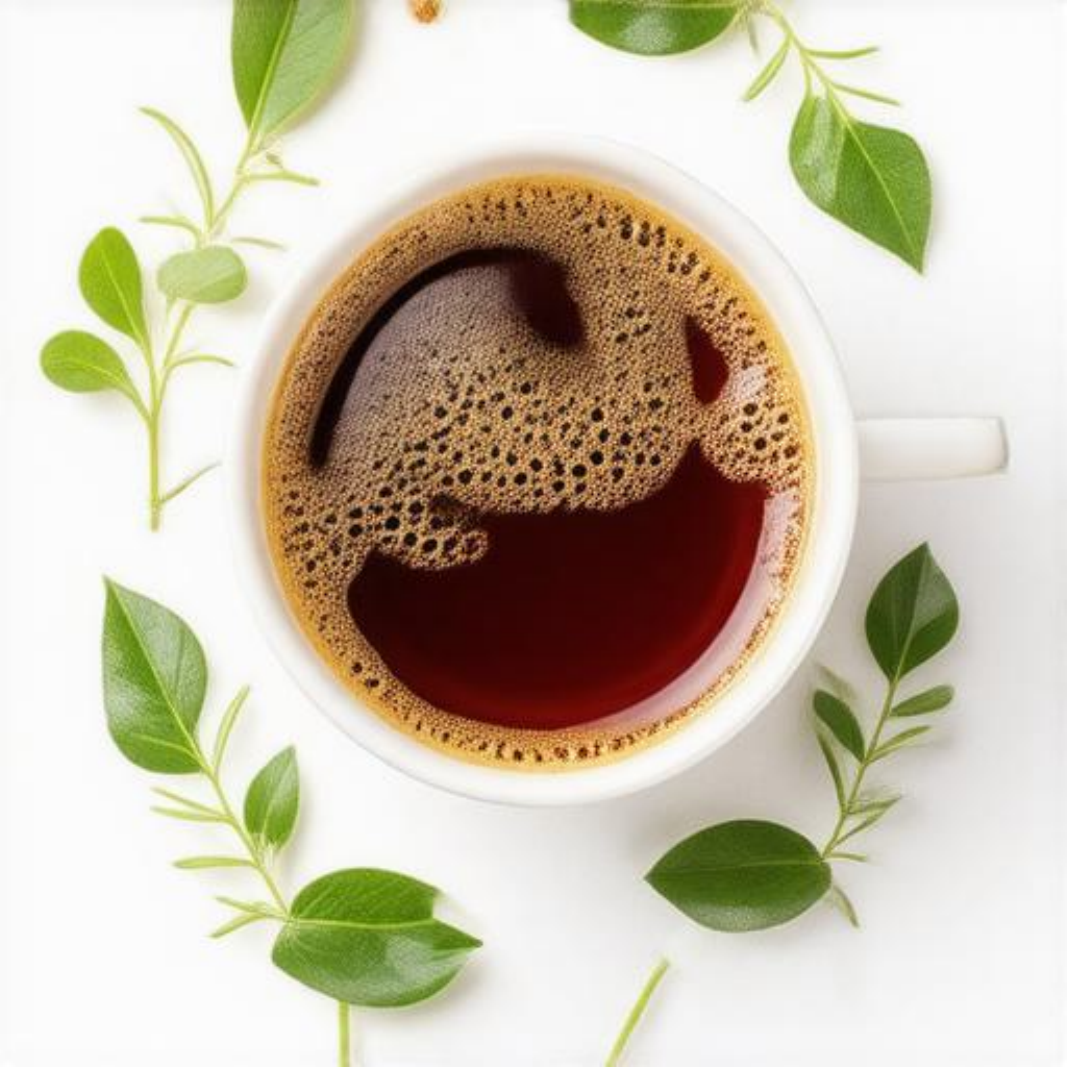} &
\img{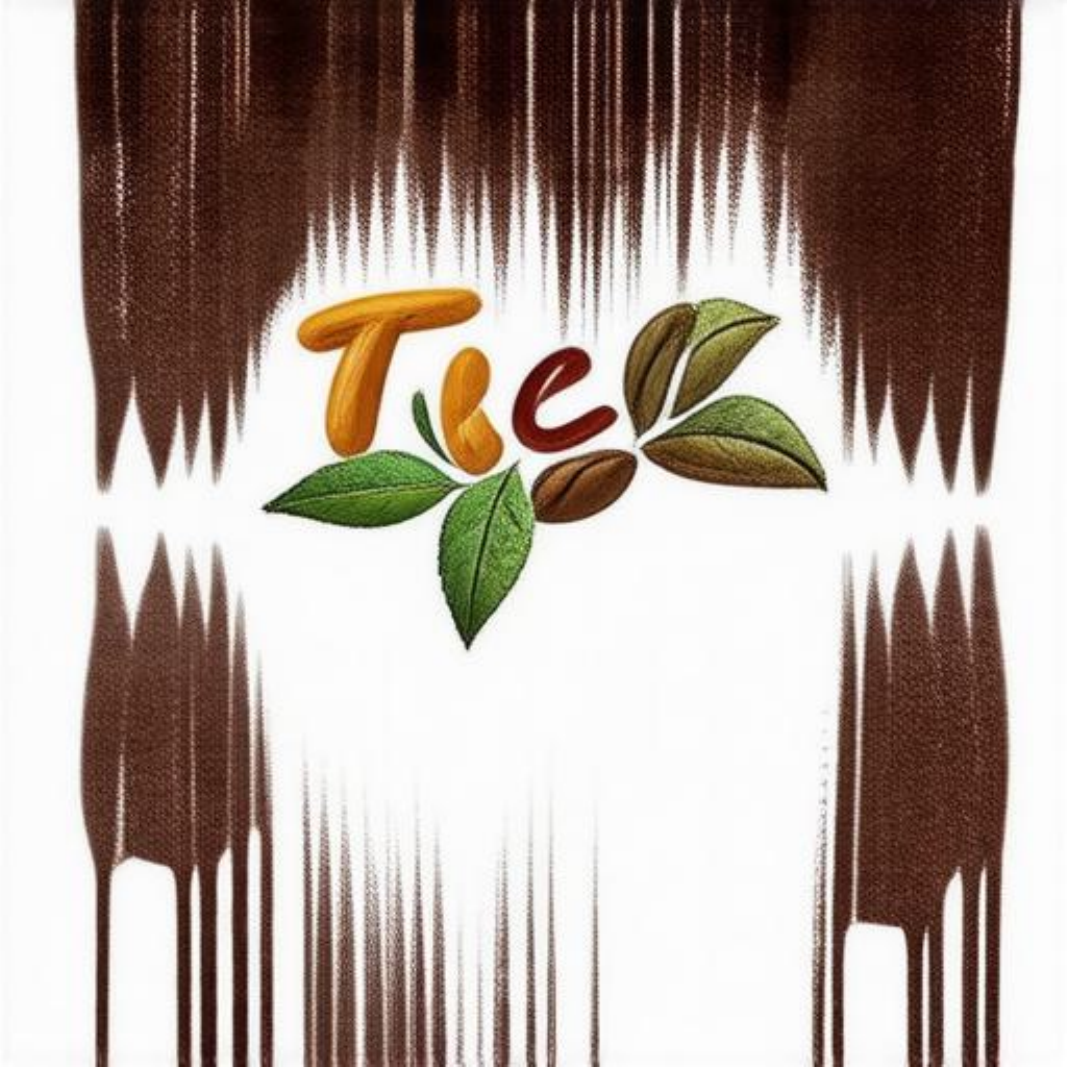} &
\img{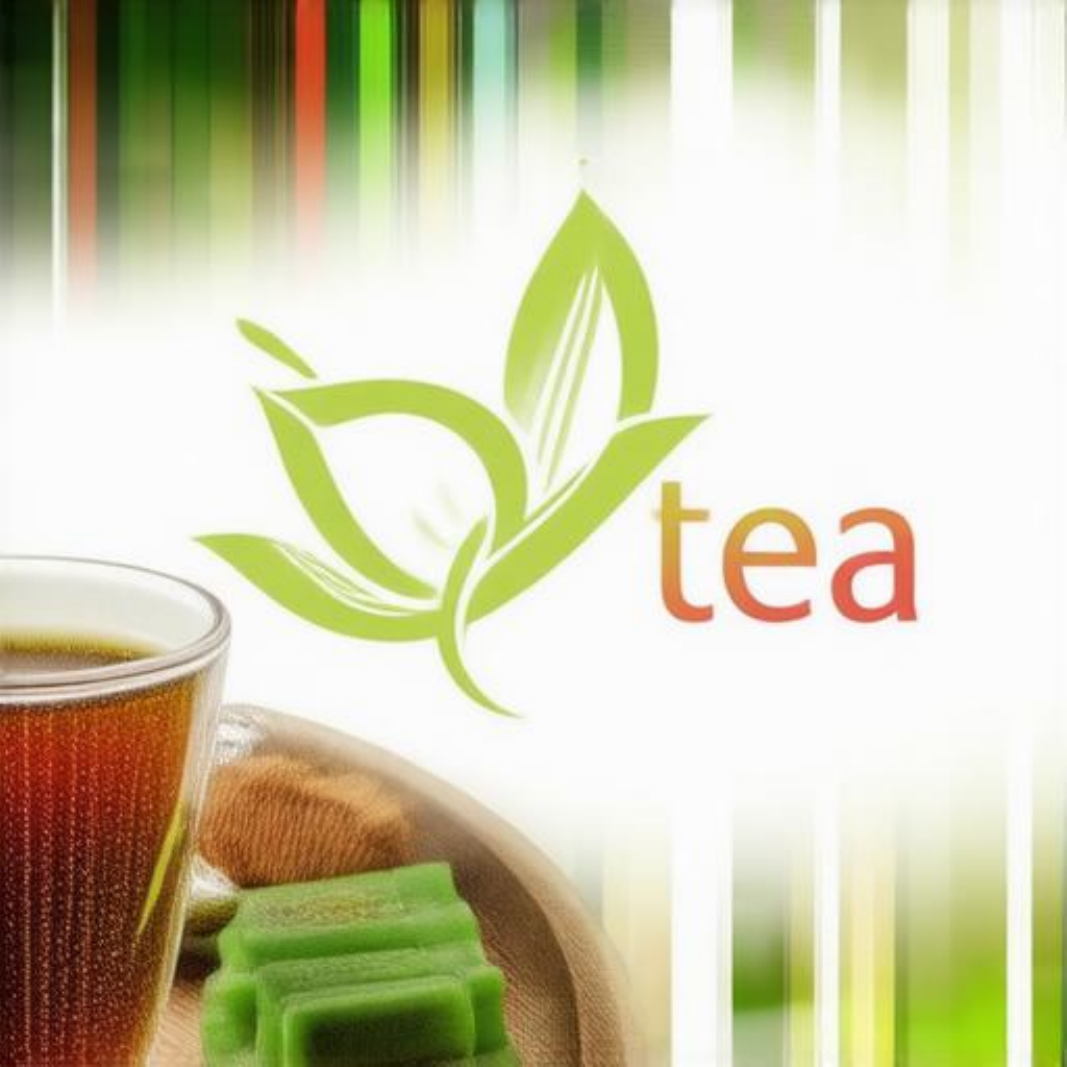} &
\img{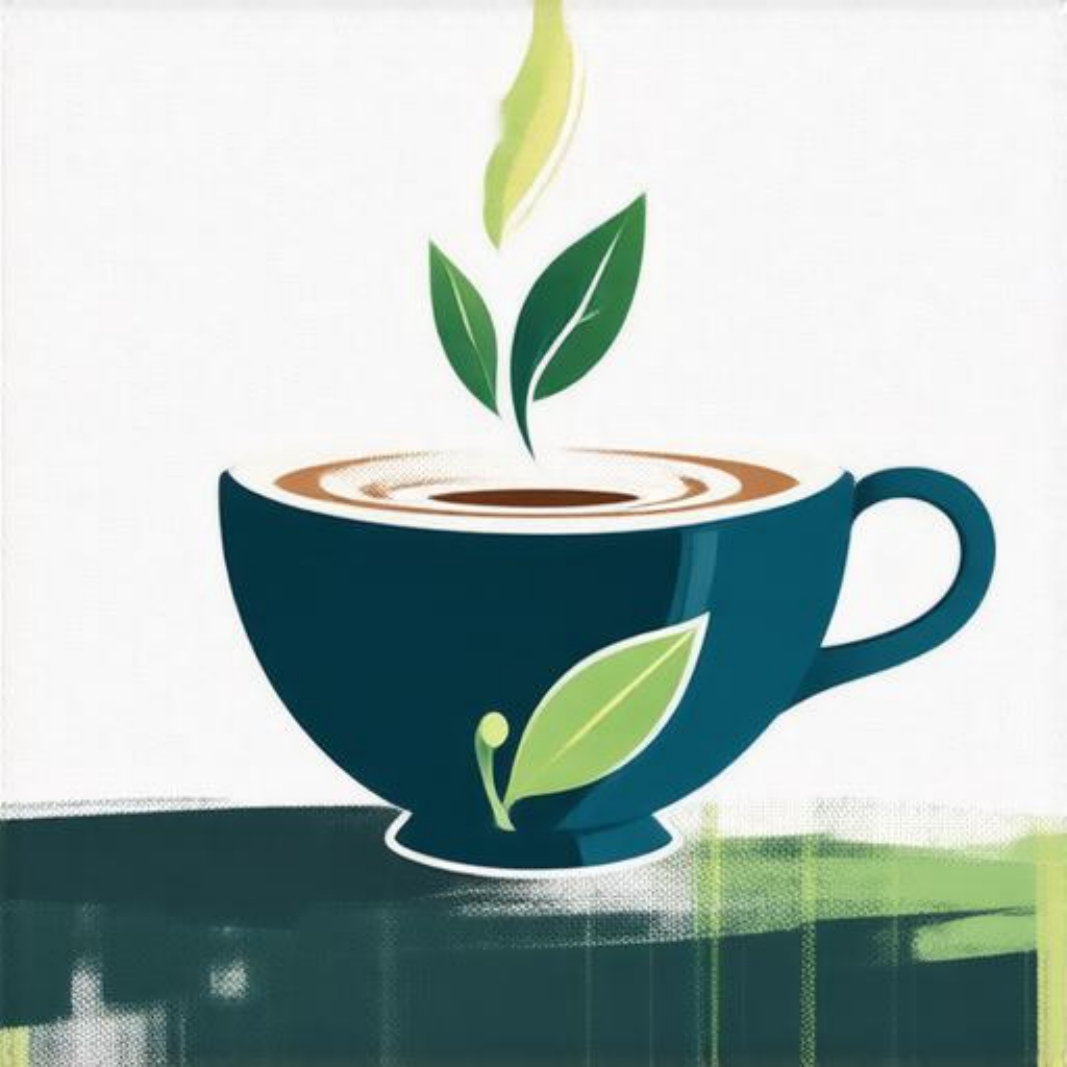} \\

\bottomrule
\end{tabular}%
} %
} %

\caption{Example of a low-distinguishability prompt with high intra-model variation, with generations from two models across five seeds.}

\label{fig:bad_prompts}
\end{figure*}

\begin{figure*}[t]
\centering
\renewcommand{\arraystretch}{1}
\setlength{\tabcolsep}{1pt}

\begin{minipage}{0.95\linewidth}
\centering
{\small \textbf{Prompt:} ``create a painting of a lake surrounded by forest in the winter, foggy weather, snow falling, andre kohn style painting.''}
\end{minipage}

{\tiny
\resizebox{0.95\linewidth}{!}{%
\begin{tabular}{@{}>{\centering\arraybackslash}m{2.0cm}ccccc@{}}

\toprule
\textbf{Model} & \multicolumn{5}{c}{Generated Images (five different seeds)}\\
\midrule

\makecell[c]{FLUX.1-dev} &
\img{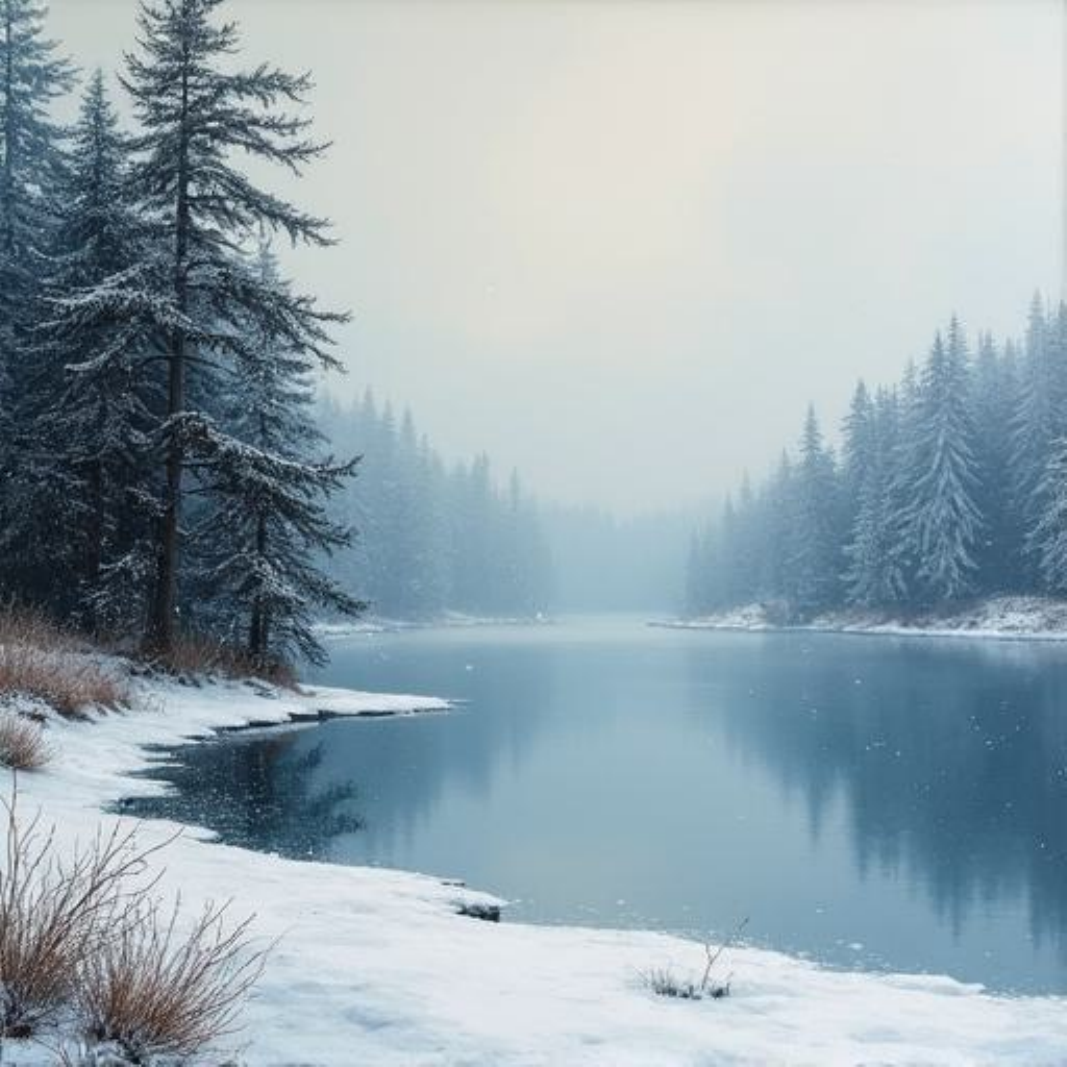} &
\img{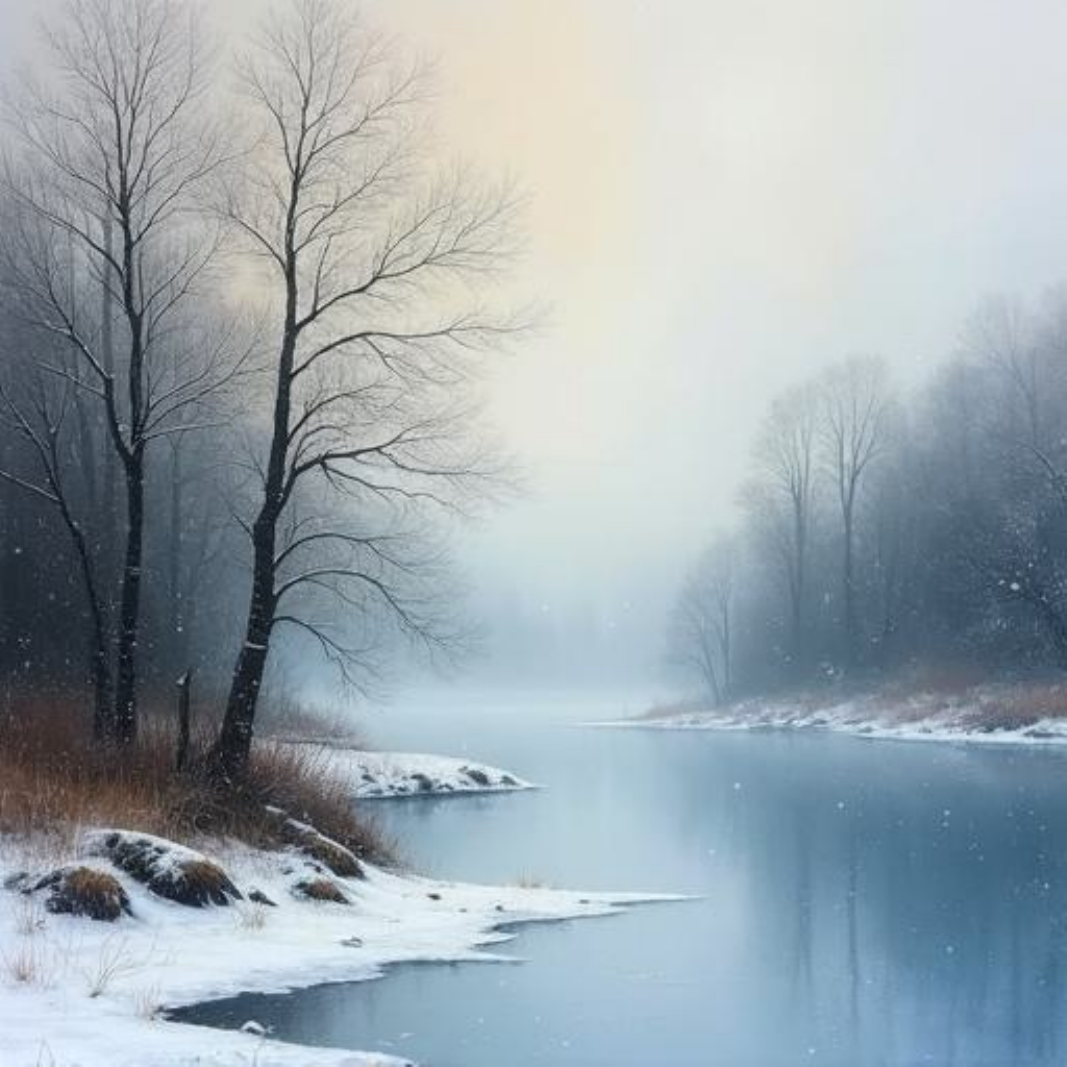} &
\img{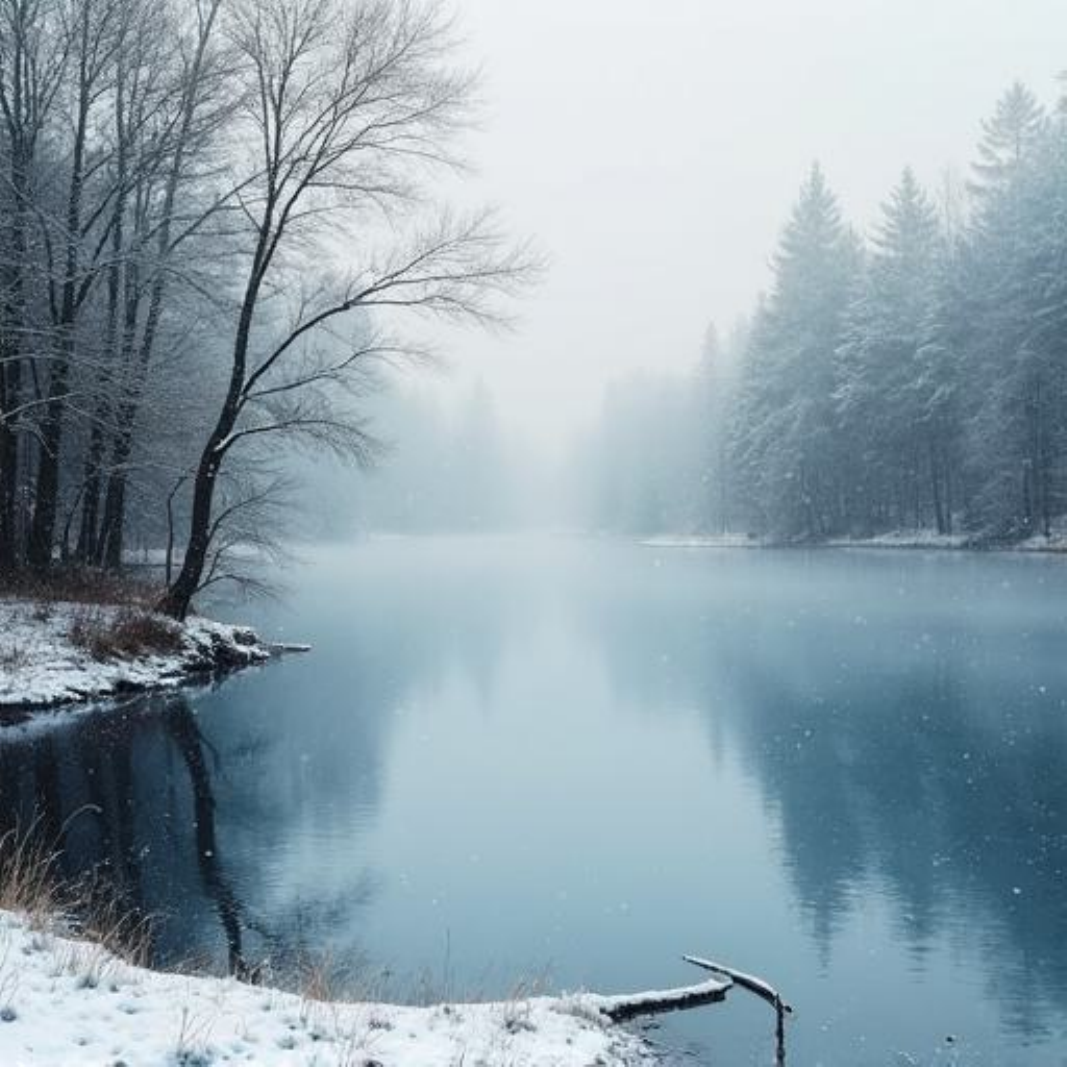} &
\img{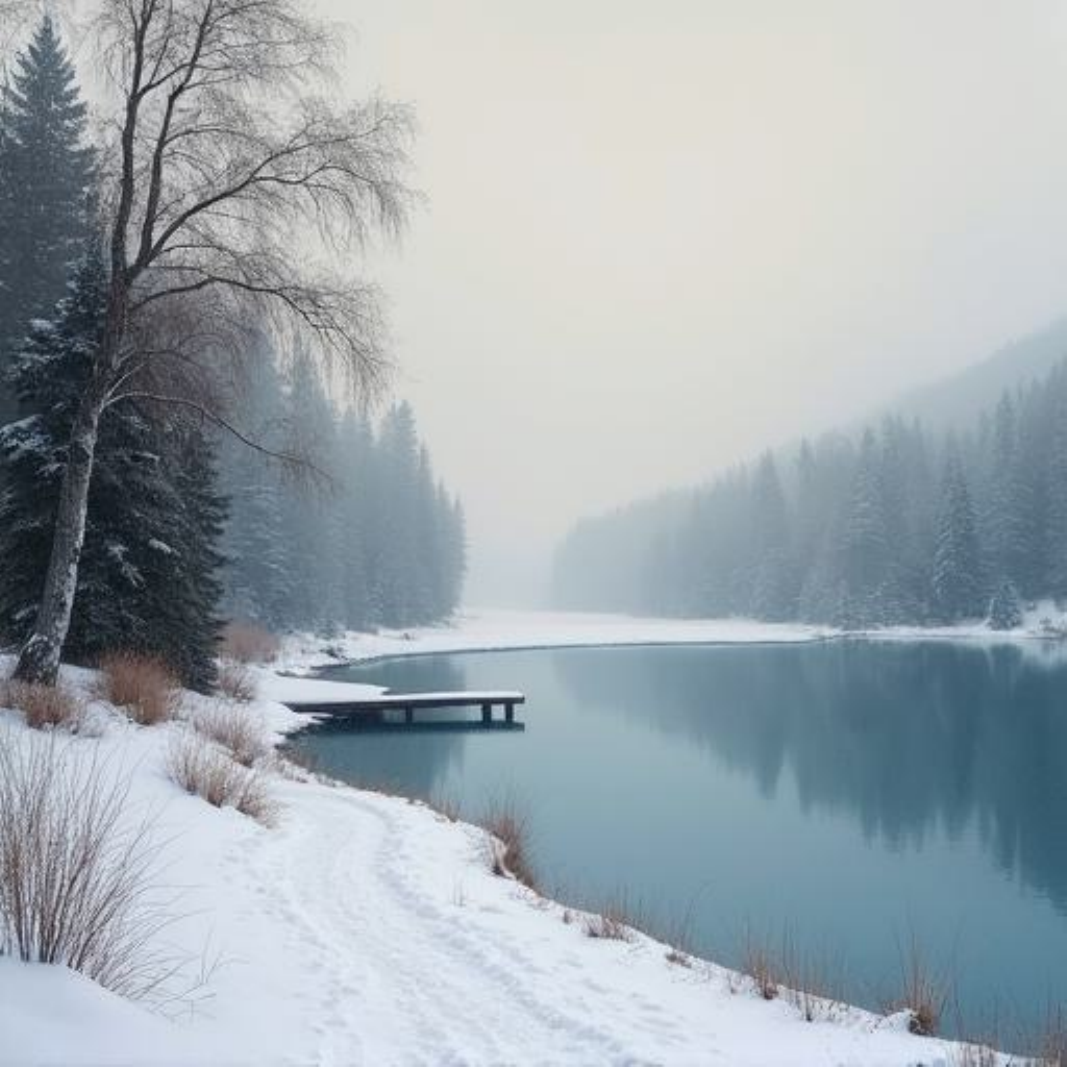} &
\img{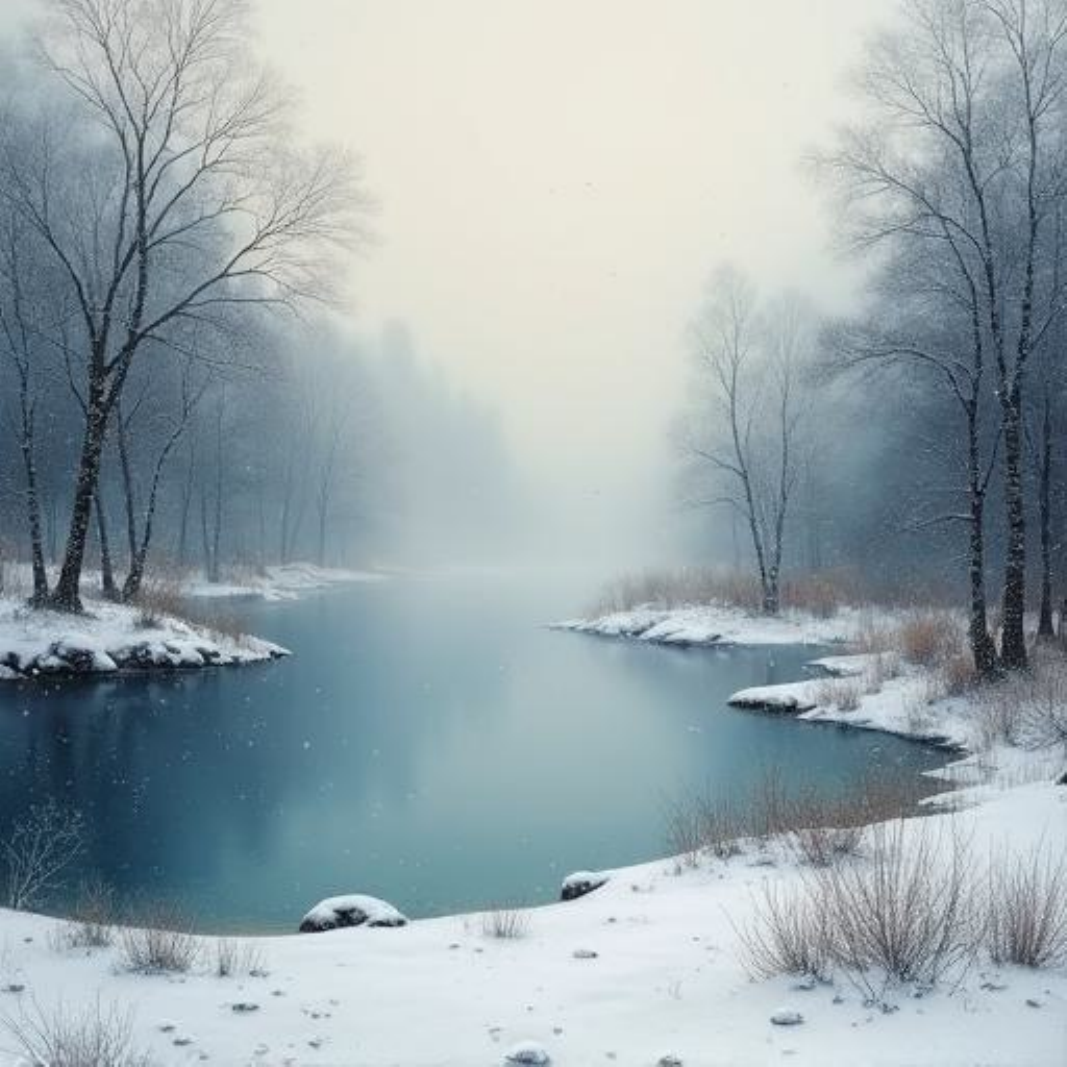} \\

\makecell[c]{SDXL Turbo} &
\img{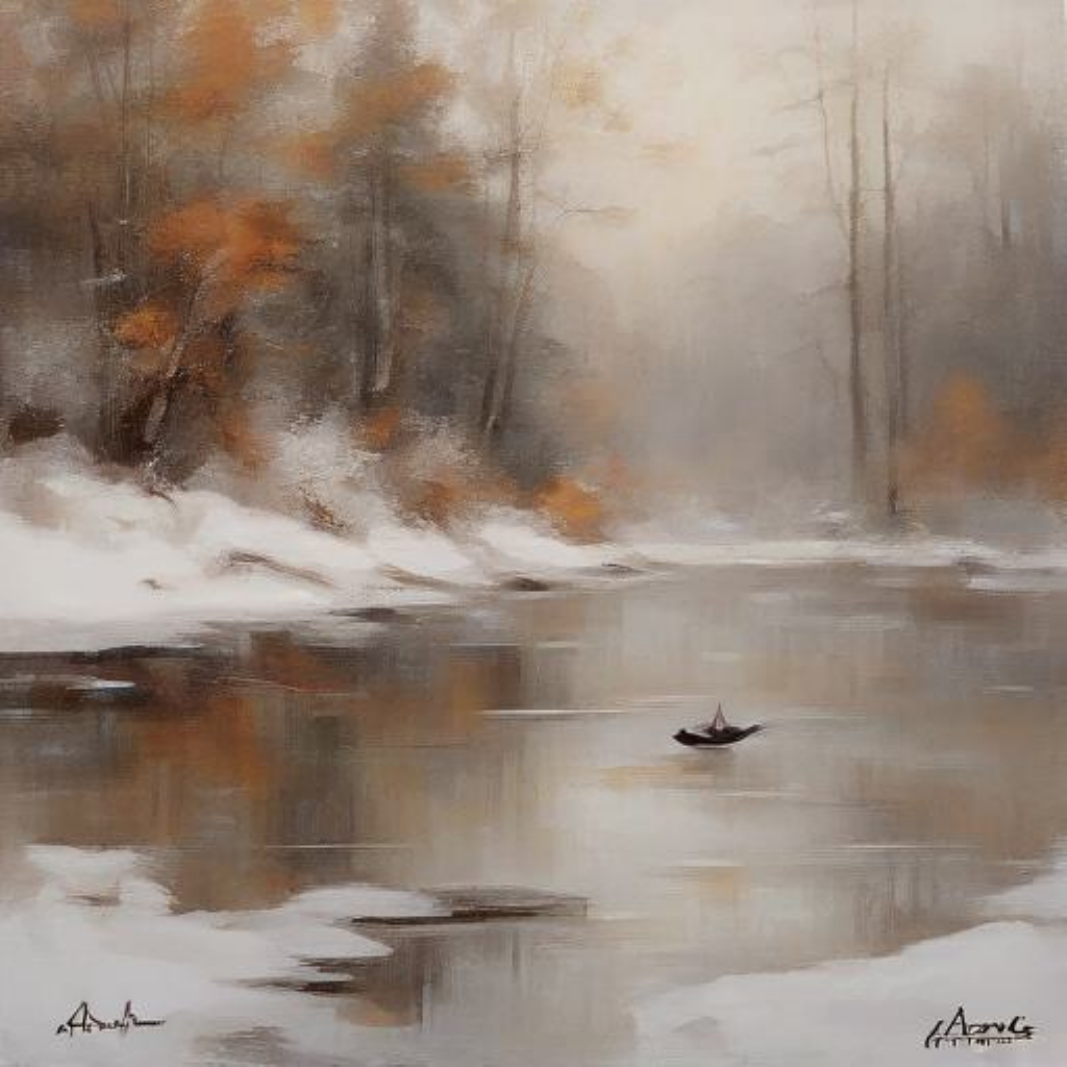} &
\img{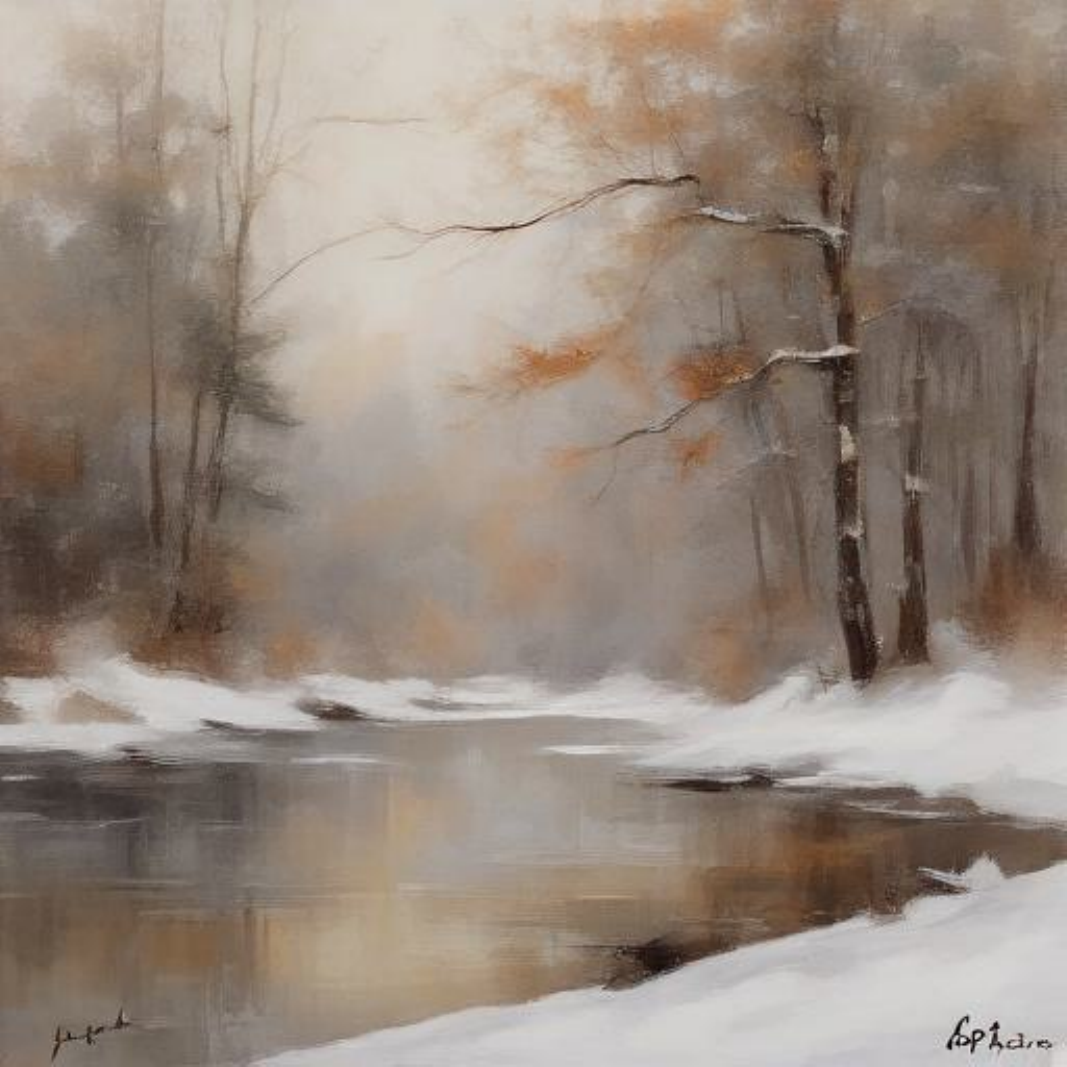} &
\img{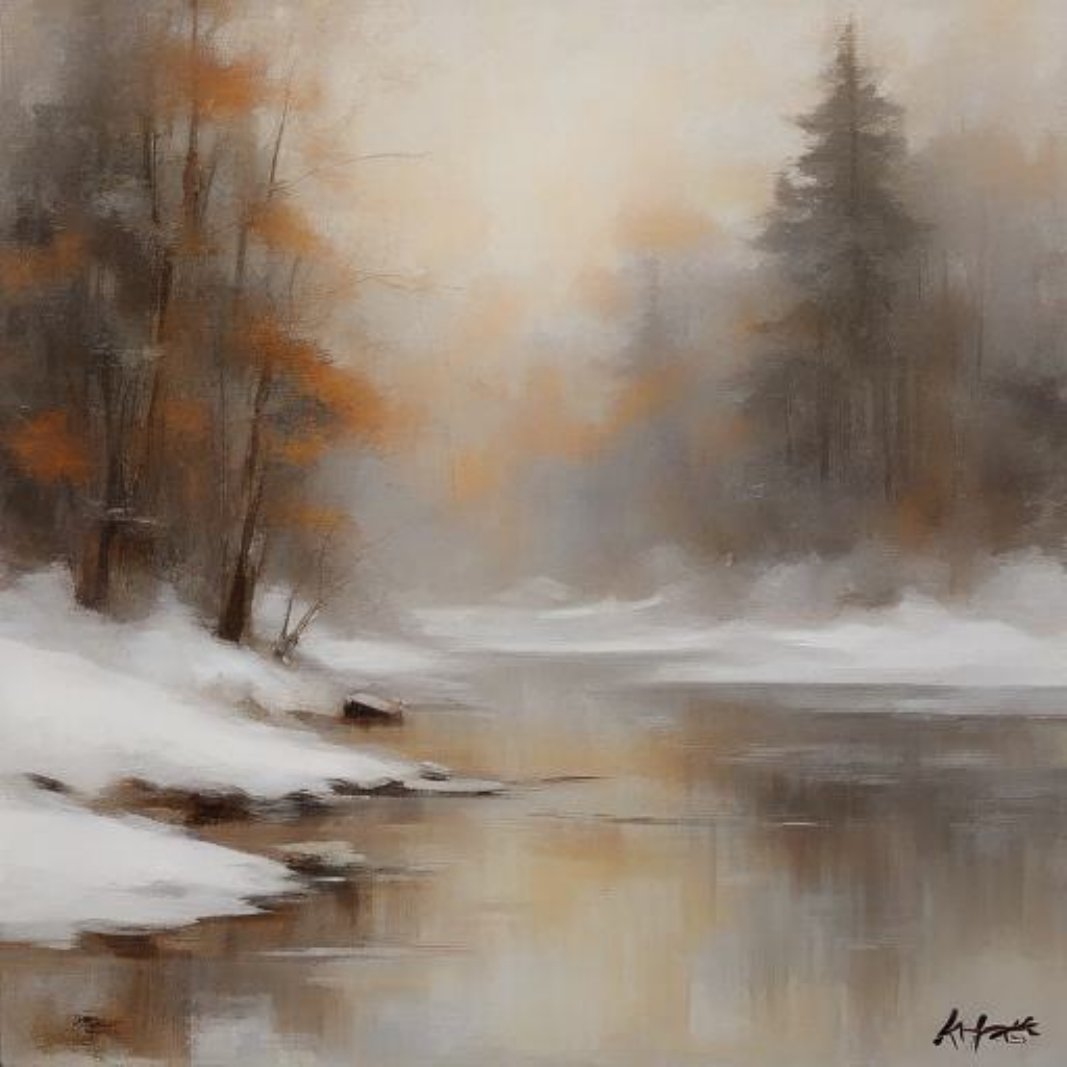} &
\img{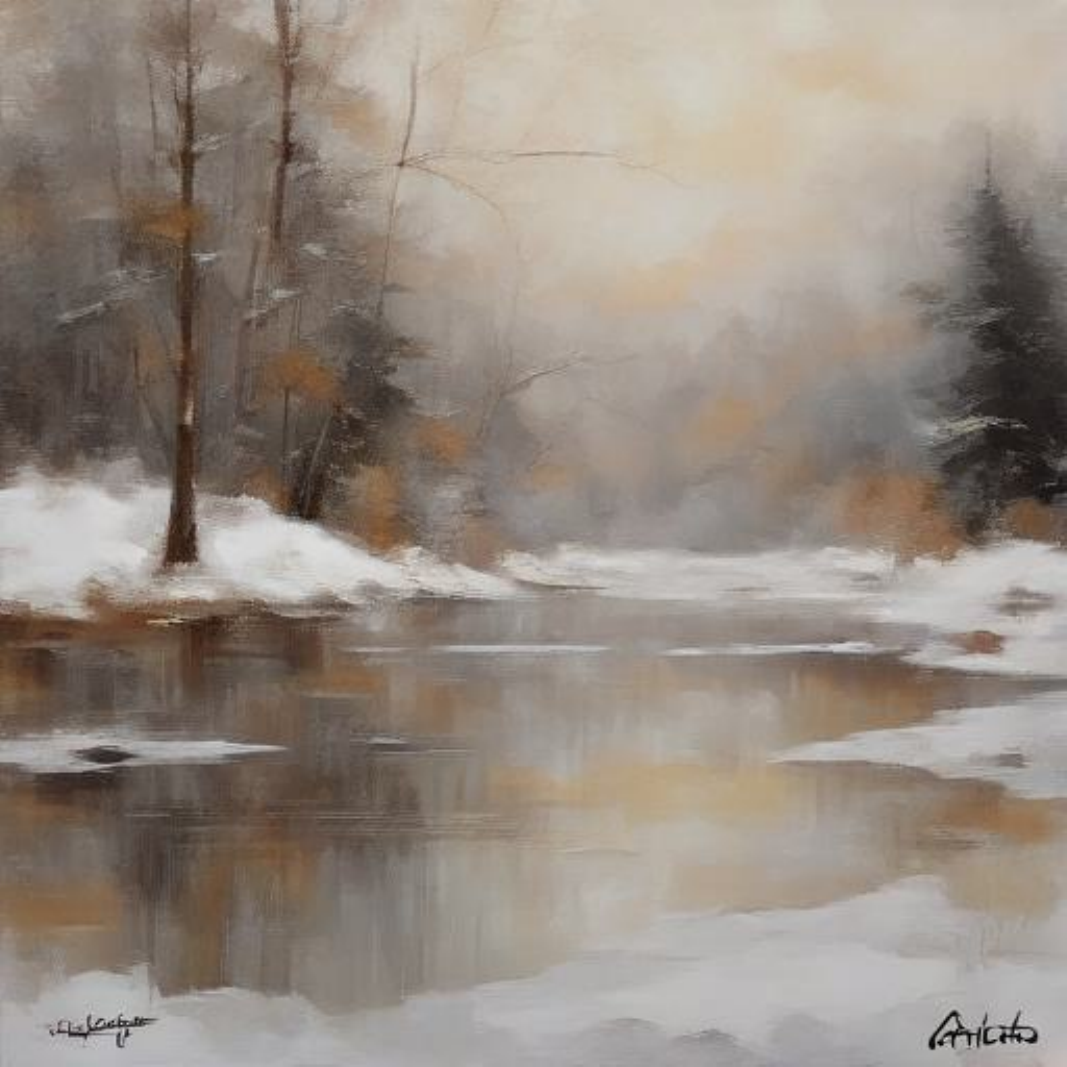} &
\img{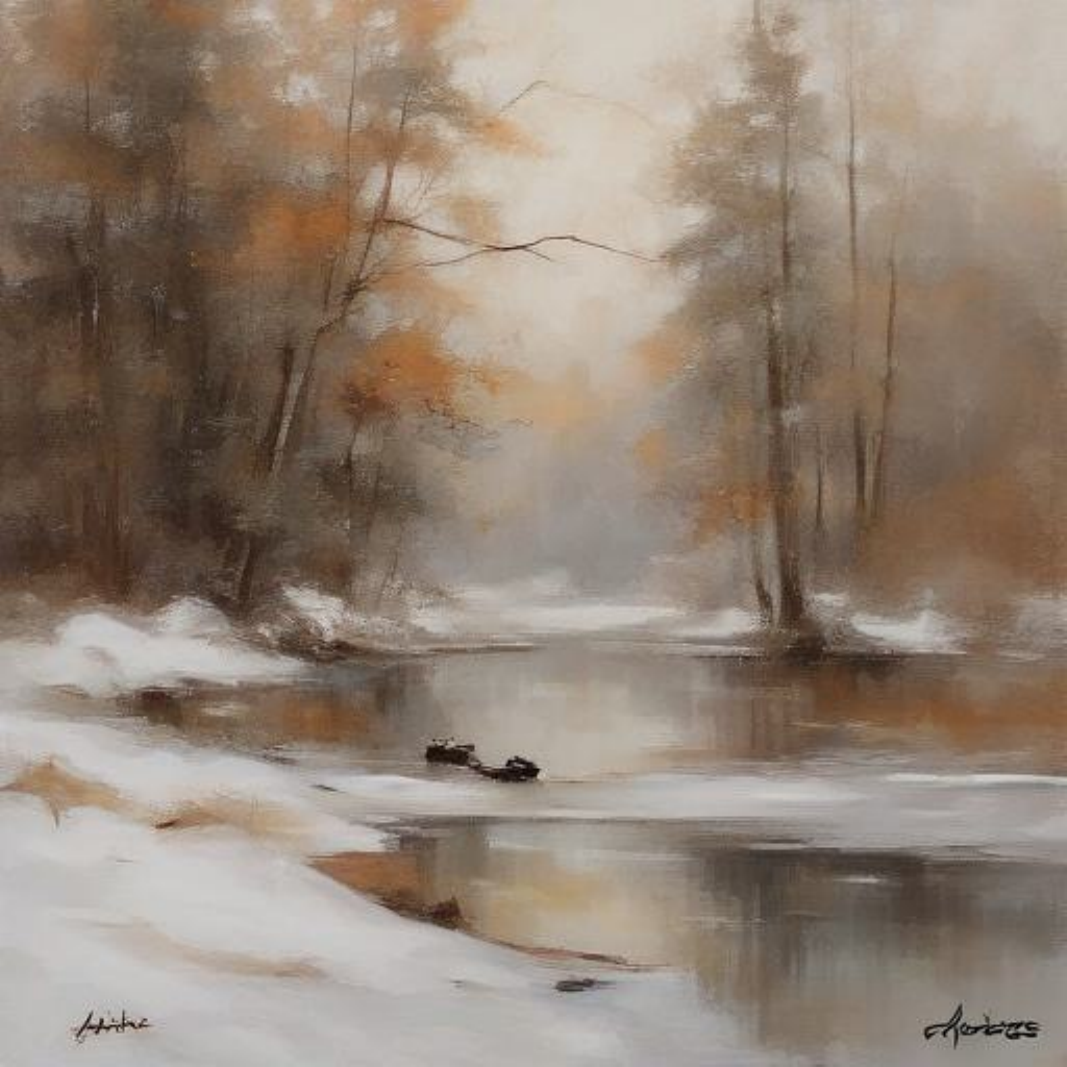} \\

\bottomrule
\end{tabular}%
} %
} %

\caption{Example of a high-distinguishability prompt with low intra-model variation, with generations from two models across five seeds.}

\label{fig:good_prompts}
\end{figure*}

Beyond the leaderboard-scale evaluation, we conduct a large-scale analysis to examine distinguishability across diverse visual concepts. Using approximately two million Midjourney prompts, we cluster them into 920 semantic groups and sample 15 prompts per cluster. Details of the processing and filtering procedure are provided in Appendix \ref{app:midjourney_prompts_analysis}. For each prompt, we generate 15 images from 11 representative models, producing over two million generated samples in total. We then compute distinguishability scores for all prompts and visualize the most and least distinguishable concepts in \Cref{fig:frequency_distinguishability}. Prompts corresponding to stylistically rich or visually constrained categories (\eg ``oil painting,’’ ``anime portrait’’) exhibit high distinguishability, whereas generic or ambiguous prompts (\eg ``city street,’’ ``landscape,’’ ``logo’’) yield lower separability. Representative examples of a high-distinguishability and a low-distinguishability prompt, along with their generations from two models, are provided in \Cref{fig:bad_prompts,fig:good_prompts} (Appendix).

\paragraph{Prompt Control and Attack Performance.}
In our main setting, the adversary has no control over leaderboard prompts. However, if prompt control is allowed, distinguishability analysis can guide prompt selection to boost attack success. For instance, sampling 30 prompts with distinguishability scores of 1 yields 100\% top-1 accuracy, showing that prompt-aware adversaries can achieve perfect deanonymization.

\subsection{Ablation Study}

We conduct ablation studies to examine how key design choices affect attack performance.

\begin{figure}[t]
    \centering
    \label{fig:deanon_accuracy_vs_k}
    \includegraphics[width=\linewidth]{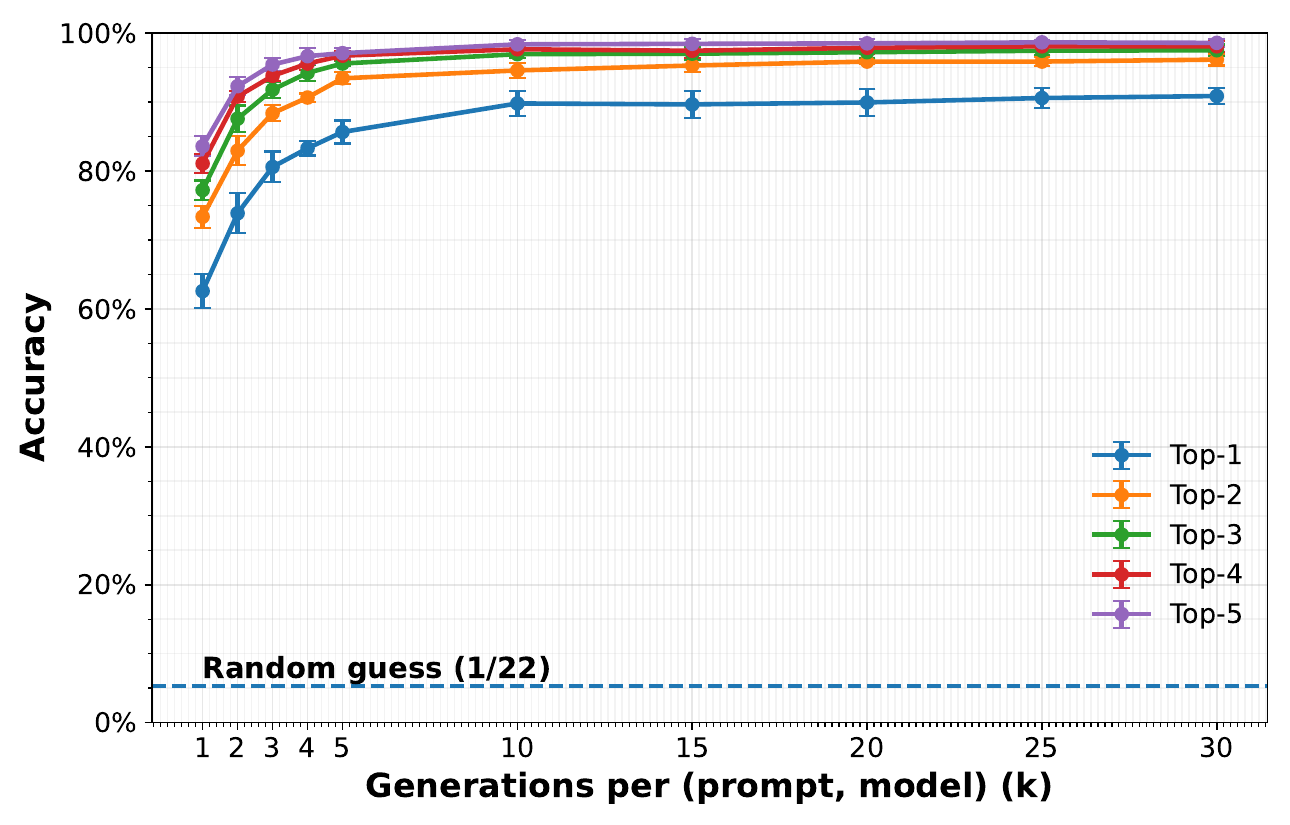}
    \caption{
Deanonymization accuracy versus number of generations $k$ per (prompt, model) pair.  
Curves show mean Top-1 through Top-5 accuracy over five runs with one-standard-deviation error bars.  
The dashed line indicates the random-guess baseline of $1/22\approx4.55\%$ accuracy.
}
\label{fig:acc_vs_dist}
\end{figure}

\paragraph{Number of Samples per Model.}
An important practical consideration for the adversary is the number of images required per model, since generating large batches can be costly, particularly for commercial APIs. \Cref{fig:acc_vs_dist} (Appendix) shows the effect of varying the number of generations $k$ per $\langle$p,\,model$\rangle$ pair on attack accuracy. With only 10 samples per model, the attack achieves nearly the same top-1 accuracy as with 30. Even a single image per model reaches about 62\% accuracy, far above random chance (1/22), showing the attack remains effective with limited generations.

\paragraph{Effect of the Image Encoder.}
We assess the impact of image encoder quality on deanonymization accuracy by testing three additional encoders. The chosen encoders include top-performing open-weight models from the MTEB leaderboard and the earlier OpenAI CLIP for comparison. As shown in \Cref{tab:image-encoder-results} (Appendix), modern encoders such as \texttt{siglip2-large} and \texttt{CLIP-ViT-bigG-14} outperform older models, underscoring the importance of high-quality embedding spaces for capturing stylistic and semantic differences between T2I models.

\section{Countermeasure}
\label{sec:defense}

Our results show that model generations form well-separated clusters in embedding space, enabling clustering-based deanonymization. To mitigate this, we propose disrupting these clusters by slightly perturbing each image so that its embedding shifts away from the source model’s region and toward those of other models. In leaderboard settings, defenders cannot alter embeddings directly and must modify the images themselves while keeping perturbations imperceptible. Inspired by recent advances in adversarial examples for vision-language models~\cite{hu2025transferable, Xie_2025_CVPR, wangTransfer, zhao2023evaluating, Lu_2023_ICCV}, we add small perturbations that (i) move the image out of its source cluster, (ii) increase overlap with other models’ embeddings, and (iii) preserve visual fidelity. The defender does not require knowledge of the attacker’s encoder, which is treated as a black box.

\begin{algorithm}[htbp]
\caption{Adversarial Post-Processing against Deanonymizition}
\label{alg:defense}
\begin{algorithmic}[1]
\State \textbf{Input:} T2I models $\mathcal{C} = \{M_1,\dots,M_n\}$ on the leaderboard; prompt $p$; visual encoder $E_v$; similarity $S(\cdot,\cdot)$; temperature $\tau$; perturbation budget $\varepsilon$; iterations $T$; learning rate $\eta$; target image $I^\ast$ generated by $M_i$ using $p$
\State \textbf{Output:} Defended image $\tilde I^\ast$
\vspace{0.3em}
\State $S_i \gets \mathcal{C} \setminus \{M_i\}$ \Comment{all other models}
\State $e^\ast \gets E_v(I^\ast)$
\For{each $M_j \in S_i$}  \Comment{same-prompt candidates}
  \State $I_j \gets M_j(p)$
  \State $e_j \gets E_v(I_j)$
\EndFor
\State $j^\star \gets \arg\min_{M_j \in S_i} S(e^\ast, e_j)$ \Comment{farthest = least similar}
\State $x^+ \gets I_{j^\star}$,\quad $x^- \gets I^\ast$
\vspace{0.3em}
\State $\delta^{(0)} \gets 0$
\For{$t = 0$ to $T-1$}
  \State $\tilde I^{(t)} \gets I^\ast + \delta^{(t)}$
  \State $z^{(t)} \gets E_v(\tilde I^{(t)})$
  \State $s^+ \gets S\big(z^{(t)}, E_v(x^+)\big)$,\quad $s^- \gets S\big(z^{(t)}, E_v(x^-)\big)$
  \State $p^+ \gets \dfrac{\exp(s^+/\tau)}{\exp(s^+/\tau) + \exp(s^-/\tau)}$
  \State $p^- \gets \dfrac{\exp(s^-/\tau)}{\exp(s^+/\tau) + \exp(s^-/\tau)}$
  \State $\mathcal{L}^{(t)} \gets -\log p^+ + \log p^-$ \Comment{contrastive loss}
  \State $\delta^{(t+1)} \gets \Pi_{\varepsilon}\!\big(I^\ast,\; \delta^{(t)} - \eta \nabla_{\delta} \mathcal{L}^{(t)}\big)$ \Comment{$\Pi_{\varepsilon}(I^\ast,\cdot)$ projects onto the $\ell_\infty$-ball of radius $\varepsilon$ around $I^\ast$.}
\EndFor
\State $\tilde I^\ast \gets I^\ast + \delta^{(T)}$
\end{algorithmic}
\end{algorithm}

\paragraph{Method.}
We adapt the adversarial attack formulation of \citet{hu2025transferable} to our deanonymization-defense setting. Our adversarial post-processing procedure is summarized in Algorithm \ref{alg:defense}. After model $M_i$ generates an image $I^\ast$ for prompt $p$, the defender queries all other leaderboard models with the same prompt, embeds their outputs, and selects the \emph{farthest} image as the positive target based on embedding similarity. The defender then optimizes a perturbation that pulls $I^\ast$ toward this farthest target while pushing it away from its original embedding.

Because the defender does not know the attacker's encoder, the perturbation must generalize across embedding models. We adopt the contrastive loss formulation fro Hu et al. which jointly attracts the embedding toward the positive example and repels it from the negative one. To further improve transferability, we optimize the perturbation using an ensemble of local encoders (\Cref{tab:surrogate}, Appendix), a technique shown to significantly enhance cross-model robustness of adversarial examples \cite{liu2017delving}.

\paragraph{Results.}
We vary the perturbation budget $\varepsilon \in \{2,4,8\}$ to evaluate the
effectiveness of our adversarial post-processing defense. The full experimental
setup is provided in Appendix~\Cref{sec:details_defense}. As shown in \Cref{tab:defense_results}, the defense consistently reduces attack accuracy, with larger perturbation budgets providing stronger protection. However, larger budgets also introduce more visible perturbations (see \Cref{fig:adv_examples}). Notably, even after applying the defense, the attack still achieves non-trivial accuracy; for example, at $\varepsilon=2$ the Top-1 accuracy remains as high as 75\%.
\begin{table}[t]
\centering
\caption{Attack accuracy before and after adversarial post-processing under different perturbation budgets $\epsilon$.}
\label{tab:defense_results}

\resizebox{\columnwidth}{!}{
\begin{tabular}{c c c c c}
\toprule
\textbf{Accuracy} & \textbf{Original} 
& $\boldsymbol{\varepsilon=2}$ 
& $\boldsymbol{\varepsilon=4}$ 
& $\boldsymbol{\varepsilon=8}$ \\
\midrule
Top-1 
& 0.90 
& 0.75\;($\downarrow$16.7\%) 
& 0.54\;($\downarrow$40.0\%) 
& 0.46\;($\downarrow$48.9\%) \\
Top-2 
& 0.96 
& 0.82\;($\downarrow$14.6\%) 
& 0.71\;($\downarrow$26.0\%) 
& 0.57\;($\downarrow$40.6\%) \\
Top-3 
& 0.97 
& 0.85\;($\downarrow$12.4\%) 
& 0.77\;($\downarrow$20.6\%) 
& 0.64\;($\downarrow$34.0\%) \\
\bottomrule
\end{tabular}}
\end{table}

\paragraph{Undoing Post-Processing.}
An adversary aware of such post-processing attempts by leaderboard maintainers may adapt and try to remove the influence of these perturbations. Following \citet{honig2025adversarial}, we evaluate two strategies: Gaussian noising, which simply adds random noise, and noisy upscaling, which uses a diffusion model but incurs higher cost. At $\varepsilon{=}2$, our defense yields a Top-1 accuracy of 0.75; applying Gaussian noising reduces it to 0.69 and noisy upscaling to 0.72. These results show that post-hoc operations do not significantly weaken our defense, indicating that standard adversarial-example mitigation strategies might be ineffective against embedding-space displacement.

\section{Discussion}
\label{sec:discussion}

\paragraph{Financial Cost Analysis.}
An important aspect of this attack is its low financial cost. To deanonymize a single leaderboard image, the adversary must generate $I \times |M|$ images, where $|M|$ is the number of models and $I$ is the number of generations per model. The total cost can be expressed as
\[
\text{Cost} = I \times \sum_{i=1}^{|M|} C_i,
\]
where $C_i$ is the price of generating one image from model $M_i$. Based on public pricing data from the Artificial Analysis leaderboard, the cumulative cost of generating one image across all commercial models is approximately \$1.08. Thus, the total cost for deanonymizing one sample is about \$1.08$I$. As shown in our ablation study, even with only five images per model (achieving roughly 85\% accuracy using 22 models), the total cost remains around \$5.4 per target image. Such a cost is negligible for large organizations or coordinated adversaries seeking to manipulate leaderboard rankings to promote their own models.

\paragraph{What Does This Mean for Arenas?}
Our results demonstrate that adversaries can easily deanonymize images in text-to-image leaderboards, raising fundamental questions about the reliability of voting-based evaluation. While image perturbations appear to mitigate this issue, such defenses can often be reversed by standard denoising techniques and may degrade visual quality, undermining fair user judgments. Other measures such as user authentication, rate limiting, malicious-user detection, and CAPTCHAs increase the cost of manipulation but remain imperfect, as CAPTCHAs can be bypassed by automated tools~\cite{lee2023holistic,teoh2025captchas}. 

Leaderboards could also leverage the proposed distinguishability metric to identify prompts that are more vulnerable to deanonymization. However, restricting evaluations to low-distinguishability prompts limits meaningful visual comparisons, challenging the purpose of human-in-the-loop ranking systems. Ultimately, there is no foolproof defense: any mitigation introduces trade-offs between fairness, usability, and transparency. We hope this work encourages the community to critically reassess the assumption of model anonymity in T2I arenas and to design future evaluations with greater resilience to such vulnerabilities.

\section{Conclusion}
We demonstrate that model anonymity in text-to-image leaderboards is fragile: even without prompt control, simple centroid-based analysis of image embeddings can deanonymize generations with high accuracy. Our findings reveal that model-specific visual signatures are inherent and persistent across architectures, prompting a re-examination of fairness and reliability in human-voting evaluation systems. While lightweight defenses such as embedding-space perturbations can reduce attack success, they introduce trade-offs in image fidelity and usability. We hope this work motivates the community to design future leaderboard and evaluation protocols that preserve transparency while mitigating deanonymization risks.

\section*{Acknowledgments}

This work has been supported by  NSF grant CNS-2247484.

{
    \small
    \bibliographystyle{ieeenat_fullname}
    \bibliography{main}

\begin{thebibliography}{67}
\providecommand{\natexlab}[1]{#1}
\providecommand{\url}[1]{\texttt{#1}}
\expandafter\ifx\csname urlstyle\endcsname\relax
  \providecommand{\doi}[1]{doi: #1}\else
  \providecommand{\doi}{doi: \begingroup \urlstyle{rm}\Url}\fi

\bibitem[Analysis(2025)]{artificial}
Artificial Analysis.
\newblock Artificial analysis: Independent analysis of ai.
\newblock \url{https://artificialanalysis.ai/}, 2025.

\bibitem[Betker et~al.(2023)Betker, Goh, Jing, Brooks, Wang, Li, Ouyang, Zhuang, Lee, Guo, et~al.]{betker2023improving}
James Betker, Gabriel Goh, Li Jing, Tim Brooks, Jianfeng Wang, Linjie Li, Long Ouyang, Juntang Zhuang, Joyce Lee, Yufei Guo, et~al.
\newblock Improving image generation with better captions.
\newblock \emph{Computer Science. https://cdn. openai. com/papers/dall-e-3. pdf}, 2\penalty0 (3):\penalty0 8, 2023.

\bibitem[Cai et~al.(2025)Cai, Chen, Chen, Li, Long, Pan, Qiu, Zhang, Gao, Xu, et~al.]{cai2025hidream}
Qi Cai, Jingwen Chen, Yang Chen, Yehao Li, Fuchen Long, Yingwei Pan, Zhaofan Qiu, Yiheng Zhang, Fengbin Gao, Peihan Xu, et~al.
\newblock Hidream-i1: A high-efficient image generative foundation model with sparse diffusion transformer.
\newblock \emph{arXiv preprint arXiv:2505.22705}, 2025.

\bibitem[Chiang et~al.(2024)Chiang, Zheng, Sheng, Angelopoulos, Li, Li, Zhu, Zhang, Jordan, Gonzalez, et~al.]{chiang2024chatbot}
Wei-Lin Chiang, Lianmin Zheng, Ying Sheng, Anastasios~Nikolas Angelopoulos, Tianle Li, Dacheng Li, Banghua Zhu, Hao Zhang, Michael Jordan, Joseph~E Gonzalez, et~al.
\newblock Chatbot arena: An open platform for evaluating llms by human preference.
\newblock In \emph{International Conference on Machine Learning}, 2024.

\bibitem[Corneanu et~al.(2025)Corneanu, Feng, and Martinez]{corneanu2025structured}
Ciprian~A Corneanu, Qianli Feng, and Aleix~M Martinez.
\newblock Structured human assessment of text-to-image generative models.
\newblock In \emph{2025 IEEE/CVF Winter Conference on Applications of Computer Vision (WACV)}, pages 4481--4490. IEEE, 2025.

\bibitem[Dubey et~al.(2024)Dubey, Jauhri, Pandey, Kadian, Al-Dahle, Letman, Mathur, Schelten, Yang, Fan, et~al.]{dubey2024llama}
Abhimanyu Dubey, Abhinav Jauhri, Abhinav Pandey, Abhishek Kadian, Ahmad Al-Dahle, Aiesha Letman, Akhil Mathur, Alan Schelten, Amy Yang, Angela Fan, et~al.
\newblock The llama 3 herd of models.
\newblock \emph{arXiv e-prints}, pages arXiv--2407, 2024.

\bibitem[Dzanic et~al.(2020)Dzanic, Shah, and Witherden]{dzanic2020fourier}
Tarik Dzanic, Karan Shah, and Freddie Witherden.
\newblock Fourier spectrum discrepancies in deep network generated images.
\newblock \emph{Advances in neural information processing systems}, 33:\penalty0 3022--3032, 2020.

\bibitem[Esser et~al.(2024)Esser, Kulal, Blattmann, Entezari, M{\"u}ller, Saini, Levi, Lorenz, Sauer, Boesel, et~al.]{esser2024scaling}
Patrick Esser, Sumith Kulal, Andreas Blattmann, Rahim Entezari, Jonas M{\"u}ller, Harry Saini, Yam Levi, Dominik Lorenz, Axel Sauer, Frederic Boesel, et~al.
\newblock Scaling rectified flow transformers for high-resolution image synthesis.
\newblock In \emph{Forty-first international conference on machine learning}, 2024.

\bibitem[Guo et~al.(2024)Guo, Jiang, Zhang, Lu, and Li]{guo2024one}
Ji Guo, Wenbo Jiang, Rui Zhang, Guoming Lu, and Hongwei Li.
\newblock One prompt to verify your models: Black-box text-to-image models verification via non-transferable adversarial attacks.
\newblock \emph{arXiv preprint arXiv:2410.22725}, 2024.

\bibitem[He et~al.(2016)He, Zhang, Ren, and Sun]{he2016deep}
Kaiming He, Xiangyu Zhang, Shaoqing Ren, and Jian Sun.
\newblock Deep residual learning for image recognition.
\newblock In \emph{Proceedings of the IEEE conference on computer vision and pattern recognition}, pages 770--778, 2016.

\bibitem[H{\"o}nig et~al.(2025)H{\"o}nig, Rando, Carlini, and Tram{\`e}r]{honig2025adversarial}
Robert H{\"o}nig, Javier Rando, Nicholas Carlini, and Florian Tram{\`e}r.
\newblock Adversarial perturbations cannot reliably protect artists from generative ai.
\newblock In \emph{International Conference on Learning Representations}, 2025.

\bibitem[Hu et~al.(2025)Hu, Yu, Zhang, Robey, Zou, Xu, Hu, and Fredrikson]{hu2025transferable}
Kai Hu, Weichen Yu, Li Zhang, Alexander Robey, Andy Zou, Chengming Xu, Haoqi Hu, and Matt Fredrikson.
\newblock Transferable adversarial attacks on black-box vision-language models.
\newblock \emph{arXiv preprint arXiv:2505.01050}, 2025.

\bibitem[Huang et~al.(2025)Huang, Nasr, Angelopoulos, Carlini, Chiang, Choquette-Choo, Ippolito, Jagielski, Lee, Liu, et~al.]{huang2025exploring}
Yangsibo Huang, Milad Nasr, Anastasios Angelopoulos, Nicholas Carlini, Wei-Lin Chiang, Christopher~A Choquette-Choo, Daphne Ippolito, Matthew Jagielski, Katherine Lee, Ken~Ziyu Liu, et~al.
\newblock Exploring and mitigating adversarial manipulation of voting-based leaderboards.
\newblock In \emph{International Conference on Machine Learning}, 2025.

\bibitem[Ilharco et~al.(2021)Ilharco, Wortsman, Wightman, Gordon, Carlini, Taori, Dave, Shankar, Namkoong, Miller, Hajishirzi, Farhadi, and Schmidt]{ilharco_gabriel_2021_5143773}
Gabriel Ilharco, Mitchell Wortsman, Ross Wightman, Cade Gordon, Nicholas Carlini, Rohan Taori, Achal Dave, Vaishaal Shankar, Hongseok Namkoong, John Miller, Hannaneh Hajishirzi, Ali Farhadi, and Ludwig Schmidt.
\newblock Openclip, 2021.

\bibitem[Kim et~al.(2024)Kim, Min, Patel, Cheng, and Yang]{kim2024wouaf}
Changhoon Kim, Kyle Min, Maitreya Patel, Sheng Cheng, and Yezhou Yang.
\newblock Wouaf: Weight modulation for user attribution and fingerprinting in text-to-image diffusion models.
\newblock In \emph{Proceedings of the IEEE/CVF Conference on Computer Vision and Pattern Recognition}, pages 8974--8983, 2024.

\bibitem[Labs(2024)]{flux2024}
Black~Forest Labs.
\newblock Flux.
\newblock \url{https://github.com/black-forest-labs/flux}, 2024.

\bibitem[Lai et~al.(2023)Lai, Ngo, Veyseh, Dernoncourt, and Nguyen]{lai2023openllmbenchmark}
Viet Lai, Nghia~Trung Ngo, Amir Pouran~Ben Veyseh, Franck Dernoncourt, and Thien~Huu Nguyen.
\newblock Open multilingual llm evaluation leaderboard, 2023.

\bibitem[Lee et~al.(2023)Lee, Yasunaga, Meng, Mai, Park, Gupta, Zhang, Narayanan, Teufel, Bellagente, et~al.]{lee2023holistic}
Tony Lee, Michihiro Yasunaga, Chenlin Meng, Yifan Mai, Joon~Sung Park, Agrim Gupta, Yunzhi Zhang, Deepak Narayanan, Hannah Teufel, Marco Bellagente, et~al.
\newblock Holistic evaluation of text-to-image models.
\newblock \emph{Advances in Neural Information Processing Systems}, 36:\penalty0 69981--70011, 2023.

\bibitem[Li et~al.(2024{\natexlab{a}})Li, Kamko, Akhgari, Sabet, Xu, and Doshi]{li2024playground}
Daiqing Li, Aleks Kamko, Ehsan Akhgari, Ali Sabet, Linmiao Xu, and Suhail Doshi.
\newblock Playground v2. 5: Three insights towards enhancing aesthetic quality in text-to-image generation.
\newblock \emph{arXiv preprint arXiv:2402.17245}, 2024{\natexlab{a}}.

\bibitem[Li et~al.(2024{\natexlab{b}})Li, Zou, Wang, Majumder, Xie, Manmatha, Swaminathan, Tu, Ermon, and Soatto]{li2024scalability}
Hao Li, Yang Zou, Ying Wang, Orchid Majumder, Yusheng Xie, R Manmatha, Ashwin Swaminathan, Zhuowen Tu, Stefano Ermon, and Stefano Soatto.
\newblock On the scalability of diffusion-based text-to-image generation.
\newblock In \emph{IEEE/CVF Conference on Computer Vision and Pattern Recognition}, 2024{\natexlab{b}}.

\bibitem[Li et~al.(2023{\natexlab{a}})Li, Wang, and Xie]{li2023clipav2}
Xianhang Li, Zeyu Wang, and Cihang Xie.
\newblock Clipa-v2: Scaling clip training with 81.1
\newblock \emph{arXiv preprint arXiv:2306.15658}, 2023{\natexlab{a}}.

\bibitem[Li et~al.(2023{\natexlab{b}})Li, Zhang, Dubois, Taori, Gulrajani, Guestrin, Liang, and Hashimoto]{alpaca_eval}
Xuechen Li, Tianyi Zhang, Yann Dubois, Rohan Taori, Ishaan Gulrajani, Carlos Guestrin, Percy Liang, and Tatsunori~B. Hashimoto.
\newblock Alpacaeval: An automatic evaluator of instruction-following models.
\newblock \url{https://github.com/tatsu-lab/alpaca_eval}, 2023{\natexlab{b}}.

\bibitem[Liu et~al.(2017)Liu, Chen, Liu, and Song]{liu2017delving}
Yanpei Liu, Xinyun Chen, Chang Liu, and Dawn Song.
\newblock Delving into transferable adversarial examples and black-box attacks.
\newblock In \emph{International Conference on Learning Representations}, 2017.

\bibitem[Lu et~al.(2023)Lu, Wang, Wang, Guan, Gao, and Zheng]{Lu_2023_ICCV}
Dong Lu, Zhiqiang Wang, Teng Wang, Weili Guan, Hongchang Gao, and Feng Zheng.
\newblock Set-level guidance attack: Boosting adversarial transferability of vision-language pre-training models.
\newblock In \emph{IEEE International Conference on Computer Vision}, pages 102--111, 2023.

\bibitem[Manning et~al.(2008)Manning, Raghavan, and Sch{\"u}tze]{manning2008vector}
Christopher Manning, Prabhakar Raghavan, and Hinrich Sch{\"u}tze.
\newblock Vector space classification.
\newblock \emph{Introduction to Information Retrieval}, pages 289--317, 2008.

\bibitem[Marra et~al.(2019)Marra, Gragnaniello, Verdoliva, and Poggi]{marra2019gans}
Francesco Marra, Diego Gragnaniello, Luisa Verdoliva, and Giovanni Poggi.
\newblock Do gans leave artificial fingerprints?
\newblock In \emph{2019 IEEE conference on multimedia information processing and retrieval (MIPR)}, pages 506--511. IEEE, 2019.

\bibitem[Matatov et~al.(2024)Matatov, Qu{\'e}r{\'e}, Amir, and Naaman]{matatov2024examining}
Hana Matatov, Marianne Aubin~Le Qu{\'e}r{\'e}, Ofra Amir, and Mor Naaman.
\newblock Examining the prevalence and dynamics of ai-generated media in art subreddits.
\newblock \emph{arXiv preprint arXiv:2410.07302}, 2024.

\bibitem[{Midjourney}(2025)]{midjourney}
{Midjourney}.
\newblock Midjourney generative image model.
\newblock \url{https://www.midjourney.com}, 2025.
\newblock Accessed: 2025-09-20.

\bibitem[Min et~al.(2025)Min, Pang, Du, Liu, Cheng, and Lin]{min2025improving}
Rui Min, Tianyu Pang, Chao Du, Qian Liu, Minhao Cheng, and Min Lin.
\newblock Improving your model ranking on chatbot arena by vote rigging.
\newblock In \emph{International Conference on Machine Learning}, 2025.

\bibitem[Minixhofer et~al.(2024)Minixhofer, Klejch, and Bell]{minixhofer2024ttsds}
Christoph Minixhofer, Ond{\v{r}}ej Klejch, and Peter Bell.
\newblock Ttsds-text-to-speech distribution score.
\newblock In \emph{2024 IEEE Spoken Language Technology Workshop (SLT)}, pages 766--773. IEEE, 2024.

\bibitem[M{\o}ller et~al.(2025)M{\o}ller, Romero, Jurgens, and Aiello]{moller2025impact}
Anders~Giovanni M{\o}ller, Daniel~M Romero, David Jurgens, and Luca~Maria Aiello.
\newblock The impact of generative ai on social media: An experimental study.
\newblock \emph{arXiv preprint arXiv:2506.14295}, 2025.

\bibitem[mrfakename et~al.(2025)mrfakename, Srivastav, Fourrier, Pouget, Lacombe, main, Gandhi, Passos, and Cuenca]{tts-arena-v2}
mrfakename, Vaibhav Srivastav, Clémentine Fourrier, Lucain Pouget, Yoach Lacombe, main, Sanchit Gandhi, Apolinário Passos, and Pedro Cuenca.
\newblock Tts arena 2.0: Benchmarking text-to-speech models in the wild.
\newblock \url{https://huggingface.co/spaces/TTS-AGI/TTS-Arena-V2}, 2025.

\bibitem[(MTEB)(2025{\natexlab{a}})]{mteb_arena}
Massive Text Embedding~Benchmark (MTEB).
\newblock Mteb arena.
\newblock \url{https://huggingface.co/spaces/mteb/arena}, 2025{\natexlab{a}}.
\newblock Accessed: 2025-03-01.

\bibitem[(MTEB)(2025{\natexlab{b}})]{mteb_leaderboard}
Massive Text Embedding~Benchmark (MTEB).
\newblock Mteb leaderboard.
\newblock \url{https://huggingface.co/spaces/mteb/leaderboard}, 2025{\natexlab{b}}.
\newblock Accessed: 2025-03-01.

\bibitem[Muennighoff et~al.(2023)Muennighoff, Tazi, Magne, and Reimers]{muennighoff2023mteb}
Niklas Muennighoff, Nouamane Tazi, Loic Magne, and Nils Reimers.
\newblock Mteb: Massive text embedding benchmark.
\newblock In \emph{Proceedings of the 17th Conference of the European Chapter of the Association for Computational Linguistics}, pages 2014--2037, 2023.

\bibitem[Naseh et~al.(2024)Naseh, Thai, Iyyer, and Houmansadr]{naseh2024iteratively}
Ali Naseh, Katherine Thai, Mohit Iyyer, and Amir Houmansadr.
\newblock Iteratively prompting multimodal llms to reproduce natural and ai-generated images.
\newblock In \emph{Conference on Language Modeling}, 2024.

\bibitem[Nie et~al.(2023)Nie, Kim, Yang, and Ren]{nie2023attributing}
Guangyu Nie, Changhoon Kim, Yezhou Yang, and Yi Ren.
\newblock Attributing image generative models using latent fingerprints.
\newblock In \emph{International Conference on Machine Learning}, pages 26150--26165. PMLR, 2023.

\bibitem[Paech(2023)]{paech2023eq}
Samuel~J Paech.
\newblock Eq-bench: An emotional intelligence benchmark for large language models.
\newblock \emph{arXiv preprint arXiv:2312.06281}, 2023.

\bibitem[Park et~al.(2024)Park, Eirich, Luckow, and Sedlmair]{park2024we}
Hyerim Park, Joscha Eirich, Andre Luckow, and Michael Sedlmair.
\newblock " we are visual thinkers, not verbal thinkers!": A thematic analysis of how professional designers use generative ai image generation tools.
\newblock In \emph{Proceedings of the 13th Nordic Conference on Human-Computer Interaction}, pages 1--14, 2024.

\bibitem[Pasquini et~al.(2025)Pasquini, Kornaropoulos, and Ateniese]{pasquini2025llmmap}
Dario Pasquini, Evgenios~M Kornaropoulos, and Giuseppe Ateniese.
\newblock {LLMmap}: Fingerprinting for large language models.
\newblock In \emph{USENIX Security Symposium}, 2025.

\bibitem[Penedo et~al.(2024)Penedo, Kydlíček, Cappelli, Sasko, and Wolf]{penedo2024datatrove}
Guilherme Penedo, Hynek Kydlíček, Alessandro Cappelli, Mario Sasko, and Thomas Wolf.
\newblock Datatrove: large scale data processing, 2024.

\bibitem[Podell et~al.(2024)Podell, English, Lacey, Blattmann, Dockhorn, M{\"u}ller, Penna, and Rombach]{podell2024sdxl}
Dustin Podell, Zion English, Kyle Lacey, Andreas Blattmann, Tim Dockhorn, Jonas M{\"u}ller, Joe Penna, and Robin Rombach.
\newblock Sdxl: Improving latent diffusion models for high-resolution image synthesis.
\newblock In \emph{International Conference on Learning Representations}, 2024.

\bibitem[Qin et~al.(2025)Qin, Zhuo, Xin, Du, Li, Fu, Lu, Yuan, Li, Liu, et~al.]{qin2025lumina}
Qi Qin, Le Zhuo, Yi Xin, Ruoyi Du, Zhen Li, Bin Fu, Yiting Lu, Jiakang Yuan, Xinyue Li, Dongyang Liu, et~al.
\newblock Lumina-image 2.0: A unified and efficient image generative framework.
\newblock \emph{arXiv preprint arXiv:2503.21758}, 2025.

\bibitem[Radford et~al.(2021)Radford, Kim, Hallacy, Ramesh, Goh, Agarwal, Sastry, Askell, Mishkin, Clark, Krueger, and Sutskever]{Radford2021LearningTV}
Alec Radford, Jong~Wook Kim, Chris Hallacy, A. Ramesh, Gabriel Goh, Sandhini Agarwal, Girish Sastry, Amanda Askell, Pamela Mishkin, Jack Clark, Gretchen Krueger, and Ilya Sutskever.
\newblock Learning transferable visual models from natural language supervision.
\newblock In \emph{ICML}, 2021.

\bibitem[Rombach et~al.(2022)Rombach, Blattmann, Lorenz, Esser, and Ommer]{Rombach_2022_CVPR}
Robin Rombach, Andreas Blattmann, Dominik Lorenz, Patrick Esser, and Bj\"orn Ommer.
\newblock High-resolution image synthesis with latent diffusion models.
\newblock In \emph{Proceedings of the IEEE/CVF Conference on Computer Vision and Pattern Recognition (CVPR)}, pages 10684--10695, 2022.

\bibitem[Sammet and Krestel(2023)]{sammet2023domain}
Jill Sammet and Ralf Krestel.
\newblock Domain-specific keyword extraction using bert.
\newblock In \emph{Proceedings of the 4th Conference on Language, Data and Knowledge}, pages 659--665, 2023.

\bibitem[Schuhmann et~al.(2022)Schuhmann, Beaumont, Vencu, Gordon, Wightman, Cherti, Coombes, Katta, Mullis, Wortsman, et~al.]{schuhmann2022laion}
Christoph Schuhmann, Romain Beaumont, Richard Vencu, Cade Gordon, Ross Wightman, Mehdi Cherti, Theo Coombes, Aarush Katta, Clayton Mullis, Mitchell Wortsman, et~al.
\newblock Laion-5b: An open large-scale dataset for training next generation image-text models.
\newblock In \emph{Advances in Neural Information Processing Systems}, 2022.

\bibitem[Song and Itti(2025)]{song2025riemannian}
Hae~Jin Song and Laurent Itti.
\newblock Riemannian-geometric fingerprints of generative models.
\newblock \emph{arXiv preprint arXiv:2506.22802}, 2025.

\bibitem[Song et~al.(2024)Song, Khayatkhoei, and AbdAlmageed]{song2024manifpt}
Hae~Jin Song, Mahyar Khayatkhoei, and Wael AbdAlmageed.
\newblock {ManiFPT}: Defining and analyzing fingerprints of generative models.
\newblock In \emph{Proceedings of the IEEE/CVF Conference on Computer Vision and Pattern Recognition}, pages 10791--10801, 2024.

\bibitem[Suri et~al.(2026)Suri, Chaudhari, Peng, Naseh, Oprea, and Houmansadr]{suri2026exploiting}
Anshuman Suri, Harsh Chaudhari, Yuefeng Peng, Ali Naseh, Alina Oprea, and Amir Houmansadr.
\newblock Exploiting leaderboards for large-scale distribution of malicious models.
\newblock In \emph{IEEE Symposium on Security and Privacy (S\&P)}, 2026.

\bibitem[Teoh et~al.(2025)Teoh, Lin, Li, Liu, Sollomoni, Harel, and Dong]{teoh2025captchas}
Xiwen Teoh, Yun Lin, Siqi Li, Ruofan Liu, Avi Sollomoni, Yaniv Harel, and Jin~Song Dong.
\newblock Are $\{$CAPTCHAs$\}$ still bot-hard? generalized visual $\{$CAPTCHA$\}$ solving with agentic vision language model.
\newblock In \emph{USENIX Security Symposium}, 2025.

\bibitem[Tschannen et~al.(2025)Tschannen, Gritsenko, Wang, Naeem, Alabdulmohsin, Parthasarathy, Evans, Beyer, Xia, Mustafa, Hénaff, Harmsen, Steiner, and Zhai]{tschannen2025siglip2multilingualvisionlanguage}
Michael Tschannen, Alexey Gritsenko, Xiao Wang, Muhammad~Ferjad Naeem, Ibrahim Alabdulmohsin, Nikhil Parthasarathy, Talfan Evans, Lucas Beyer, Ye Xia, Basil Mustafa, Olivier Hénaff, Jeremiah Harmsen, Andreas Steiner, and Xiaohua Zhai.
\newblock Siglip 2: Multilingual vision-language encoders with improved semantic understanding, localization, and dense features, 2025.

\bibitem[Wang et~al.(2024)Wang, Dong, Zhu, Qin, Liu, Fang, Wang, and Liu]{wangTransfer}
Haodi Wang, Kai Dong, Zhilei Zhu, Haotong Qin, Aishan Liu, Xiaolin Fang, Jiakai Wang, and Xianglong Liu.
\newblock Transferable multimodal attack on vision-language pre-training models.
\newblock In \emph{IEEE Symposium on Security and Privacy}, pages 1722--1740, 2024.

\bibitem[Wei and Tyson(2024)]{wei2024understanding}
Yiluo Wei and Gareth Tyson.
\newblock Understanding the impact of ai-generated content on social media: The pixiv case.
\newblock In \emph{Proceedings of the 32nd ACM International Conference on Multimedia}, pages 6813--6822, 2024.

\bibitem[Wu et~al.(2025)Wu, Li, Zhou, Lin, Gao, Yan, ming Yin, Bai, Xu, Chen, Chen, Tang, Zhang, Wang, Yang, Yu, Cheng, Liu, Li, Zhang, Meng, Wei, Ni, Chen, Cao, Peng, Qu, Wu, Wang, Yu, Wen, Feng, Xu, Wang, Zhang, Zhu, Wu, Cai, and Liu]{wu2025qwenimagetechnicalreport}
Chenfei Wu, Jiahao Li, Jingren Zhou, Junyang Lin, Kaiyuan Gao, Kun Yan, Sheng ming Yin, Shuai Bai, Xiao Xu, Yilei Chen, Yuxiang Chen, Zecheng Tang, Zekai Zhang, Zhengyi Wang, An Yang, Bowen Yu, Chen Cheng, Dayiheng Liu, Deqing Li, Hang Zhang, Hao Meng, Hu Wei, Jingyuan Ni, Kai Chen, Kuan Cao, Liang Peng, Lin Qu, Minggang Wu, Peng Wang, Shuting Yu, Tingkun Wen, Wensen Feng, Xiaoxiao Xu, Yi Wang, Yichang Zhang, Yongqiang Zhu, Yujia Wu, Yuxuan Cai, and Zenan Liu.
\newblock Qwen-image technical report, 2025.

\bibitem[Wu et~al.(2023)Wu, Jiang, Khan, Fu, Ruis, Grefenstette, and Rocktäschel]{2023ChatArena}
Yuxiang Wu, Zhengyao Jiang, Akbir Khan, Yao Fu, Laura Ruis, Edward Grefenstette, and Tim Rocktäschel.
\newblock Chatarena: Multi-agent language game environments for large language models.
\newblock \url{https://github.com/chatarena/chatarena}, 2023.

\bibitem[Xie et~al.(2025)Xie, Bie, Mao, Song, Wang, Chen, and Chen]{Xie_2025_CVPR}
Peng Xie, Yequan Bie, Jianda Mao, Yangqiu Song, Yang Wang, Hao Chen, and Kani Chen.
\newblock Chain of attack: On the robustness of vision-language models against transfer-based adversarial attacks.
\newblock In \emph{IEEE/CVF Conference on Computer Vision and Pattern Recognition}, pages 14679--14689, 2025.

\bibitem[Yang et~al.(2024)Yang, Singh, and Menczer]{yang2024characteristics}
Kai-Cheng Yang, Danishjeet Singh, and Filippo Menczer.
\newblock Characteristics and prevalence of fake social media profiles with ai-generated faces.
\newblock \emph{arXiv preprint arXiv:2401.02627}, 2024.

\bibitem[Yao and Juarez(2025)]{yao2025authprint}
Kai Yao and Marc Juarez.
\newblock Authprint: Fingerprinting generative models against malicious model providers.
\newblock \emph{arXiv preprint arXiv:2508.05691}, 2025.

\bibitem[Yu et~al.(2019)Yu, Davis, and Fritz]{yu2019attributing}
Ning Yu, Larry~S Davis, and Mario Fritz.
\newblock Attributing fake images to gans: Learning and analyzing gan fingerprints.
\newblock In \emph{IEEE/CVF Conference on Computer Vision and Pattern Recognition}, 2019.

\bibitem[Yu et~al.(2021)Yu, Skripniuk, Abdelnabi, and Fritz]{yu2021artificial}
Ning Yu, Vladislav Skripniuk, Sahar Abdelnabi, and Mario Fritz.
\newblock Artificial fingerprinting for generative models: Rooting deepfake attribution in training data.
\newblock In \emph{Proceedings of the IEEE/CVF International conference on computer vision}, pages 14448--14457, 2021.

\bibitem[Zhai et~al.(2023)Zhai, Mustafa, Kolesnikov, and Beyer]{zhai2023sigmoid}
Xiaohua Zhai, Basil Mustafa, Alexander Kolesnikov, and Lucas Beyer.
\newblock Sigmoid loss for language image pre-training.
\newblock \emph{arXiv preprint arXiv:2303.15343}, 2023.

\bibitem[Zhang et~al.(2025{\natexlab{a}})Zhang, Yu, Min, Xin, Wei, Shi, Huang, Kong, Xin, Jiang, et~al.]{zhang2025generative}
Ruihan Zhang, Borou Yu, Jiajian Min, Yetong Xin, Zheng Wei, Juncheng~Nemo Shi, Mingzhen Huang, Xianghao Kong, Nix~Liu Xin, Shanshan Jiang, et~al.
\newblock Generative ai for film creation: A survey of recent advances.
\newblock In \emph{IEEE/CVF Conference on Computer Vision and Pattern Recognition}, 2025{\natexlab{a}}.

\bibitem[Zhang et~al.(2025{\natexlab{b}})Zhang, Li, Long, Zhang, Lin, Yang, Xie, Yang, Liu, Lin, Huang, and Zhou]{qwen3embedding}
Yanzhao Zhang, Mingxin Li, Dingkun Long, Xin Zhang, Huan Lin, Baosong Yang, Pengjun Xie, An Yang, Dayiheng Liu, Junyang Lin, Fei Huang, and Jingren Zhou.
\newblock Qwen3 embedding: Advancing text embedding and reranking through foundation models.
\newblock \emph{arXiv preprint arXiv:2506.05176}, 2025{\natexlab{b}}.

\bibitem[Zhao et~al.(2025)Zhao, Rush, and Goyal]{zhao-etal-2025-challenges}
Wenting Zhao, Alexander~M Rush, and Tanya Goyal.
\newblock Challenges in trustworthy human evaluation of chatbots.
\newblock In \emph{Findings of the Association for Computational Linguistics: NAACL}, 2025.

\bibitem[Zhao et~al.(2023)Zhao, Pang, Du, Yang, Li, Cheung, and Lin]{zhao2023evaluating}
Yunqing Zhao, Tianyu Pang, Chao Du, Xiao Yang, Chongxuan Li, Ngai-Man~Man Cheung, and Min Lin.
\newblock On evaluating adversarial robustness of large vision-language models.
\newblock In \emph{Advances in Neural Information Processing Systems}, pages 54111--54138, 2023.

\bibitem[Zhou and Lee(2024)]{zhou2024generative}
Eric Zhou and Dokyun Lee.
\newblock Generative artificial intelligence, human creativity, and art.
\newblock \emph{PNAS nexus}, 3\penalty0 (3):\penalty0 pgae052, 2024.

\end{thebibliography}
}

\clearpage
\setcounter{page}{1}
\maketitlesupplementary

\begin{table*}[ht]
\centering
\caption{
Full list of T2I models used in our experiments, along with their provider, image resolution, and the number of inference steps.
Where inference-step counts are available on the ArtificialAnalysis methodology page, we adopt those values directly.
For models not mentioned there, we use the default values documented on their respective Hugging Face model pages.  
For OpenAI and Midjourney models we did not explicitly set the number of inference steps and used their internal default generation settings.  
}
\small
\begin{tabular}{l l c c}
\toprule
\textbf{Model} & \textbf{Company / Provider} & \textbf{Resolution (W$\times$H)} & \textbf{Inference Steps} \\
\midrule
DALL·E 3 HD~\cite{betker2023improving} & OpenAI & 1024×1024 & -- \\
FLUX.1-dev~\cite{flux2024} & Black Forest Labs & 1024×1024 & 28 \\
FLUX.1-schnell~\cite{flux2024} & Black Forest Labs & 1024×1024 & 4 \\
FLUX.1-Krea-dev~\cite{flux2024} & Black Forest Labs & 1024×1024 & 28 \\
FLUX-1.1-pro & Black Forest Labs & 1024×1024 & 28 \\   %
flux.1-kontext-pro & Black Forest Labs & 1024×1024 & 28 \\ %
Stable Diffusion v1.5~\cite{Rombach_2022_CVPR} & Stability AI & 512×512 & 50 \\
Stable Diffusion 2.1~\cite{Rombach_2022_CVPR} & Stability AI & 768×768 & 50 \\
Stable Diffusion XL~\cite{podell2024sdxl} & Stability AI & 1024×1024 & 30 \\
SDXL Turbo~\cite{podell2024sdxl} & Stability AI & 1024×1024 & 4 \\
Stable Diffusion 3.5 Large Turbo & Stability AI & 1024×1024 & 4 \\
Stable Diffusion 3.5 Large & Stability AI & 1024×1024 & 35 \\
Stable Diffusion 3 Medium~\cite{esser2024scaling} & Stability AI & 1024×1024 & 30 \\
Stable Diffusion 3.5 Medium & Stability AI & 1024×1024 & 40 \\
GPT-Image-1 & OpenAI & 1024×1024 & -- \\
Midjourney v6~\cite{midjourney} & Midjourney & 1024×1024 & -- \\
Lumina 2~\cite{qin2025lumina} & Alpha-VLLM & 1024×1024 & 50 \\
HiDream~\cite{cai2025hidream} & HiDream.ai & 1024×1024 & 50 \\
Playground v2.5~\cite{li2024playground} & Playground AI & 1024×1024 & 50 \\
Playground v2~\cite{li2024playground} & Playground AI & 1024×1024 & 50 \\  %
Playground v1 & Playground AI & 512×512 & -- \\
Qwen-Image~\cite{wu2025qwenimagetechnicalreport} & Alibaba & 1024×1024 & 50 \\
\bottomrule
\end{tabular}
\label{tab:models}
\end{table*}

\begin{table*}[ht]
\centering
\caption{
Mean deanonymization accuracy for each target model.  
Values are averaged over all prompts and reported as percentages.
}
\small
\begin{tabular}{l c}
\toprule
\textbf{Target Model} & \textbf{Mean Accuracy (\%)} \\
\midrule
DALL·E 3 HD                 & 99.9 \\
SDXL Turbo                  & 99.9 \\
GPT-Image-1                 & 99.8 \\
Playground v2               & 99.7 \\
HiDream                     & 99.7 \\
Playground v2.5             & 99.6 \\
FLUX.1-Krea-dev             & 99.6 \\
Lumina 2                    & 99.5 \\
flux.1-kontext-pro          & 99.5 \\
Playground v1               & 99.4 \\
Stable Diffusion 3.5 Medium & 99.3 \\
Stable Diffusion 3 Medium   & 99.2 \\
Stable Diffusion 3.5 Large Turbo & 99.2 \\
Stable Diffusion XL         & 99.1 \\
Midjourney v6               & 99.0 \\
FLUX.1-schnell              & 98.9 \\
Stable Diffusion 3.5 Large  & 98.8 \\
Qwen-Image                  & 98.7 \\
Stable Diffusion 2.1        & 98.6 \\
Stable Diffusion v1.5       & 98.3 \\
FLUX-1.1-pro                & 97.9 \\
FLUX.1-dev                  & 97.8 \\
\bottomrule
\end{tabular}
\label{tab:mean_accuracy_per_model}
\end{table*}

\begin{table*}[t]
    \centering
    \small
    \caption{Deanonymization performance for our method across different image encoders.}
    \label{tab:image-encoder-results}
    \begin{tabular}{lccc}
        \toprule
        \multirow{2}{*}{\textbf{Image Encoder}} & \multicolumn{3}{c}{\textbf{Accuracy} (\%)} \\
        & Top-1 & Top-2 & Top-3 \\
        \midrule
        laion/CLIP-ViT-L-14-DataComp.XL~\citep{ilharco_gabriel_2021_5143773}  & \(87.86\) & \(94.71\) & \(96.29\) \\
        google/siglip2-large-patch16-512~\citep{tschannen2025siglip2multilingualvisionlanguage} & \(90.36\) & \(95.79\) & \(97.29\) \\
        openai/clip-vit-large-patch14~\citep{Radford2021LearningTV}    & \(84.07\) & \(92.71\) & \(95.64\) \\
        laion/CLIP-ViT-bigG-14-laion2B~\citep{ilharco_gabriel_2021_5143773}   & \(\mathbf{90.86}\) & \(\mathbf{96.14}\) & \(\mathbf{97.50}\) \\
        \bottomrule
    \end{tabular}
\end{table*}

\section{Details on Our Countermeasure}
\label{sec:details_defense}
\paragraph{Setup.}
We randomly sampled $100$ images to evaluate the effectiveness of our adversarial post-processing defense. Following prior work \citep{hu2025transferable}, we adopt the same optimization hyperparameters: a learning rate of $0.1$ and a contrastive-loss
temperature of $\tau = 0.1$. For the visual encoder, we use an ensemble of surrogate models listed in \Cref{tab:surrogate}. To test transferability across model families, the target encoder is set to \texttt{google/siglip2-large-patch16-512},
which is distinct from all surrogate models.
\begin{table*}[t]
\centering
\small
\caption{Local encoder ensemble used for adversarial optimization.}
\label{tab:encoders}
\begin{tabular}{l c}
\toprule
\textbf{Model} & \textbf{Training Dataset} \\
\midrule
ViT-H-14-378-quickgelu      & dfn5b \\
ViT-H-14-quickgelu          & dfn5b \\
ViT-SO400M-14-SigLIP-384~\citep{zhai2023sigmoid}    & webli \\
ViT-SO400M-14-SigLIP~\citep{zhai2023sigmoid}        & webli \\
ViT-L-16-SigLIP-384~\citep{zhai2023sigmoid}         & webli \\
ViT-bigG-14~\citep{ilharco_gabriel_2021_5143773}                 & laion2b\_s39b\_b160k \\
ViT-H-14-CLIPA-336~\citep{li2023clipav2}          & datacomp1b \\
ViT-H-14-quickgelu          & metaclip\_fullcc \\
\bottomrule
\end{tabular}
\label{tab:surrogate}
\end{table*}

\begin{figure*}[t]
\centering
\begin{subfigure}{0.23\linewidth}
    \centering
    \includegraphics[width=\linewidth]{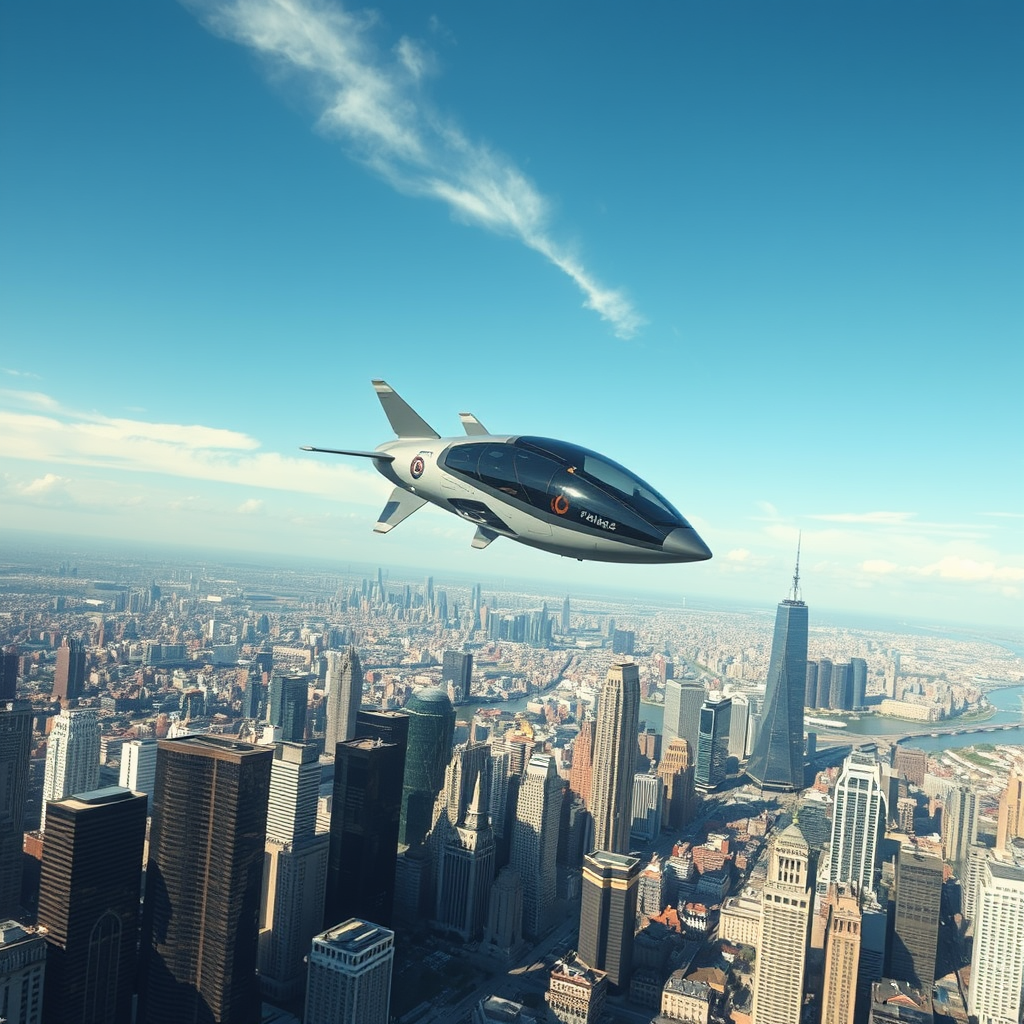}
    \caption{Original}
\end{subfigure}\hfill
\begin{subfigure}{0.23\linewidth}
    \centering
    \includegraphics[width=\linewidth]{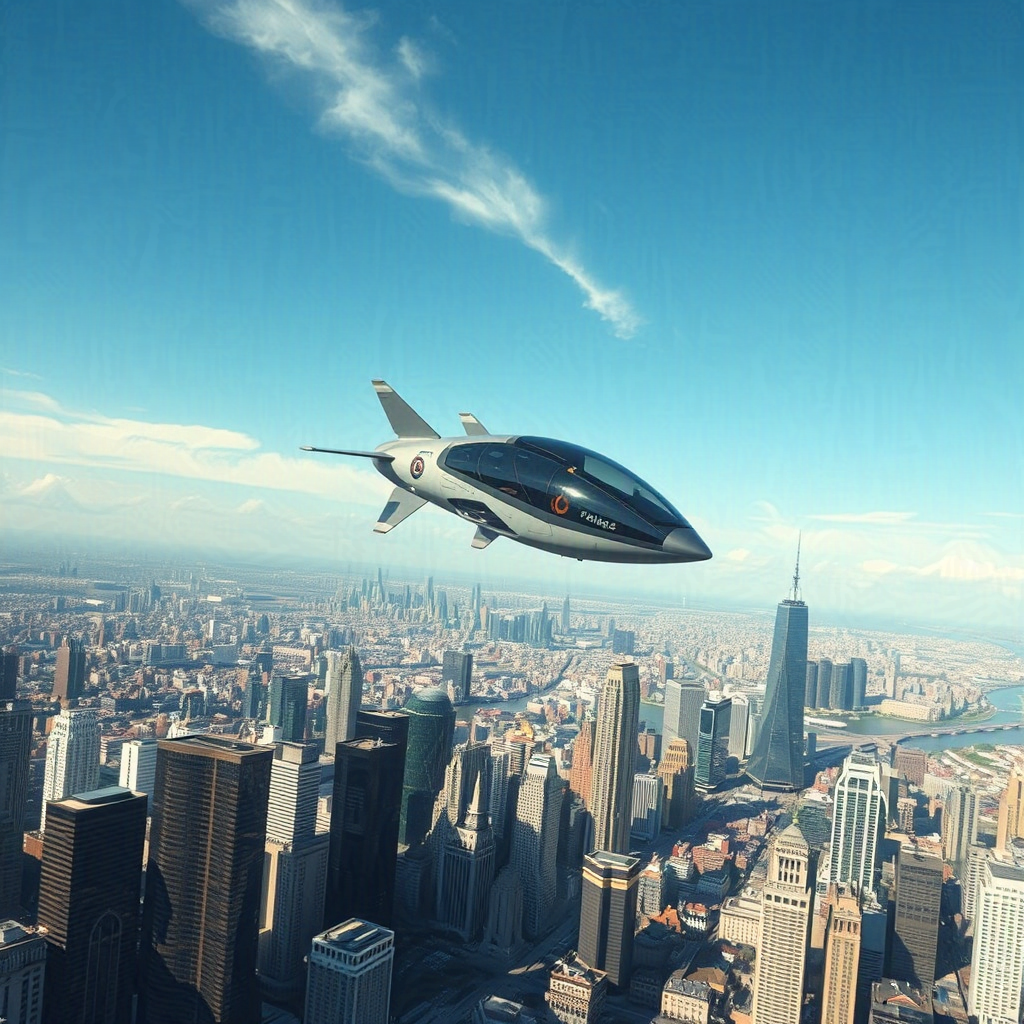}
    \caption{$\varepsilon = 2$}
\end{subfigure}\hfill
\begin{subfigure}{0.23\linewidth}
    \centering
    \includegraphics[width=\linewidth]{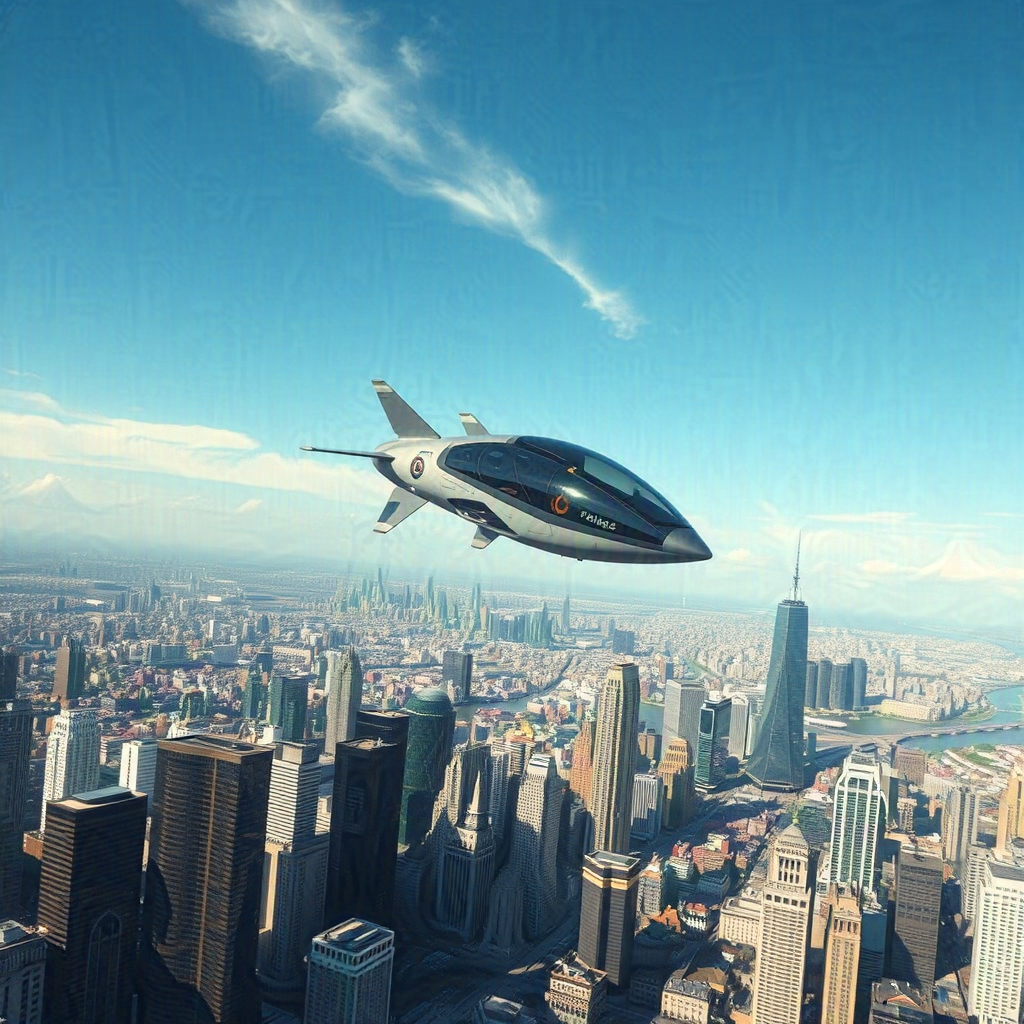}
    \caption{$\varepsilon = 4$}
\end{subfigure}\hfill
\begin{subfigure}{0.23\linewidth}
    \centering
    \includegraphics[width=\linewidth]{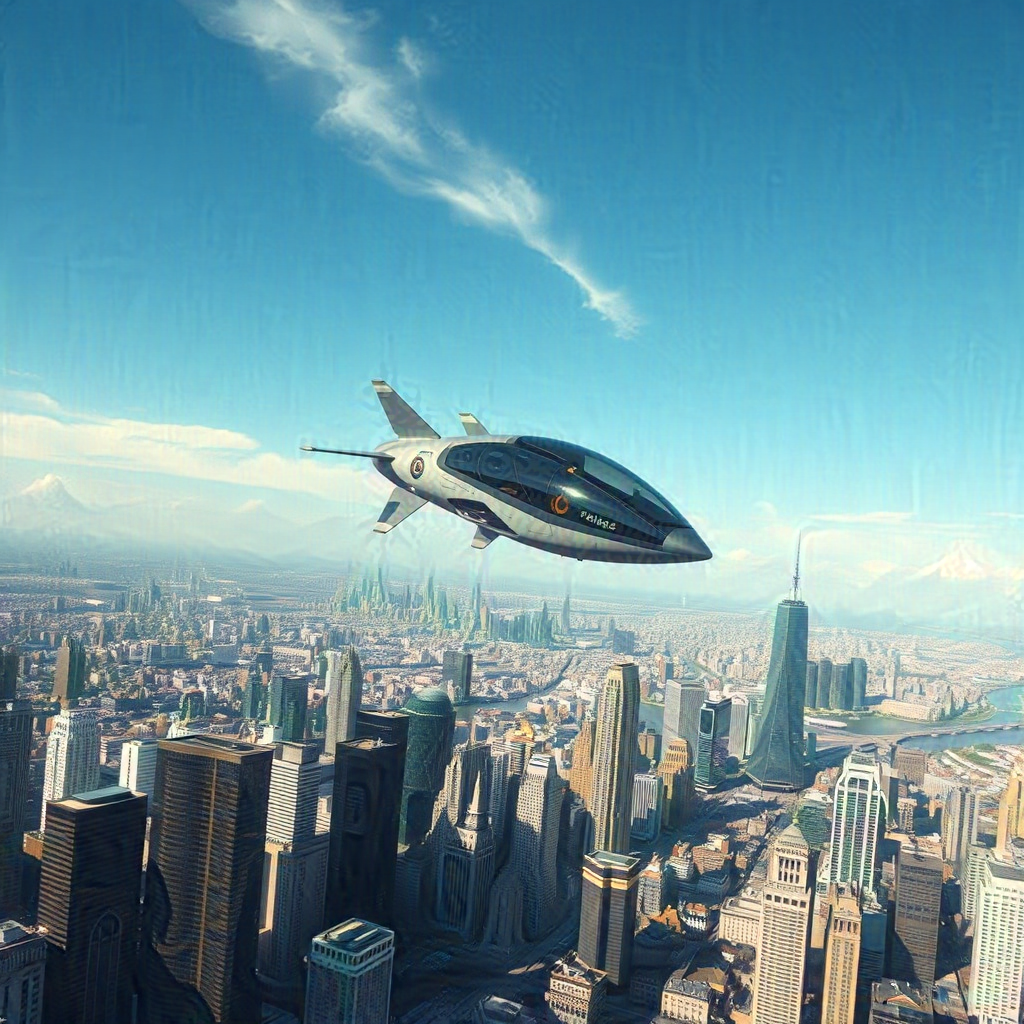}
    \caption{$\varepsilon = 8$}
\end{subfigure}
\caption{Original image and adversarial examples generated under different perturbation budgets $\varepsilon$. As expected, visual artifacts are more apparent for larger perturbation budgets.}
\label{fig:adv_examples}
\end{figure*}

\paragraph{Adversarial examples.}
\Cref{fig:adv_examples} shows the original image and our generated adversarial examples under different perturbation budgets $\varepsilon \in \{2, 4, 8\}$. As expected, larger $\varepsilon$ leads to more visible perturbations, but also yields stronger defense performance (see \Cref{tab:defense_results}).

\section{Automated Prompt Analysis}
\label{app:midjourney_prompts_analysis}

To better understand and identify any trends in distinguishability and the underlying prompt topics, we construct a pipeline on top of $\approx2.5M$ prompts collected from Midjourney \citep{naseh2024iteratively}.
We perform basic filtering and cleaning based on length (25-300 characters), language (English), and quality (requiring >75\% alphanumeric characters, no excessive repetition, and reasonable word lengths). Finally, MinHash deduplication was applied to remove near-duplicate prompts, reducing the dataset down to $\approx1.4$M prompts. We use datatrove \citep{penedo2024datatrove} for implementing these filters.

We embed all prompts using the \texttt{Qwen3-Embedding-4B}~\citep{qwen3embedding} model to obtain text representations for clustering.
 We then automate keyword extraction from each cluster using KeyBERT \citep{sammet2023domain}, extracting top-10 in the 1-3 word range across all prompts per cluster.
Each cluster was then assigned a concise label ($\leq 4$ words) using Llama 3.1 8B Instruct \cite{dubey2024llama}, which generated tags based on the cluster's keywords and 10 sampled prompts for context. The model was prompted to produce a single descriptive label. These LLM-assigned tags were mapped back to all prompts in their respective clusters.

\begin{table*}[t]
\centering
\footnotesize
\caption{Per-model deanonymization performance in the one-vs-rest setting, where the adversary has access only to its target model. Models are sorted in decreasing order of accuracy.}

\label{tab:one-vs-rest-no-access}
\begin{tabular}{lcccccc}
\toprule
\textbf{Target Model} & \textbf{Accuracy} & \textbf{FPR} & \textbf{FNR} & 
\textbf{TPR @1\%} & \textbf{TPR @5\%} & \textbf{ROC-AUC} \\
\midrule
playground\_v2\_5                & 0.986 & 0.003 & 0.292 & 0.871 & 0.988 & 0.993 \\
sdxl\_turbo                      & 0.985 & 0.000 & 0.318 & 0.971 & 0.983 & 0.997 \\
playground\_v2                   & 0.984 & 0.003 & 0.258 & 0.902 & 0.968 & 0.995 \\
gpt\_image\_1                    & 0.984 & 0.000 & 0.345 & 0.917 & 1.000 & 0.997 \\
hidream                          & 0.979 & 0.008 & 0.271 & 0.750 & 0.976 & 0.989 \\
stable\_diffusion\_3\_5\_large\_turbo & 0.969 & 0.017 & 0.305 & 0.544 & 0.864 & 0.974 \\
dalle3\_hd                       & 0.969 & 0.025 & 0.191 & 0.520 & 0.883 & 0.979 \\
flux.1-kontext-pro               & 0.954 & 0.035 & 0.304 & 0.534 & 0.851 & 0.961 \\
stable\_diffusion\_3\_5\_medium  & 0.954 & 0.032 & 0.404 & 0.331 & 0.702 & 0.951 \\
stable\_diffusion\_3\_5\_large   & 0.954 & 0.037 & 0.252 & 0.549 & 0.852 & 0.968 \\
stable\_diffusion\_3\_medium\_diffusers & 0.953 & 0.037 & 0.307 & 0.300 & 0.730 & 0.957 \\
lumina2                          & 0.944 & 0.044 & 0.381 & 0.178 & 0.651 & 0.922 \\
flux\_1\_krea\_dev               & 0.943 & 0.042 & 0.291 & 0.340 & 0.810 & 0.950 \\
flux\_1\_dev                     & 0.932 & 0.051 & 0.399 & 0.213 & 0.623 & 0.930 \\
midjourney\_v6                   & 0.931 & 0.059 & 0.276 & 0.229 & 0.707 & 0.939 \\
playground\_v1                   & 0.926 & 0.061 & 0.345 & 0.154 & 0.598 & 0.919 \\
flux\_1\_schnell                 & 0.921 & 0.071 & 0.250 & 0.274 & 0.682 & 0.919 \\
stable\_diffusion\_xl            & 0.908 & 0.079 & 0.367 & 0.179 & 0.510 & 0.923 \\
FLUX-1.1-pro                     & 0.896 & 0.092 & 0.318 & 0.260 & 0.557 & 0.915 \\
stable\_diffusion\_2\_1          & 0.876 & 0.116 & 0.272 & 0.177 & 0.466 & 0.891 \\
stable\_diffusion\_v1\_5         & 0.850 & 0.142 & 0.255 & 0.117 & 0.398 & 0.869 \\
qwen\_image                      & 0.694 & 0.306 & 0.304 & 0.087 & 0.235 & 0.717 \\
\bottomrule
\end{tabular}
\end{table*}

\end{document}